\documentclass[10pt,twocolumn,letterpaper]{article}

\usepackage{cvpr}              % To produce the CAMERA-READY version
\usepackage{tcolorbox}
\usepackage{placeins}
\tcbset{
  myboxstyle/.style={
    colback=gray!5,      % Background color
    colframe=gray!80,    % Frame color
    boxrule=0.5pt,       % Frame thickness
    arc=4pt,             % Rounded corners
    outer arc=4pt,
    boxsep=5pt,          % Separation between box and content
    left=5pt,            % Left padding
    right=5pt,           % Right padding
    top=5pt,             % Top padding
    bottom=5pt,          % Bottom padding
  }
}

\newif\ifclean

\newcommand{\cc}[1]{\ifclean{#1}\else {\color{brown}{#1}}\fi}
\newcommand{\KG}[1]{{\color{black}#1}} % purple
\newcommand{\SX}[1]{\ifclean{#1}\else {\color{blue}{#1}}\fi}
\newcommand{\CR}[1]{{\color{black}#1}}
\newcommand{\KGCR}[1]{{\color{black}#1}}

\newcommand{\tablestyle}[2]{\setlength{\tabcolsep}{#1}\renewcommand{\arraystretch}{#2}\centering\footnotesize}
\definecolor{LighterGray}{gray}{0.93}
\newcommand{\thickhline}{\Xhline{2\arrayrulewidth}}

\newcommand{\best}[1]{\textbf{\underline{#1}}} 
\newcommand{\sbest}[1]{\textbf{#1}}    
\newcommand{\tbest}[1]{\textit{#1}} 

\newcommand{\custompar}[1]{
  \par
  \vspace{2pt}
  \noindent\textbf{#1}
}

\cleantrue

\definecolor{cvprblue}{rgb}{0.21,0.49,0.74}
\usepackage[pagebackref,breaklinks,colorlinks,allcolors=cvprblue]{hyperref}
\usepackage{makecell}
\usepackage{pifont}
\usepackage{multirow}
\usepackage{arydshln}
\usepackage{booktabs}
\usepackage{float}

\usepackage{graphicx}        % For including images
\usepackage{caption}         % For \captionof
\usepackage{colortbl}        % For \rowcolor
\usepackage{afterpage}
\usepackage{needspace}
\usepackage{stfloats}
\usepackage{tabularx}
\usepackage[accsupp]{axessibility}

\def\paperID{5366} % *** Enter the Paper ID here
\def\confName{CVPR}
\def\confYear{2025}

\title{Progress-Aware Video Frame Captioning}

\author{
    Zihui Xue$^{1}$,\; Joungbin An$^{1}$,\; Xitong Yang\thanks{Work conducted as an independent researcher},\; Kristen Grauman$^{1}$ \\
    $^{1}$\KGCR{University of Texas} at Austin \\
}

\begin{document}
\maketitle

\begin{abstract}
While image captioning provides isolated descriptions for individual images, and video captioning offers one single narrative for an entire video clip, our work explores an important
middle ground: progress-aware video captioning at the frame level. This novel task aims to generate temporally fine-grained captions that not only accurately describe each frame but also capture the subtle progression of actions throughout a video sequence. \cc{Despite the strong capabilities of existing leading vision language models, they often struggle to discern the nuances of frame-wise differences.} To address this, we propose ProgressCaptioner, a captioning model designed to capture the fine-grained temporal dynamics within an action sequence. Alongside, we develop the FrameCap dataset to support training and the FrameCapEval benchmark to assess caption quality. The results demonstrate that ProgressCaptioner significantly surpasses leading captioning models, producing precise captions that accurately capture action progression and set a new standard for temporal precision in video captioning. Finally, we showcase practical applications of our approach, specifically in aiding keyframe selection and advancing video understanding, highlighting its broad utility.\footnote{Project webpage: \url{https://vision.cs.utexas.edu/projects/ProgressCaptioner}.}
% and effectiveness.  
\end{abstract}    

%\KGnote{title: consider if we want ``action progression" to be a highlight}
% \SXnote{To be decided: \\
% (1) title:
% \begin{itemize}
%     \item ProgressCaptioner: Towards Temporally Fine-grained Video Captioning at the Frame-level
%     \item Progress-Aware Video Captioning: Capturing Fine-Grained Temporal Dynamics / Action Progression at the Frame Level
%     \item Progress-Aware Video Captioning at the Frame Level
%     \item From Frames to Progress: Towards Temporally Fine-grained Video Captioning
% \end{itemize}
% (2) task name: progress-aware video captioning at the frame level?\\
% (3) method name: ProgressCaptioner?\\
% (4) dataset/benchmark name: FrameCap data; FrameCapEval benchmark?}
% \KGnote{names seem fine to me.  consider if we want to remove ``frame level" from the title? because it sounds more elegant. we can still emphasize frame level in abstract.  Could be simply ``progress-aware video captioning".} 

%\SXnote{uncomment clearntrue above to enable a clean version for the current page estimation}

% \KGnote{brown is places we could cut to save space.  I'm using track changes.}

\section{Introduction} \label{sec:intro}

%\KGnote{somewhere in intro, need to get in front of the ``each frame" issue, vs. which frames to caption.  reader will wonder within Intro.} \SXnote{see p4 added clarification}

Visual captioning~\cite{li2019visual}—the task of generating textual descriptions of visual content—is a fundamental problem in computer vision with extensive practical applications. Existing captioning paradigms are broadly divided into two categories: image captioning and video captioning, with a clear distinction between them. Image captioning~\cite{hossain2019comprehensive} generates a single, isolated description for each image, with no contextual linkage among different images. In contrast, video captioning~\cite{abdar2023review} assigns a single caption for the entire video clip, aggregating information across frames without addressing the specifics of individual frames.
% but failing to detach details at the frame level.

\begin{figure}[!t]
  \centering
   \includegraphics[width=1.0\linewidth]{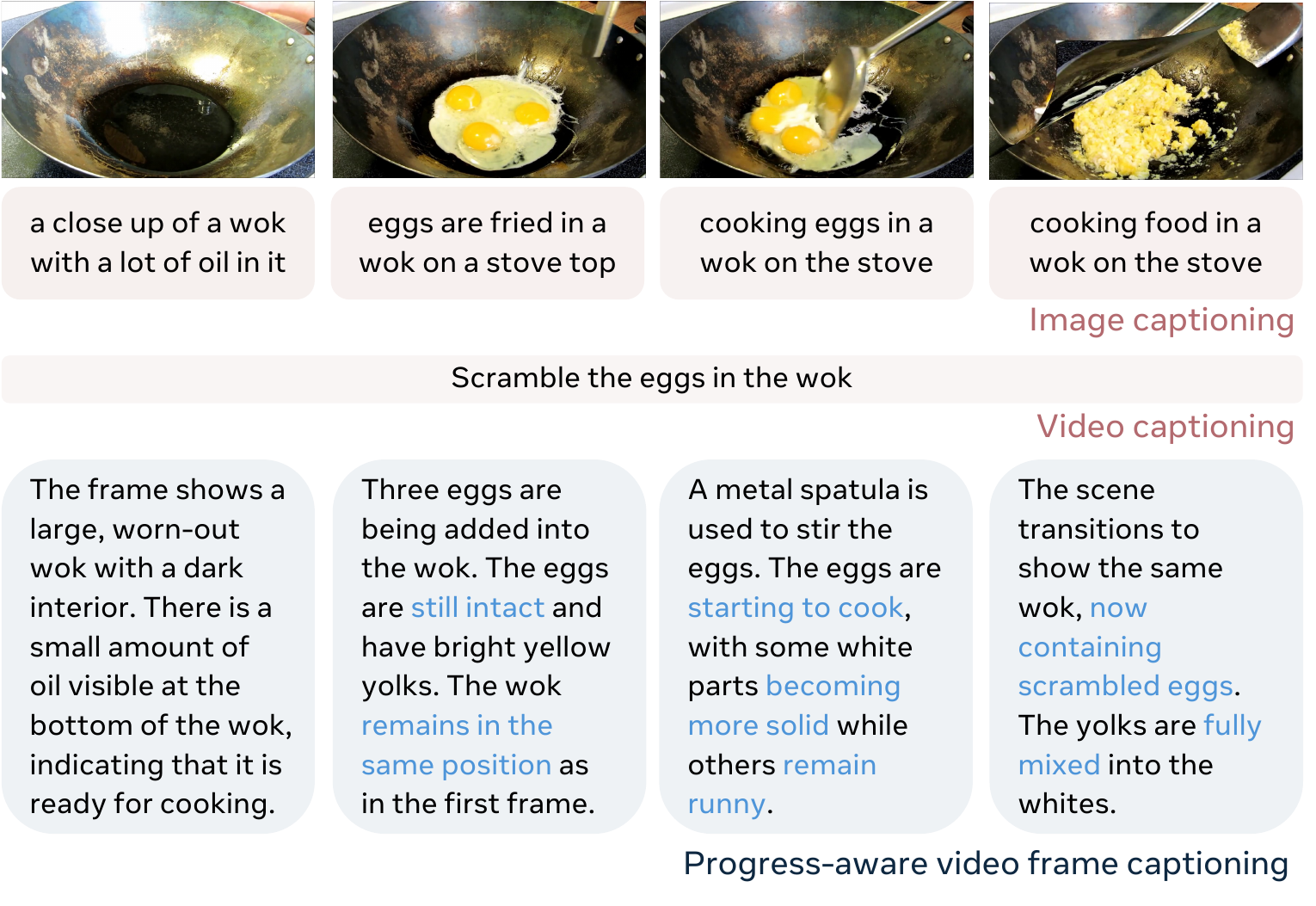}
   \vspace*{-5mm}
   \caption{We propose progress-aware video frame captioning \KG{(bottom)}, which aims to generate a 
   % caption sequence that corresponds to each individual frame within an action sequence.
  sequence of captions that capture the temporal dynamics within \KG{a video.} %an action sequence.}
   %\KGnote{``each individual frame" sounds alarming, or begs the question of how frames are selected... is there another way to phrase it here?} \SXnote{does this look better?}
   Unlike traditional image and video captioning \KG{(top)} that focus on broad event-level descriptions, our task delves into the detailed, progressive dynamics of an action, necessitating precise, temporally fine-grained %descriptive 
   capabilities.  \KG{Blue text highlights how the progress-aware captions build successively on the earlier content to highlight what is changing.}
   % \KGnote{``remains in same position" can also be made green, same with ``now containing"}
   } %(e.g., ``still intact", ``starting to cook", ``remain runny", ``fully mixed").}}
 %  \KGnote{is this going to show the failure modes of both image captioning and video captioning?  and last image says ``identical to the first frame" which is likely because it was generated for a pair, but frame 3 is not the ``first frame".}}
   \label{fig:intro}
\end{figure}

Figure~\ref{fig:intro} illustrates this dichotomy. Employing an image captioning model like BLIP~\cite{li2023blip} to describe each frame of the video results in captions that are \emph{local, not temporally context-aware}, and \KG{may exhibit} little variation across the sequence.  Conversely, video captioning provides a \emph{global, not temporally fine-grained} overview of the event, as exemplified by the YouCook2~\cite{youcook} ground truth label ``scramble the eggs in the wok''. In both scenarios, the nuances of how the action unfolds over time %\cc{or the detailed changes in the object's state throughout the sequence} 
are missed. This raises the question: Can we develop temporally fine-grained captions that capture the subtle, progressive nature of action sequences? % where such dynamics occur?  
\KG{Figure~\ref{fig:intro} (bottom) illustrates what we seek.}
% detailed, progressive dynamics of

\begin{figure}[t!]
    \centering
    \includegraphics[width=1.0\linewidth]{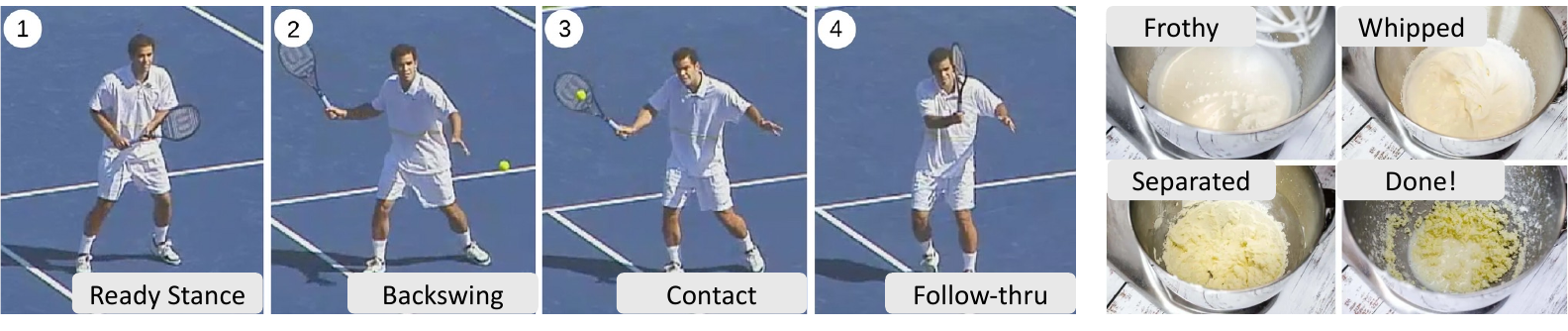}
    \vspace*{-4mm}
    \caption{Use cases of video frame captioning: finer-grained captions enable detailed, step-by-step guidance for daily tasks.
    % cooking tasks, detailed sports coaching, and precise robot learning from demonstration videos, surpassing the granularity of event-level descriptions.
    }
    \label{fig:usecase}
\end{figure}

Having such progress-aware captions could benefit a great variety of downstream tasks, bringing improved video understanding~\cite{zeng2017leveraging,yang2021just}, more precise video retrieval~\cite{wu2023cap4video,wu2024cap4video++,wang2023unified}, and enriched video generation~\cite{singer2022make,polyak2024movie}. Moreover, such a capability could open up new AR/VR and robotics applications. For instance, in AI coaching, a captioning system could meticulously analyze an expert's tennis forehand, %demonstration video, 
%dissecting each frame to provide precise descriptions that 
simplifying the learning process for users. Similarly, \KG{for how-to video creation, it could elicit and describe the key object state changes at each stage (e.g., ``how to make whipped cream")—useful for both content creators as well as visually-impaired users learning a new skill.} \CR{See Fig.~\ref{fig:usecase}}.  
% \KGCR{SHOULD WE TRY TO GET A FIG LIKE IN REBUTTAL?}
%in culinary applications, such a system could detail a ``whipping cream'' video, by capturing and describing key object state changes at each stage. 
%This detailed narration \KG{would also} offer a valuable aid to visually impaired individuals, helping them grasp  visual cues that are critical for understanding the process.

% , and the video-level event name is known. 
%To address this, %that seamlessly integrates the principles of traditional image and video captioning
\KG{Towards this end, we introduce a novel captioning} \KG{task}—\emph{progress-aware video captioning at the frame level}. %\KGnote{consider if we want to also bill ``action progression" at the top as well, since it is central and also because superficially ``video captioning at the frame level" sounds dry.}  
This task involves generating %a caption for each individual frame from an action clip
%frame-level 
captions %for \KG{each frame in} a sequence, 
%in a manner
that not only coherently depict action progression but also tailor each description specifically to its corresponding frame.\footnote{Without loss of generality, we obtain the input frame sequence by uniformly sampling from the action clips at a fixed rate (1FPS). These frames may or may not demonstrate visual action progression from one to the next, demanding that the model discern the difference when generating progress-aware captions.}  % effectively capture and convey the temporal dynamic, thereby being progress-aware.}
%\KGnote{``each individual", same comment} \SXnote{revised}
% \cc{See Figure~\ref{fig:intro} for an example of the desired caption sequence.} 
Our task is uniquely characterized by its demand for fine-grained temporal sensitivity. By ``fine-grained'', we refer to generating detailed descriptions that elucidate the stages or procedural steps of the action, effectively conveying how the action is performed throughout the video sequence. 

As discussed, traditional video captioning~\cite{chen2011collecting,xu2016msr,zhou2018towards,wang2019vatex} settles for broad event-level descriptions, where a description like scrambling eggs for the video in Figure~\ref{fig:intro} would be considered entirely accurate. In contrast, we seek progress-aware captions that detail each stage of the action, such as ``eggs still intact'', ``starting to cook'' and ``fully mixed''.
\cc{\SX{While recent works~\cite{chai2024auroracap,wang2024tarsier,chen2024sharegpt4video,llava-video,cai2024temporalbench,chen2024panda} enhance the overall descriptiveness of video captions,} they continue to produce a single video-level description without distinguishing the nuances \KGCR{across time}.  %between individual frames. 
Our task delves deeper, exploring how each frame contributes to the narrative of the action's progression, thereby setting a new standard for fine-grained temporal precision in video captioning.} 
% Different from previous research~\cite{chen2015microsoft, you2016image, vinyals2016show, agrawal2019nocaps,chen2011collecting,xu2016msr,zhou2018towards,wang2019vatex} that aims at general, coarse-grained event-level descriptions, our task delves into the temporally fine-grained aspect of captioning—given a video with \emph{known} human activity, the model is challenged to accurately describe the temporal progression within this specific action.  
%\KGnote{in intro or in onset of approach, we need to clearly motivate why ``known activity" is not a strong assumption.  I guess it's that inferring the overall activity (``what") is pretty accurate with SOTA models.  should we even have an experiment that demonstrates this?  rather than provide human activity label, we infer it first?} \SXnote{removed this requirement, it's optional to provide action label during inference}
% In all, the task necessitates a significant enhancement in current captioning models’ capability to describe temporal progression precisely, marking a departure from the more general, coarse-grained event-level descriptions that have dominated previous research.

\begin{figure}[!t]
  \centering
   \includegraphics[width=1.0\linewidth]{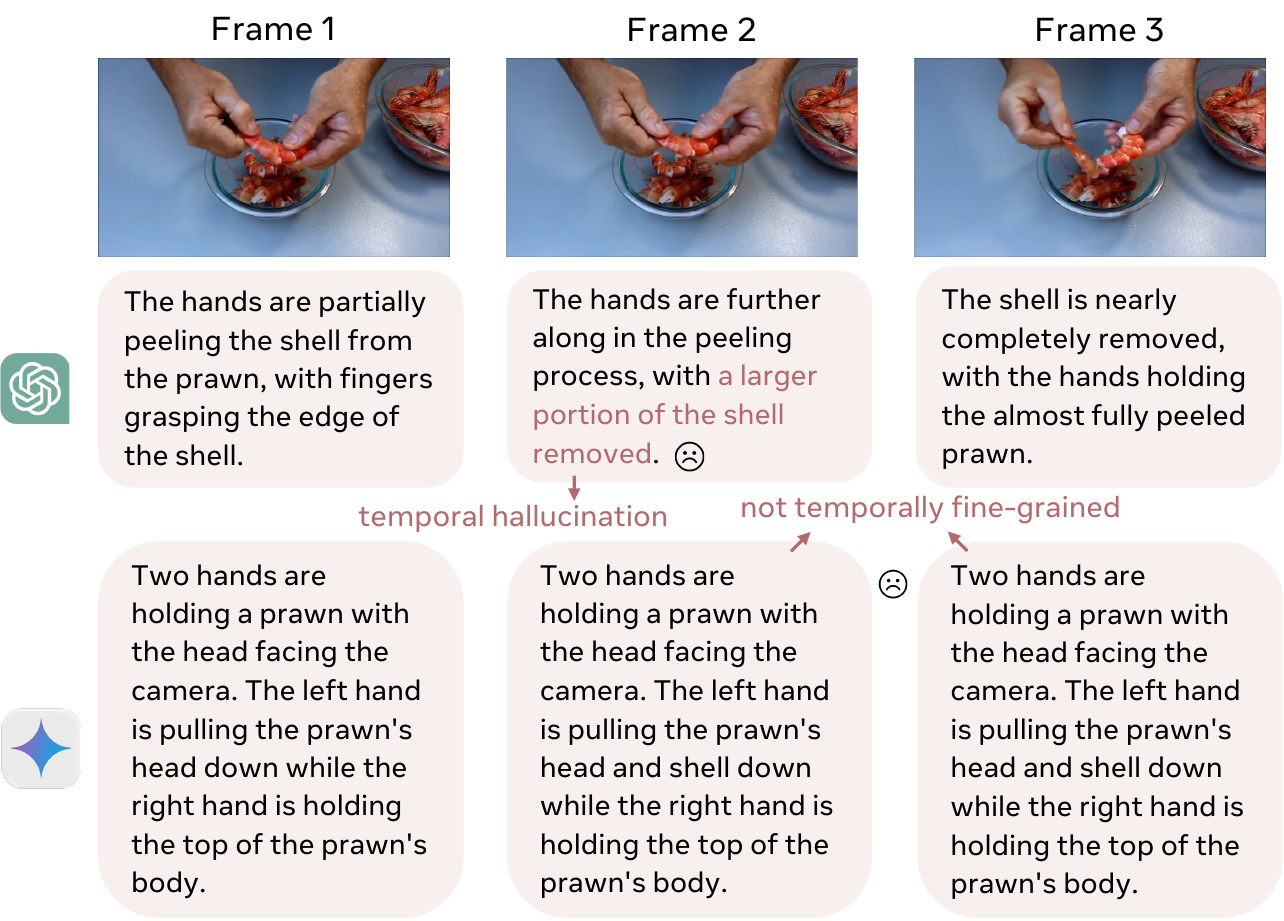}
   \vspace*{-4mm}
   \caption{Issues of existing VLMs in video frame captioning: (1) Lack of temporal granularity. See captions for frames 2 and 3, produced by Gemini-1.5-Pro (row 2), which fail to distinguish subtle differences between the frames. (2) Temporal hallucination. See frame 2's caption produced by GPT-4o (row 1), which inaccurately suggests progression that is not visible.}
   \label{fig:vlm_issue}
\end{figure}

% \KGnote{I reduced this para some, and I think we could condense more if needed.}
Despite great advancements of vision language models (VLMs)~\cite{li2023blip,ye2023mplug,videollama,llava,liu2024improved,lin2024vila,qwen2,achiam2023gpt,llava-ov,llava-video} that have markedly improved visual captioning, we observe that these models still struggle with this nuanced task. Two main issues persist: first, the lack of temporally fine-grained captions; when shown adjacent frames that depict subtle variations in action progression (such as frames 2 and 3 in Figure~\ref{fig:vlm_issue}), the generated captions can be overly coarse, failing to differentiate between the frames (see row 2, Gemini-1.5-Pro's captions). Second, we identify and term a notable issue of ``temporal hallucination'', where the captions suggest temporal progression in disagreement with what the visual frames exhibit. See frame 2 of Figure~\ref{fig:vlm_issue}, where GPT-4o's generated captions (row 1) incorrectly advance the action sequence. 
% when the frames themselves show no visual change. 
%\KGnote{is the hallucination only when there is no visual change, or also where there's a different visual change than the one described?}  
% This issue is prevalent in VLMs that integrate strong language models. 
The prevalence of such errors can be attributed to models' reliance on the common statistics of activity sequences, which %overshadow the %necessity of matching individual statements accurately to corresponding 
%need to
mistakenly override matching specific statements to specific frames. 
% \cc{While the described action progression may occur within the video, the models are not sufficiently trained to link specific action stages to the appropriate frames.}
\KG{Meanwhile, image captioners—even if trained with fine-grained annotations—treat frames in isolation and hence lack the temporal context to say what is \emph{progressing} versus what is \emph{present}.}  %%%  added this, do you think we can back this up better in teh approach explanation below?
\CR{
We introduce ProgressCaptioner, a model for generating progress-aware frame-level video captions. Our approach uniquely interleaves pseudo labeling with two learning stages. Stage I develops a \emph{frame pair} captioning model, and stage II extends this to \emph{full frame sequences}. This process also creates the FrameCap training dataset, comprising action videos along with high-quality frame-wise captions, which are initially generated by an ensemble of VLMs and then filtered using our proposed evaluation methods.}

To assess the quality of frame-wise captions and benchmark ProgressCaptioner against leading VLMs, we introduce the FrameCapEval benchmark comprised of %. This benchmark sources 
videos from four public video action datasets. %, alongside a human study. % and incorporates the two novel automatic evaluation tasks we design, along with a human study for additional validation. 
ProgressCaptioner consistently outperforms leading open-source VLMs with a 1.8$\times$ to 2.7$\times$ improvement in caption quality and also achieves the highest selection rate in user studies, even surpassing the much larger proprietary models GPT-4o~\cite{achiam2023gpt} and Gemini-1.5-Pro~\cite{reid2024gemini}.
% The results underscore the superior performance of ProgressCaptioner: it consistently delivers high-quality, temporally fine-grained, and progress-aware captions.  
%\KGnote{add a punchline on results here - make a big deal about outperforming even much large proprietary VLMs on fine-grained action captioning, e.g., by as much as X\% relative gains despite at least 10$\times$ smaller model (and Y\% relative gains over similarly sized models?}
Finally, we highlight potential applications enabled by our advanced captions: %with two use-cases: 
keyframe selection and enhanced video understanding. We hope that our task, model, and benchmark can inspire future development in temporally fine-grained video captioning. 

%\KGnote{we'll need more here to hint about technical novelty claims.  novelty of the problem is clear so far, but novelty of approach is not stated. e.g., should we give a hint about how the pseudolabeling is clever?}

% To address these challenges, we propose sliding captioning, that utilizes a frame pair as the functional unit, to effectively balance between incorporating context from adjacent frames and generating precise frame-wise descriptions. Motivated by this, we develop a data collection pipeline to gather large-scale frame pairs from existing video datasets. This process employs multiple captioning models alongside a filtering mechanism, leveraging VLMs in the loop as a judge... dataset statistics... Furthermore, we establish a comprehensive evaluation benchmark to assess the quality of frame-wise captions produced existing models...our model training + results highlights

\section{Related Work} \label{sec:related}

% \subsection{Image Captioning}
\custompar{Image Captioning}
Image captioning %, situated at the intersection of computer vision and natural language processing, 
has been extensively studied in recent years~\cite{chen2015microsoft, you2016image, vinyals2016show, agrawal2019nocaps}. A related line of work is image difference captioning, where the task 
is to describe differences between two images~\cite{jhamtani2018learning, park2019robust, sachdeva2023change} or sets of images~\cite{dunlap2024describing}. Building on the success of generative models, recent benchmarks~\cite{awal2024vismin, jiao2024img,yu2014fine} 
%\KGnote{interesting, these sound similar to my student Aron Yu's work on fine-grained attribute learning, can also cite} \SXnote{added} 
challenge models to distinguish between two visually similar images, 
% (one original and one synthesized), % based on a designed prompt),
advancing fine-grained image understanding. However, %the discussion is 
\KG{all the above models are} restricted to static \SX{(typically synthesized)}
% \KG{(typically independently captured)} 
image pairs and address coarse-grained differences like object presence or absence.  Temporal intricacies—accurately describing how an action progresses—remain unexplored.
% focusing on spatial differences (i.e., object presence, attributes, or spatial relationships). 
%%\KGnote{are we airtight on the contrast with the image difference captioning work?  one could argue that a temporal change is equivalent to seeing the object/spatial changes from one image to the next.  The refs~\cite{jhamtani2018learning, park2019robust, sachdeva2023change,dunlap2024describing} seem most important to be very strongly distinct.  are any of them possible baselines?} \SXnote{Does this contrast (static image pairs vs consecutive frames in a sequence) looks good? For baselines, all these adopt old model architectures, require training data and won't be better than current VLMs}

% \subsection{Video Captioning}
\custompar{Video Captioning}
Video captioning~\cite{abdar2023review} aims to produce a single description that encapsulates %the overarching content of a given 
a video clip. While traditional benchmarks~\cite{chen2011collecting,xu2016msr,zhou2018towards,wang2019vatex} offer a brief %coarse-level 
one-sentence caption for each video, recent efforts expand this scope, %. These improvements include 
extending captioning 
to hours-long videos~\cite{islam2024video}, enriching the granularity of details~\cite{chen2024sharegpt4video,chai2024auroracap,wang2024tarsier}, enhancing caption uniqueness~\cite{perrett2024s}, integrating a \KGCR{causal} temporal narrative~\cite{nadeem2024narrativebridge}, or introducing LLM summarization~\cite{li2024wolf}.
% adopting a world summarization framework~\cite{li2024wolf}. 
%\KGnote{what does ``world summarization" mean?} \SXnote{their paper title, revised}

%\noindent
Adjacent to traditional video captioning are the tasks of visual storytelling~\cite{huang2016visual,li2019video} (creating a coherent story for a sequence of snapshots), dense video captioning~\cite{krishna2017dense, yang2023vid2seq, zhou2024streaming} (temporal localization and captioning of all events in an untrimmed video), audio description~\cite{han2023autoad,han2023autoad2,han2024autoad} (detailed narrations of visual events in videos (e.g., movies) for visually impaired audiences), \CR{and video paragraph captioning~\cite{yu2016video} (producing a multi-sentence paragraph describing the video)}. However, \KG{all} these works \KG{still} address ``what is happening'' at a coarse-grained event level, e.g., noting that someone is making a souffle within a specific time range. 
%KGnote{butter - that's a pretty fine-grained  example, pick something less low-level? like `making a souffle"?} 
%A more temporally fine-grained understanding at the frame level—accurately describing frames within the 
The ability to break down frame-level details—such as whisking egg whites, folding ingredients, and observing the souffle rising—is still lacking.
% to depict various melting stages—is still lacking.  
%\KGnote{1. if we are making this the distinguishing factor, then it seems we have to be more specific about what is level of fine-grainedness we mean in the intro; how do we characterize that boundary?  is it object state changes plus body pose changes specifically, or...?  }
%as described now, it sounds like we are tackling a very niche and potentially unimportant area (explain different steps as the butter melts) but I don't think that's actually the case.
%2. how good are these methods \cite{han2023autoad,han2023autoad2,han2024autoad,yang2023vid2seq, zhou2024streaming}?}
%\SXnote{1. action description, what is the action in the video vs. how an action progresses in the video? Our model can also do the first, just the first (what) problem is pretty mature while second (how) is not well addressed? Shall we state this in intro more directly?} \KGnote{if the made-up example above (butter) can be swapped with something more telling, it will help.  and yes in intro could we say a one-liner about what qualifies as ``fine-grained", what is the level of detail we seek?} \SXnote{added in Intro p4}
%\SXnote{2. I think current VLMs are better than them, also these models are trained to give brief event-level descriptions for a very long video, quite different from our setting: frame-wise descriptions for a relatively short action clip.}
% missing the subtle nuances and intricate temporal progression that occur within an event.

% \subsection{Visual Language Models}
\custompar{Vision Language Models} Recent advancements in VLMs~\cite{li2023blip,ye2023mplug,videollama,llava,liu2024improved,lin2024vila,qwen2,achiam2023gpt,llava-ov,llava-video} have greatly enhanced the capabilities of both image and video captioning.
% Contemporary VLMs are crafted by integrating a vision encoder with expansive language models (LMs) through projection layers. 
Despite their strong performance, VLMs often exhibit ``hallucination''~\cite{wang2024videohallucer,guan2024hallusionbench}, and 
% a common issue where they produce inaccurate or unsubstantiated responses not supported by the visual inputs. 
\SX{preference learning~\cite{rafailov2024direct,zhao2023beyond} has proven effective in mitigating this issue.}

%\noindent
Compared to image-LMs, video-LMs crucially require the integration of temporal dynamics understanding, spurring a series of work on evaluating temporal perception~\cite{grunde2021agqa,xiao2021next,liu2024tempcompass,li2023vitatecs}. %, which typically evaluates the ability to answer general questions about the video content. 
While these assessments ensure that a model can generate an accurate overall video summary or answer general questions, they entail neither temporal localization %\SXnote{not sure if it is accurate? nextQA we tested, has A appears before/after B questions, and is sort of related to temporal localization?} nor %the model's ability to  %%% KG that's not localization though.
nor discernment of  %and describe the %temporally 
fine-grained differences between frames. % within the sequence.}
% These benchmarks typically assess through question-answering formats, and do not cover the topic of frame-wise captioning.
%\KGnote{convey more explicitly that Q\&A  means that overall summaries are enough and temporal localization not needed?}
%Similar 
%\KG{Related} to our formulation,
The OSCaR benchmark~\cite{nguyen2024oscar} \KG{focuses on} %proposes
object state change (OSC) captioning, yet it is limited to just three frames and specifically OSC videos, with models and captions not publicly released yet preventing direct comparison. Additionally, their approach relies on human annotation and a single advanced GPT model. In contrast, our approach features a scalable data collection pipeline that reduces reliance on these labor-intensive resources, employs novel automatic evaluation tasks, and broadens the scope beyond OSC videos.  %\KG{using an unreleased dataset}.
%\KGnote{this contrast to OSCaR does not sound strong/convincing enough.} \SXnote{their dataset is not publicly released so we can not test our model; shall we emphasize how our pseudo labeling and model development is uniquely designed to contrast with them here? may be a bit long?} \KGnote{not sure - it's the closest work right? so it may be useful to get ahead of it but briefly and not too defensive?} \SXnote{yes it's the closet in terms of the problem formulation; they do not really have a method, more of a benchmark paper, updated the text}
%Finally, another body of work
Finally, unlike methods for long-form video and event localization with VLMs~\cite{mangalam2023egoschema,chen2024rextime,ren2024timechat}, % explores VLMs' capability to understand long-form videos and localize events, whereas 
our focus is distinctly more temporally fine-grained, concentrating on how individual frames evolve within a single event. 

\section{Approach}  \label{sec:method}

\begin{figure}[!t]
  \centering
   \includegraphics[width=1.0\linewidth]{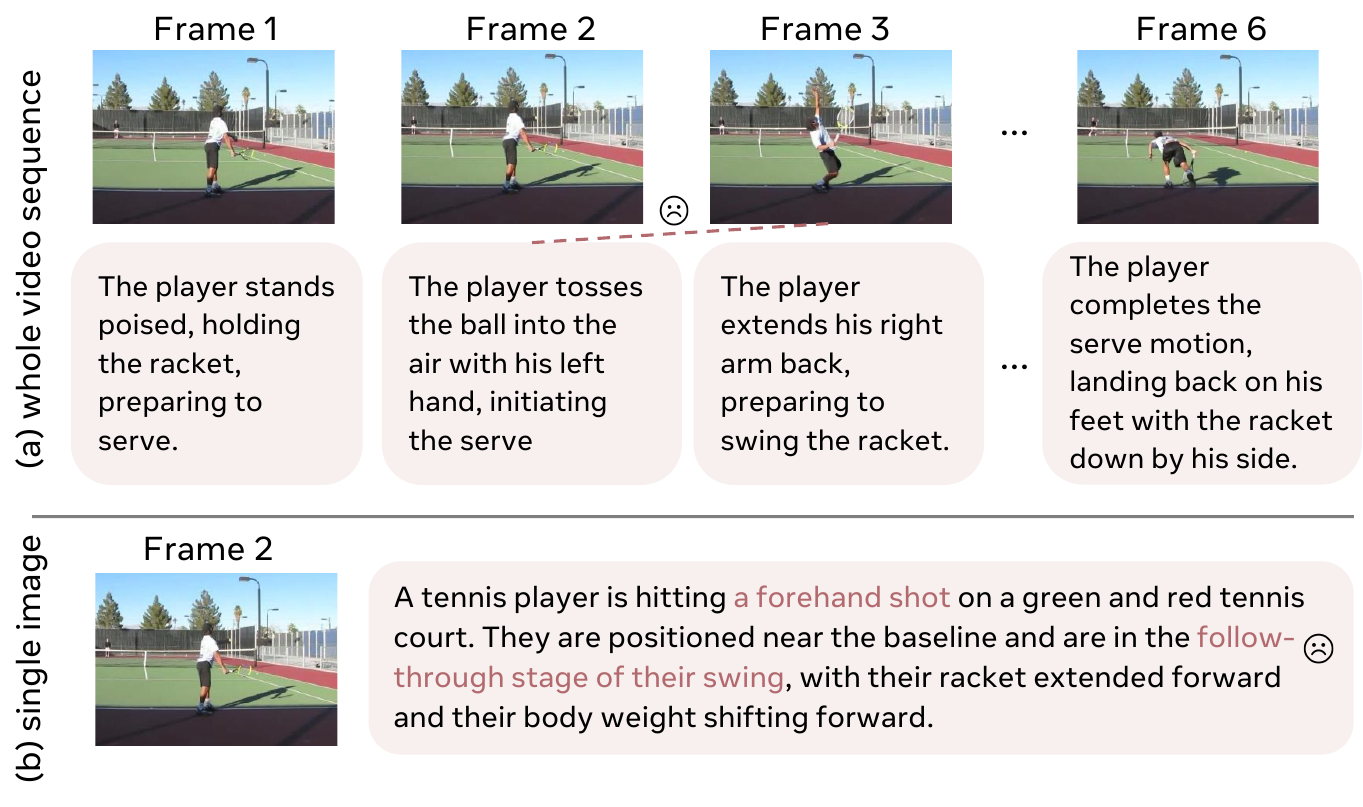}
    \vspace*{-7mm}
   \caption{Captioning outcomes using Gemini-1.5-Pro~\cite{reid2024gemini}.} % on a tennis serve video sequence. 
   % \cc{(a): providing the entire sequence leads to a brief and temporally misaligned caption sequence; (b): captioning individual frames results in inaccurate captions that lack context and temporal coherence.}} %\KGnote{is the long caption meant to be the output for the single image/frame 2 only? also the images are so small, maybe crop to the tennis player more?} \SXnote{yes; updated the figure} 
   %\KGnote{there is some redundancy between the first 3 figures of the paper.} \KGnote{could reduce this caption a lot since text already unpacks  it.}}
   \label{fig:context}
\end{figure}
%\KGnote{overall the approach reads a bit like string of heuristics if the reviewer is feeling grumpy.  try to counter this by giving foreshadowing of how this is going to work, make clear where the particularly insightful and novel steps are, and keeping it formally written.  Maybe also an intermediate method fig with real examples (in addition to the flowchart).  I tried to do this a bit but you may be able to improve it more.} \SXnote{rewrote sec 3.2}

% In Sec.~\ref{sec:method_task}, we delve into the specific challenges and nuances of the frame-wise video captioning problem. Building on this analysis, we detail the development of our model, ProgressCaptioner in Sec.~\ref{sec:method_model}. 
We delve into the specific challenges of our progress-aware video frame captioning problem in Sec.~\ref{sec:method_task} and outline ProgressCaptioner's development in Sec.~\ref{sec:method_model}. 

\subsection{Progress-aware Video Frame Captioning} \label{sec:method_task}

\custompar{Problem Formulation} Our objective is to develop a captioning model that, \KG{given a video}, produces accurate temporally fine-grained captions. Formally, for a sequence of $T$ frames, denoted as $\mathcal V = \{v_i\}_{i=1}^{T}$, the captioning model generates a corresponding sequence of captions $\mathcal C = \{c_i\}_{c=1}^T$, where each $c_i$ describes the $i$-th frame $v_i$ \KGCR{(recall we sample at 1FPS)}. This captioning process has three key requirements% \SX{to capture progress awareness}
: (1) \emph{Accuracy}, where each caption $c_i$ must faithfully represent what is visually occurring in frame $v_i$, without hallucinating from the context of other frames; (2) \emph{\SX{Temporal Specificity}}, where each caption $c_i$ specifically attends to $v_i$, without being overly generic to be applicable to multiple frames in the sequence; \SX{(3) \emph{Progressive Coherence}: The sequence of captions $\{c_i\}_{i=1}^{T}$ should build upon each other to reflect the essential changes in the action over time.} 
%\KGnote{is item 2 better dscribed as progress-aware (vs temporal granularity)?  and should we include that we want the successive captions to build on each other, in the sense of describing essential things that are changing over time?} \SXnote{I feel progress-aware encapsulates both: (1) when there is no progression, should avoid hallucination; (2) when there is progression, should capture. About (3) progressive coherence, I agree it is a requirement, yet we do not have a good evaluation metric corresponding to this, would adding a (3) raise reviewer's doubt on how we evaluate this? the two evaluation tasks are about (1) (2)}

%\KGnote{is this the para to state formally what is the fps assumption / how we get the $T$ frames worth captioning from a typical video? after foreshadowing the ``each frame" issue I flagged in intro.} \SXnote{added in intro}

\custompar{FrameCap Dataset} To train our captioning model, we require a dataset that pairs frame sequences ($\mathcal V$) with corresponding captions ($\mathcal C$). Existing datasets~\cite{chen2011collecting,xu2016msr,zhou2018towards,wang2019vatex} provide only a single, generic caption for an entire video clip, lacking the frame-wise caption format we need. To address this gap and train our model, we develop the FrameCap dataset. Given the prohibitive expense of collecting human-labeled caption sequences as our ground truth ($\mathcal C$), especially at scale, we leverage leading VLMs as powerful tools to create a pseudo caption sequence $\hat{\mathcal C}$ from $\mathcal V$. For video sources, we refer to two large-scale datasets that focus on fine-grained human activities: HowToChange~\cite{xue2024learning} (featuring object state change videos from YouTube) and COIN~\cite{tang2019coin} (featuring daily activities from YouTube). 
% Both datasets are organized hierarchically, enabling more fine-grained video analysis that goes beyond coarse-grained event descriptions.

\custompar{Caption Sequence Construction} Prompting VLMs for our desired caption sequence is nontrivial. %To address this challenge, 
We identify two key problems: (1) \emph{Input considerations}: how many context frames from $\{v_i\}_{i}^T$ should be provided? %to the VLM to accurately generate a caption for the current frame? 
% for accurate captioning?
(2) \emph{Output assessment}: what issues arise in VLM-generated captions, and how can we filter to retain only high-quality ones? To explore these questions, we conduct preliminary experiments by prompting
% \footnote{Detailed prompts are provided in the Supplementary.} 
leading VLMs to perform the frame-wise captioning task. We share our findings below.

\custompar{Observation I} 
%First, regarding the number of input context frames, the most intuitive answer \KG{would} seem to be all $T$ frames. 
Intuitively, inputting all $T$ frames would seem best. 
% \cc{As humans, with the ability to view the entire sequence, we can most effectively describe any single frame in a way that differentiates it from the others}. 
However, current VLM capabilities do not support this extensive context. Specifically, providing too many frames at once often leads to descriptions that lack detail and exhibit temporal inaccuracies, with VLMs also risking memory overload, as similarly observed in~\cite{chen2024sharegpt4video}. Conversely, providing a single frame at a time reduces the task to image captioning, which is not optimal either, resulting in captions that lack temporal context and coherence.

Figure~\ref{fig:context} shows a \KG{representative} trial with Gemini-1.5-Pro~\cite{reid2024gemini}. % on a tennis serve video. 
Inputting the full sequence (case (a)) yields brief per-frame descriptions with temporal misalignment (i.e., the second caption erroneously describes what is visually occurring in the third frame). On the other hand, captioning frames in isolation (case (b)) removes essential temporal context, where the model mistakes the initial stage of a tennis serve for the follow-through of a forehand swing. 
These findings underscore the importance of finding a balanced approach and motivate us to adopt a \emph{frame pair} as the stepping stone of our captioning model. % development.

\begin{figure*}[!t]
  \centering
   \includegraphics[width=0.85\linewidth]{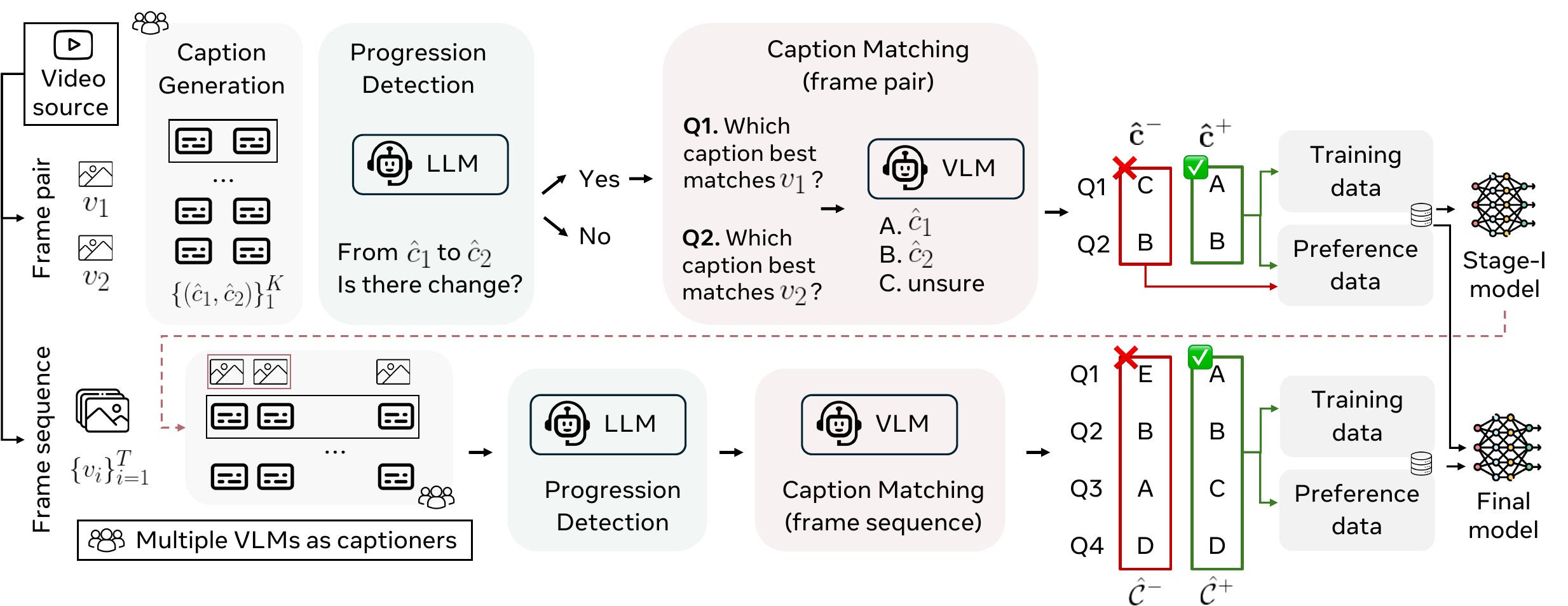}
   \vspace*{-2mm}
   \caption{Framework of ProgressCaptioner, designed in two stages. In Stage-I, we prepare frame pairs and generate corresponding caption pairs using multiple VLMs. Each pair undergoes our designed progression detection and caption matching evaluations, to decide if they are selected for model supervised fine-tuning or rejected, with the latter contributing to preference data to aid in model preference learning. 
   The Stage-I model training then proceeds using this collected data.
   In Stage-II, the trained stage-I model labels frame sequences with a two-frame sliding window, in conjunction with other VLMs. These sequences are again assessed through progression detection and caption matching to classify them as selected or rejected. All collected data from both stages contribute to the final training of ProgressCaptioner. %\KGnote{in caption matching box should we indicate items are permuted randomly across samples?} \SXnote{you mean the multi-choice options? we do not permute actually}
   }
   \label{fig:method}
   \vspace*{-4mm}
\end{figure*}
% we discuss how to assess the corresponding caption pair ($\hat c_1, \hat c_2$) produced by existing VLMs.
\custompar{Observation II} Next, building upon the use of a frame pair ($v_1, v_2$), is the caption pair ($\hat c_1, \hat c_2$) produced by existing VLMs of sufficient quality to be directly adopted? Our preliminary experiments reveal two main issues: (a) lack of temporal %\KGnote{is ``temporal" the right word here?} \SXnote{try to say for two adjacent frames, their captions are not fine-grained enough to tell the two frames differences, what word would be better here?} 
granularity, and (b) temporal hallucination, as showcased in Figure~\ref{fig:vlm_issue}. 
To dissect these issues, we analyze the captions in relation to the visual progression between frames $v_1$ and $v_2$. Specifically, if there is a visible progression from $v_1$ to $v_2$ (e.g., the slight peeling of a shrimp's shell from frame 2 to frame 3 in Figure~\ref{fig:vlm_issue}), the captions should adequately reflect this change. Overly similar captions in such a scenario signify a failure in temporal granularity. Conversely, when there is no change between frames (e.g. frames 1 and 2 in Figure~\ref{fig:vlm_issue}), the captions should remain consistent. We %identify the problem as a 
deem it a temporal hallucination when captions erroneously indicate progression in disagreement with the visuals. 
% This issue often arises in VLMs equipped with strong language decoders, which can introduce a language bias toward generating unwarranted narrative changes.
\CR{This deficiency in existing VLMs motivates our development of a new captioning model and specialized evaluations for high-quality caption selection.}

\subsection{ProgressCaptioner} \label{sec:method_model}
% Having thoroughly analyzed the problem and armed with the new insights discussed above, we now detail the development of our captioning model, ProgressCaptioner.
The observations above drive the design of our model, ProgressCaptioner, which unfolds in two stages. Based on our findings that current VLMs have trouble maintaining caption quality when handling extensive $T$-frame inputs, our approach begins with frame pair captioning. In the first stage, we develop a ProgressCaptioner to excel at describing the nuances between adjacent frames. The second stage then leverages the first-stage model to pseudo label the full $T$-frame sequence with a two-frame sliding window. This staged approach refines caption quality along with model development, enhancing the captioning process iteratively with more precise pseudo labels.

% We propose a two-stage pseudo labeling pipeline interleaved with our model development, leveraging several leading VLMs as captioners. \KG{After amassing confident pseudo-labels $\langle v_i, \hat{c}_i \rangle$, Stage I trains a model for two-frame pairs. Using the resulting model, Stage II extends the two-frame model to the $T$-frame context.}

\custompar{Frame Pair Data Preparation} Starting with a frame pair $\mathbf{v} = (v_1,v_2$), we employ $K$ captioning models to generate an initial set of caption pairs $\{(\hat c_1, \hat c_2)\}_1^K$. Acknowledging the potential inaccuracies in these captions, as per observation II, we design two automatic evaluation tasks to assess caption quality. The first task, \emph{progression detection}, examines progress awareness: it checks whether the captions appropriately reflect visual changes between $v_1$ and $v_2$. Specifically, an LLM assesses each caption pair $(\hat c_1, \hat c_2)_k$ to determine if they suggest a visible physical change. We utilize majority voting across multiple LLMs' assessments for all $K$ caption pairs to establish a consensus visual-change label. Caption pairs that align with this consensus are marked as passing; others are marked as failing.

For pairs passing progression detection, we proceed to our second evaluation task—\emph{caption matching}—to assess how precisely $\hat c_1$ and $\hat c_2$ describe $v_1$ and $v_2$, respectively. The task is designed as a multi-choice question format, where a VLM is given $\hat c_1$, $\hat c_2$, and an ``unsure'' option, and tasked with matching the correct caption to each frame. A caption pair is considered good if the evaluation VLM correctly identifies $\hat c_1$ for $v_1$ and $\hat c_2$ for $v_2$. \KG{Because the captions will all be topically related, this is essentially a matching task with ``hard negatives" that lets us automatically gauge the precision of the proposed captions for the target images.}

This automatic pipeline distinguishes between high-quality caption pairs, denoted by $\mathbf{\hat c}^+=(\hat c_1^+, \hat c_2^+)$, and those that exhibit inaccuracies or hallucinations, denoted by $\mathbf{\hat c}^-=(\hat c_1^-, \hat c_2^-)$, forming training data for Stage I.\footnote{\SX{We encourage readers to view data examples provided in Supp. for a better understanding of our data refinement process, as well as %additional 
details on pseudo labeling, prompts used, and the $K$ VLMs we employ.}}% we employ, are also discussed in Supp.}}}

\custompar{\KG{Stage I Training}} Following the success of versatile VLMs in captioning tasks~\cite{xu2024pllava,llava,wang2024tarsier,li2024wolf}, we initialize ProgressCaptioner with the LLAVA-OV-7B~\cite{llava-ov} checkpoint to inherit its pretrained capabilities. Stage-I training utilizes frame and caption pair data \KG{collected on HowToChange and COIN YouTube videos} \textless $\mathbf{v}, \mathbf{\hat c}^+, \mathbf{\hat c}^-$\textgreater~ through two principal methods: supervised fine-tuning (SFT) and direct preference optimization (DPO).
The SFT process is straightforward given our dataset; we perform instructional tuning to tailor the general capabilities of the original VLM to our specific frame-wise captioning requirements using \textless $\mathbf{v}, \mathbf{\hat c}^+$\textgreater. 
The subsequent DPO step targets the prevalent issue of hallucination in VLMs and is innovatively driven by our proposed automatic evaluation critics. Preference optimization~\cite{rafailov2024direct} in LLM training typically requires human-provided preference data to steer LLM responses towards more desirable outputs. Here, we employ progression detection and caption matching to automatically construct preference data $\mathbf{\hat c}^+$ and $\mathbf{\hat c}^-$, eliminating the reliance on manual labeling. This preference data \textless $\mathbf{v}, \mathbf{\hat c}^+, \mathbf{\hat c}^-$\textgreater~ is adopted in DPO training to further enhance model performance with feedback from LLM and VLM evaluations.
% Preference optimization~\cite{rafailov2024direct}, an effective tool in LLM training, typically relies on human-generated preference data to guide LLM towards hallucination-free outputs. Here, we eliminate the need for human labeling and instead employ progression detection and caption matching evaluation to serve as critiques. Specifically, from the pool of candidate captions $\{(\hat c_1, \hat c_2)\}_{1}^K$, we construct preference data pairs where one caption pair that meets both evaluative criteria is preferred, and another that does not is marked as rejected. In this way, the evaluation LLM and VLM are positioned in the model development loop to assess and rate generated captions, facilitating an automatic preference optimization process.
% to further enhance captioning model performance with LLM and VLM feedback. 
\custompar{Frame Sequence Data Preparation} The second stage expands our pseudo labeling scheme from 2 to $T$ frames, where our Stage-I ProgressCaptioner first generates captions using a two-frame sliding window. To increase data diversity and volume, we also incorporate captions produced by other VLMs, with both two-frame and full $T$-frame contexts, since captions of low-quality are also useful (after undergoing our evaluation tasks, those that are rejected enrich the preference data). Once the initial set of caption sequences is generated, we conduct progression detection to identify $M$ visually distinct frames from the original $T$-frame sequence, denoted as $\mathcal V_M = \{v_i\}_{i=1}^M$, using majority voting; $M$ varies based on the distinctiveness of each frame sequence's content. The caption matching task is then employed to encompass $M$ frames, with a selection pool of all $M$ captions, $\mathcal{\hat C}_M = \{\hat c_i\}_{i=1}^M$, plus an ``unsure'' option. A high-quality caption sequence $\mathcal{\hat C^+}$ is identified when the evaluation VLM correctly selects $\hat c_i$ for $v_i$ across all frames. Conversely, a caption sequence is deemed problematic, $\mathcal{\hat C^-}$, if the VLM incorrectly answers more than half of the caption selections. This process forms our Stage-II data.
%\KGnote{this seems at odds with our claim that the existing ones don't keep coherence.} \SXnote{added clarification}

\custompar{\KG{Stage II Training}}
% Having successfully developed the Stage-I ProgressCaptioner, which excels at frame pair captioning, we now proceed to Stage II. This next stage expands our pseudo labeling scheme from 2 to $T$ frames and integrates the stage-I model alongside other VLMs to generate an initial set of caption sequences. Specifically, we apply our ProgressCaptioner in a sliding window fashion across two-frame segments to generate precise and detailed per-frame descriptions. Simultaneously, other VLMs operate within a $T$-frame context window to provide a diverse range of captions that maintain coherence across the sequence. Once the initial set of caption sequences is generated, we conduct progression detection to remove visually similar frames from the sequence, then proceed to multi-frame caption matching on the refined sequence. Frames and their corresponding captions that successfully pass this check are selected as training data for SFT in the second stage. \KGnote{notate?} Additionally, for each frame sequence, a pair of caption sequences—one that passes and another that fails the checks—forms the basis for the second-stage DPO training. This process ensures the collection of high-quality, temporally fine-grained frame and caption sequences that are well-aligned.
% In Stage II, ProgressCaptioner is trained with SFT using data prepared during both stages to refine the model’s understanding of frame-wise dynamics. We then apply DPO, utilizing preference data accumulated from both stages, to further hone the model’s accuracy and mitigate hallucinatory errors. 
Following the same pipeline as stage I, ProgressCaptioner is first trained through SFT using data prepared during both stages, which includes frame-caption pairs \textless $\mathbf{v}, \mathbf{\hat c}^+$\textgreater~and frame-caption sequences \textless $\mathcal{V}, \mathcal{\hat C}^+$\textgreater. Subsequently, we conduct DPO with preference data collected from both stages \textless $\mathbf{v}, \mathbf{\hat c}^+, \mathbf{\hat c}^-$\textgreater~and \textless $\mathcal{V}, \mathcal{\hat C}^+, \mathcal{\hat C}^-$\textgreater~ to further refine performance and mitigate hallucination. \CR{This sequential approach results in our final captioning model, that accepts inputs ranging from 2 to $T$ frames. This flexibility allows users to control the temporal context, balancing the need for local frame-wise changes (smaller window) and global event progressions (larger window).}
The framework is illustrated in Figure~\ref{fig:method}.

% data statistics, filter pass rate
% implementation
\section{Experiments} \label{sec:exp}
\begin{table}[!t]
\tablestyle{3pt}{1.1}
\begin{tabular}{lc|cc|cc|cc}
\thickhline
\multirow{2}{*}{Model} & \multirow{2}{*}{Size} & \multicolumn{2}{c|}{HTC} & \multicolumn{2}{c|}{COIN} & \multicolumn{2}{c}{Penn\&K} \\
 & & Cap & Prog & Cap & Prog & Cap & Prog\\
\hline
\emph{Proprietary models} & \\
\hline
Gemini-1.5-Pro~\cite{reid2024gemini} (img) & - & 28.4 & 59.7 & \tbest{24.3} & 58.6 & 15.3 & 51.2 \\
Gemini-1.5-Pro~\cite{reid2024gemini} & - & \tbest{31.4} & 63.8 & \sbest{25.0} & \tbest{63.8} & 17.6 & \tbest{60.3}\\
GPT-4o~\cite{achiam2023gpt} & - & \sbest{32.4} & \tbest{64.2} & 21.3 & 58.4 & \tbest{18.2} & \sbest{63.2}\\
\hline
\emph{Open-source models} & \\
\hline
Idefics2~\cite{idefics2} & 8B & 2.0 & 54.4 & 2.9 & 52.2 & 12.5 & 50.9\\
VILA~\cite{lin2024vila} & 8B & 6.9 & 53.6 & 5.1 & 48.2 & 15.9 & 51.4\\
Qwen2-VL~\cite{qwen2} & 7B & 13.7 & \sbest{69.6} & 11.0 & \best{70.8} & 8.5 & 58.8\\
LLAVA-Video~\cite{llava-video} & 7B & 3.9 & 59.3 & 8.8 & 53.0 & 9.7 & 51.8\\
LLAVA-OV~\cite{llava-ov} (img) & 7B & 5.9 & 56.3 & 17.6 & 55.4 & 11.9 & 55.5\\
LLAVA-OV~\cite{llava-ov} & 7B & 7.8 & 59.0 & 5.9 & 57.3 & 5.1 & 50.8\\
\rowcolor{LighterGray} 
PL (VLM ensemble) & - & 18.6 & 62.5 & 17.6 & 60.1 & \sbest{19.3} & 52.4\\
% ProgressCaptioner (ours) & 7B & \best{32.4} & \best{70.8} & \best{27.9} & \sbest{66.0} & \best{25.6} & \best{68.2}\\
% just got a better trained model version
ProgressCaptioner (ours) & 7B & \best{37.3} & \best{73.6} & \best{32.3} & \sbest{66.1} & \best{31.3} & \best{63.7}\\
\thickhline
\end{tabular}
\caption{Results on the FrameCapEval Benchmark, \KG{composed of video from four public datasets}. Cap and Prog denote caption matching and progression detection accuracy, respectively. \CR{PL denotes the pseudo labeling baseline adopting filtered captions from multiple VLMs.} ProgressCaptioner greatly outperforms SOTA open-source VLMs and \KG{even} %achieves comparable or better performance than 
the leading proprietary models, despite being a 7B model. The \best{best} results are bolded and underlined, the \sbest{second best} are bolded, and the \emph{third best} are italicized. %Moreover, 
The results confirm our model's generalizability from in-domain datasets (HTC for HowToChange and COIN) to external datasets not seen during training (Penn\&K for Penn Action and Kinetics). 
%\KGnote{in text, how to explain the anomaly of qwen2 being so high on COIN Prog?} \SXnote{Qwen is generally good at progression detection task (also high Prog on HTC), add clarification in text. Just got a new model result and updated the table (higher caption matching accuracy), but we do not have time for another user-study :(}
}
\label{tab:result}
\end{table}

\begin{figure*}[!t]
  \centering
   \includegraphics[width=0.95\linewidth]{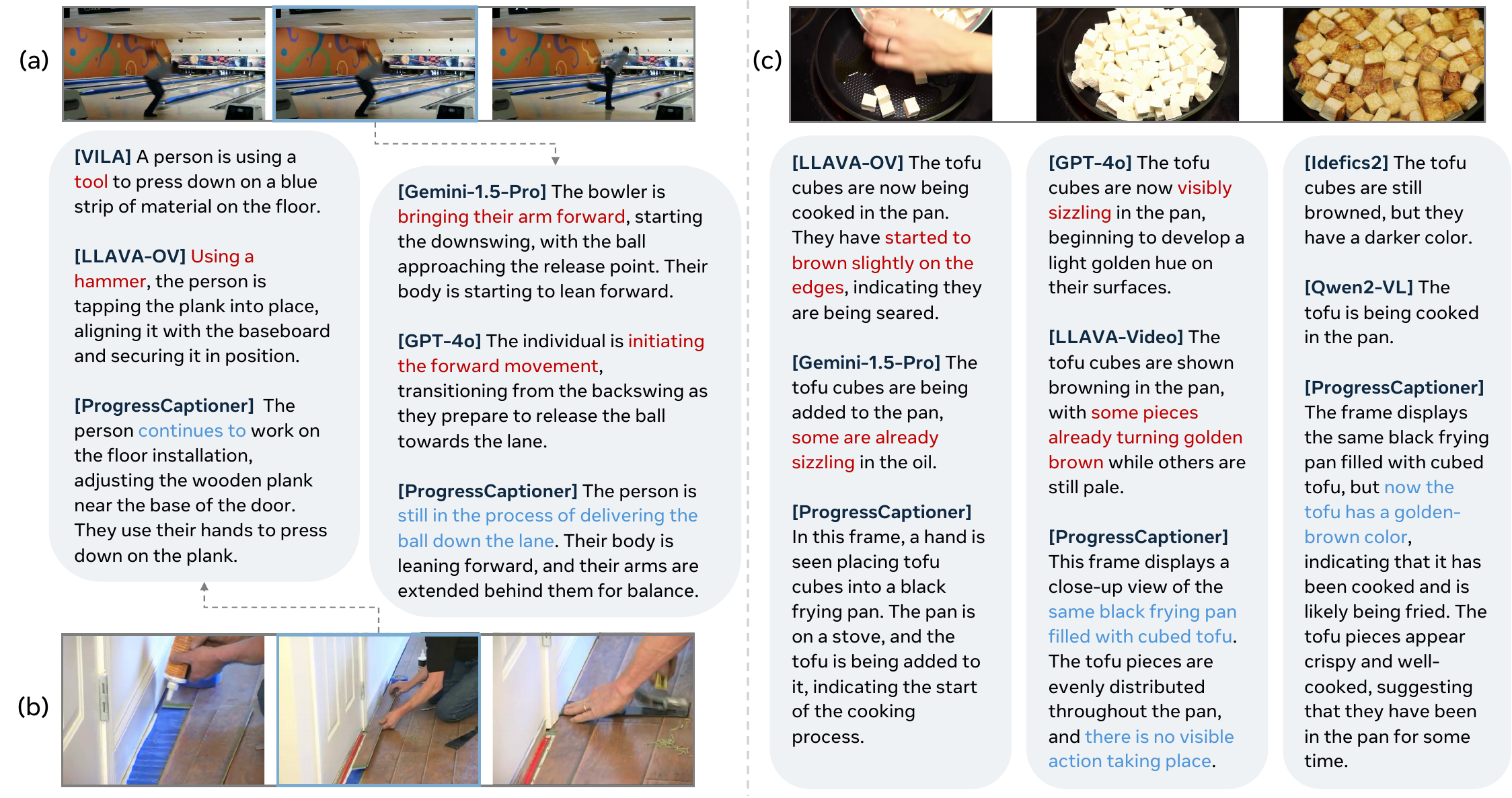}
   \vspace*{-2mm}
   \caption{Qualitative comparisons of ProgressCaptioner with SOTA VLMs on three action sequences. For sequences (a) and (b), only the middle frame predictions are displayed. %Due to space constraints, a selection of frames and model predictions are displayed; 
   See Supp. for all models' predictions on full sequences and more examples.  Inaccuracies in descriptions are highlighted in red.
   % Notably, even top VLMs often produce descriptions that do not align accurately with the corresponding frames, exhibiting temporal mismatch or hallucination. In contrast, ProgressCaptioner consistently delivers hallucination-free and temporally fine-grained captions of high quality, including phrases explicitly calling out progress (blue). More in Supp. 
   Even top VLMs often produce descriptions that misalign with the corresponding frames, while ProgressCaptioner delivers hallucination-free and temporally fine-grained captions, including phrases explicitly calling out progress (blue). 
   % See Supp. for more qualitatives.
   %\KGnote{h1.ow about adding some text color (like blue) on phrases within our outputs that show \emph{action progression}?  like ``same black frying pan", ``but now the tofu has a golden brown color", ``background remains consistent", ``still in the process of delivering" etc.  So we call out those positive properties that we have said are important in the setup, not just generic accuracy.} \SXnote{added} \KGnote{2. decide if we would rather have 2 rows of image examples with just 1 baseline captioner shown?} \SXnote{I feel it's better to show different captioners' performance? helps demonstrate that they all have inaccuracies/ hallucination is prevalent. Not sure if we have space to put both examples? exceed 1 page now}
   % \KGnote{here and figure 1 should we make it blue rather than green? otherwise it reads like red is wrong and green is the only place that's right, whereas we are right more than that.  if so update both captions too.}
   }
   \label{fig:qualitative}
\end{figure*}

\begin{figure}[!t]
  \centering
   \includegraphics[width=0.8\linewidth]{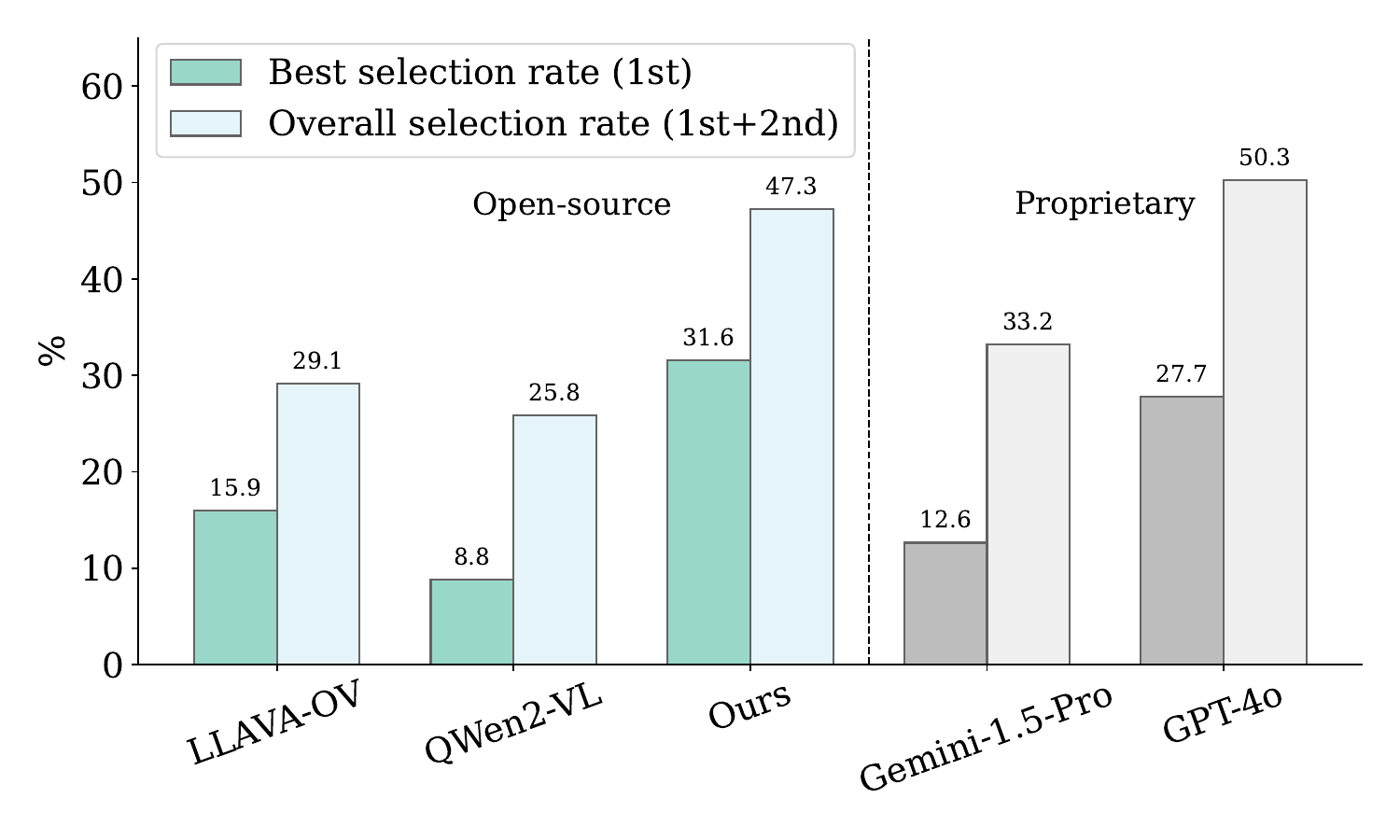}
   \vspace*{-5mm}
   \caption{User study results comparing ProgressCaptioner with top competitors show it as 
   % \cc{We report the best and %overall 
   % top-2 selection rate (\%) for each model, representing the percentage of times participants select captions from each model as the best, and cumulatively as the best or second best.} 
   %\KGnote{skip the top-3 part? and revise caption.} 
   %\SXnote{updated it top-2 so that the difference with gpt-4o looks smaller}
   % ProgressCaptioner is
   the most preferred model (see text). %, with the highest best selection rate and strong overall rankings, closely competing with GPT-4o. %\KGnote{how about using gray shading or something on all the closed models, here and in Table 1? to draw clear visual difference.  and/or sort them with a vertical line in between the open and closed bars on this chart like is done on the table, with a label.} \SXnote{updated}
   }
   \label{fig:user_study}
\end{figure}

\begin{figure*}[!t]
  \centering
   \includegraphics[width=0.75\linewidth]{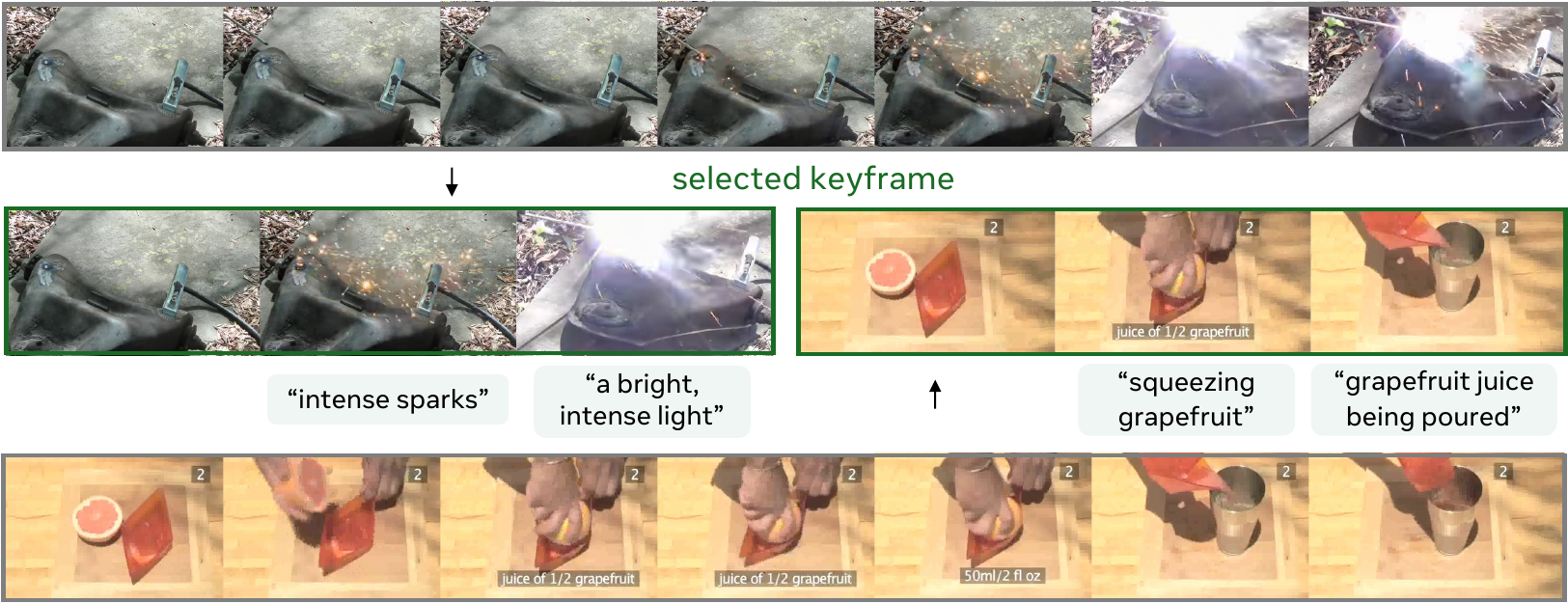}
   \caption{
   \SX{ProgressCaptioner facilitates keyframe selection and enriches the selected keyframes with progress-aware descriptions. 
   % \cc{, effectively condensing the original frame sequences, while enriching the selected frames with descriptions that precisely characterize the action process.}
   }
   % Left: ProgressCaptioner facilitates the selection of representative frames that clearly depict action progression from densely sampled frame sequences, while effectively removing duplicates and enhancing the selected frames with captions. 
   % \cc{See Supp. for more examples. } 
   % \KGnote{I really wish these could be bigger, they are hard to see even zooming, yet this is a cool application.}
   % Right: The detailed per-frame captions produced by ProgressCaptioner allow for zero-shot frame-wise classification when integrated with an LLM. 
   %\KGnote{1. rather than use a chunk of space to explain how the frame selection is done, consider showing two examples' outputs of selected frames; I think the how will be clear from the text. second one not cooking or at least very visually different.  caption could say ``Here our model reduced the original X frames of video into 4 essential keyframes."  2. consier moving the righthand visual to separate figure or combined with the tables that go with it.} \SXnote{updated}
   }
   \vspace*{-4mm}
   \label{fig:keyframe}
\end{figure*}

% In this section, key 
We tackle two questions below: (1) How to evaluate frame-wise caption quality of existing models? And how does ProgressCaptioner perform? (Sec.~\ref{sec:exp_benchmark}); (2) What applications are enabled by precise progress-aware captions? (Sec.~\ref{sec:exp_app})

\subsection{Benchmarking Video Frame Captioning} \label{sec:exp_benchmark}
% To systematically evaluate the quality of generated video frame-wise captions, we propose the FrameCapEval benchmark. \SX{Our benchmark sets a new standard for temporal precision in video captioning, assessing captions based on the criteria established in Sec.~\ref{sec:method_task}.} 

% To systematically evaluate frame-wise video captions, we introduce the FrameCapEval benchmark, establishing a new standard for temporal precision.

\custompar{Benchmark Data Curation} 
We establish the FrameCapEval benchmark, featuring videos from \textbf{four action understanding datasets}: HowToChange~\cite{xue2024learning} and COIN~\cite{tang2019coin} (on which ProgressCaptioner was trained), along with Penn Action~\cite{zhang2013actemes} and Kinetics~\cite{carreira2017quo}, which are unseen in training and serve to assess %the model’s 
generalization capabilities. \SX{We are mindful of the single frame bias~\cite{lei2022revealing} and manually verify all videos to exclude sequences lacking fine-grained action progression.}
% \cc{We are mindful of the single frame bias~\cite{lei2022revealing}—%a recognized issue in video understanding 
% where some actions %are not distinctly temporal and 
% can be adequately depicted with a single frame. To address this, we conduct a manual verification of all videos to eliminate frame sequences that lack fine-grained action progression, as these scenarios are straightforward and can be adequately managed by image captioning models.} 
This process yields a final set of 684 videos. 
% \cc{See Supp. for more details.}
% action-centric frame sequences that exhibit progression within a single action, , making the FrameCapEval benchmark a unique testbed to evaluate a captioning model’s temporally fine-grained description capability.

\custompar{Evaluation Metrics} We employ the automatic evaluation tasks of progression detection and caption matching (Sec.~\ref{sec:method_model}), reporting accuracy with Llama-3.1-70B-Instruct~\cite{dubey2024llama} as the evaluation LLM and Gemini-1.5-Pro~\cite{reid2024gemini} as the evaluation VLM. 
Additionally, we enhance our evaluation with a \textbf{user study} of 15 participants, reporting the average selection rate. \CR{See Supp. for experiments on evaluation metric reliability and full user study details.}

\custompar{Baselines} 
We evaluate an array of state-of-the-art VLMs, including two proprietary models, GPT-4o~\cite{achiam2023gpt} and Gemini-1.5-Pro~\cite{reid2024gemini}, and five open-source models—Idefics2~\cite{idefics2}, VILA~\cite{lin2024vila}, Qwen2-VL~\cite{qwen2}, LLAVA-Video~\cite{llava-video}, and LLAVA-OV~\cite{llava-ov}. 
\CR{We also include a pseudo labeling baseline using filtered captions produced by an ensemble of VLMs (Sec.~\ref{sec:method})}, and image captioning baselines using Gemini-1.5-Pro and LLAVA-OV. We select open-source VLM variants with fewer than 10B parameters for computational efficiency and a fair comparison with our ProgressCaptioner, which is an 7B model. \KG{The closed proprietary models are much larger and trained with much more extensive data; we include them as a useful reference point, but stress that they do not constitute an apples-to-apples comparison, to the disadvantage of our ProgressCaptioner.}
% , all of which process the entire $T$-frame sequences as input
% based on preliminary findings that larger models are more prone to language bias (see Supp. for more details)
%\KGnote{when you run the baselines, it is always on 2-frame chunks or something else? we talk about 2-frame and T-frame training for our approach but don't come back to it.} \SXnote{It's T-frame inference always, both for ours and all baselines since our final model can take $T$ frames as input; T varies from 2 to 6 for test videos, as $>$6 frames due to context length limitation, VLMs out of memory or per-frame caption quality very low, explained this in Supp.}

\custompar{Implementation} 
ProgressCaptioner is constructed with SigLIP~\cite{zhai2023sigmoid} as the vision encoder and Qwen2~\cite{qwen2} as the language model, linked through a projector, and initialized from the LLAVA-OV-7B checkpoint~\cite{llava-ov}. \CR{For benchmark evaluations in Sec.~\ref{sec:exp_benchmark}, ProgressCaptioner operates on the full $T$-frame sequence for a comprehensive temporal context. For applications presented in Sec.~\ref{sec:exp_app}, where fine-grained analysis of local frame changes is crucial, a sliding window approach is used, with the model processing frame pairs. To ensure a fair comparison, all video baseline models are provided with the same temporal context as ProgressCaptioner across all evaluations.}

% For final-stage model training, we finetune ProgressCaptioner on 274.5K data samples over 1 epoch during the SFT stage and on 47.3K preference data for 1 epoch during the DPO stage. 
% \cc{See Supp. for full implementation details.}
% The learning rates are set at 1e-5 for the LLM and projector, and 2e-6 for the vision encoder, , with a batch size of 64
% with a learning rate of 5e-7 and a batch size of 8 across 1 epoch.  We set the preference scaling parameter $\alpha$ = 1.0 and the temperature parameter $\beta$ = 0.2.

\custompar{Results} As shown in Table \ref{tab:result}, on the FrameCapEval benchmark, ProgressCaptioner greatly outperforms existing open-source VLMs \KG{of similar capacity} and \KG{even} 
% demonstrates comparable performance with 
the \KG{(much larger)} latest Gemini-1.5-Pro and GPT-4o.  
%\KGnote{we don't know their actual sizes right?} \SXnote{no, but definitely $>$100B} 
We observe that strong language-backed VLMs like GPT-4o %, with its strong language modeling capabilities, 
show high caption matching accuracy,  %due to their ability to produce more distinguishable captions across different frames. Conversely, 
whereas Qwen2-VL excels in progression detection, %particularly adept at describing frames without progression, demonstrating its strength in 
reducing hallucination. However, it tends to produce less detailed captions, leading to lower caption matching accuracy. 
% ProgressCaptioner demonstrates superior performance in frame-wise video captioning.
\SX{In contrast, ProgressCaptioner effectively balances precision and detail in frame-wise captioning, consistently leading the benchmark across both in-domain and out-of-domain datasets.}
% .
% It excels at the frame-wise video captioning task by delivering: (1) temporally fine-grained descriptions that effectively distinguish between different frames, leading to superior caption matching accuracy, and (2) progress-aware captions that accurately capture visual progression where present and minimize hallucinations where absent, yielding high progression detection accuracy. Despite being a 7B model, ProgressCaptioner demonstrates strong capabilities across both in-domain and out-of-domain datasets, spanning a broad spectrum of activities, from object state changes to various physical activities, underscoring its broad applicability. 

%Moreover, 
Figure~\ref{fig:qualitative} provides qualitative comparisons on three action sequences. Consider the (a) bowling sequence for instance: baseline models erroneously suggest progression in frame 2, like ``arm forward'', exemplifying the common issue of temporal hallucination in current VLMs. This issue recurs in the other two sequences. Conversely, ProgressCaptioner delivers high-quality captions that precisely characterize action \KG{progress} in each frame. 
% In the bowling sequence on the left, there is no actual progression between frames 2 and 3. While baseline models accurately describe frame 2, they erroneously suggest progression in frame 3, such as a ball release. This highlights the prevalent \KG{temporal} hallucination issue with current VLMs. A similar issue is observed in the browning tofu sequence (right). In contrast, ProgressCaptioner excels by delivering captions that are accurate, detailed, and precisely characterize the action \KG{progress} in each frame. 
See Supp. for more qualitative examples and an ablation of ProgressCaptioner.

\custompar{User Study} Figure~\ref{fig:user_study} presents the user study results, where ProgressCaptioner is compared against four of the strongest competitors: LLAVA-OV, Qwen2-VL, Gemini-1.5-Pro and GPT-4o. Each participant is presented with five captions produced by these models and is tasked with
%ranking them from best to third best, \KGnote{is there a precursor study done the same way, so we can say ``following~\cite{xxx}" here?} \SXnote{not exactly the same, but ~\cite{nguyen2024oscar} asks users to select 2 out of 7 captions} 
selecting the top-2, with an additional ``none'' option if the captions are deemed inadequate. ProgressCaptioner emerges as the most preferred model, with an average best caption selection rate of 31.6\%---\KG{2$\times$ to 3.6$\times$ better than the comparably sized best models from the literature~\cite{qwen2,llava-ov}, and even surpassing top-tier proprietary models that enjoy significant scaling advantages.} 
\KG{While our model outperforms all open-source \emph{and} proprietary models for top-1 preference, the more forgiving top-2 metric brings the proprietary closed models back in the game, though our model remains competitive even there (50.3\% for GPT-4o vs.~47.3\% for ours).}
%When considering the overall selection rate %s—choices as the best, second, or third best—
%\KG{among the top 3}, ProgressCaptioner secures a prominent second position, \KG{exceeding Gemini-1.5-Pro and
%rivaling %top-tier proprietary models such as 
%GPT-4o---both top-tier proprietary models with significant scaling advantages.} %\KGnote{does the top-3 bit really add anything here?  why don't we just stick with the top-1 win rate (redraw chart)?} \SXnote{top-3 provides an understanding of how often these models produce ``correct'' captions. Also without top-3, gemini is worse than LLAVA-OV, makes me feel top-1 is somewhat biased, reviewers may wonder the low performance of gemini, as contrast to what gets demonstrated in Table1. I update it to be top-2 ours vs gpt 47.3 vs 50.3 for top 3 it's 58.8 vs 65.1, let me know which looks better}
%and Gemini-1.5-Pro. 
These findings underscore our model's strong ability to produce accurate, temporally fine-grained captions. 
% \cc{See Supp.~for results breakdown.} 

\subsection{Applications of Video Frame Captioning} \label{sec:exp_app}

% \KGnote{I expanded the paragraph headers, do you like it?}

ProgressCaptioner %brings
\KG{offers} progress-aware frame-wise captions, which hold great potential for many real-world applications. We explore several practical use-cases below.
%\KGnote{overall you could get more space back by saying less in section 4.2.} \SXnote{should the header be advancing video understanding first then followed by zero-shot frame classfication and Video QA?}

\custompar{Keyframe Selection} Our first use-case leverages ProgressCaptioner's temporally precise captions as an intermediate representation to identify keyframes within a densely sampled sequence, aided by an LLM (see Supp.~for details).  %; this process pinpoints keyframes that vividly depict the progression of an action with the assistance of an LLM (see Supp. for details on the selection mechanism). 
\SX{Figure~\ref{fig:keyframe} provides two examples, showcasing how ProgressCaptioner's produced captions allow selecting distinct frames that effectively capture different stages of the welding and squeezing grapefruit action.} 
% \KGnote{update, new ex. there now}
While recent video summarization work~\cite{hua2024v2xum} explores using VLMs and LLMs for keyframe selection, it aims to
identify coarse-grained events within long videos, which is not adequate for our problem scenario.  %, which necessitates fine-grained temporal description capability. 
See Supp. for a side-to-side comparison and more qualitatives that underscore this distinction.
% Figure~\ref{} also compares the latest summarization pipeline proposed in V2Xum~\ref{} with ours. As can be seen, when the frame sequences exhibit many redundancies (left example), they fail to identify meaningful subsets (include many similar frames); when the frames indeed demonstrate fine-grained differences (right example), they fail to include them. 

\begin{figure}[!t]
  \centering
   \includegraphics[width=0.9\linewidth]{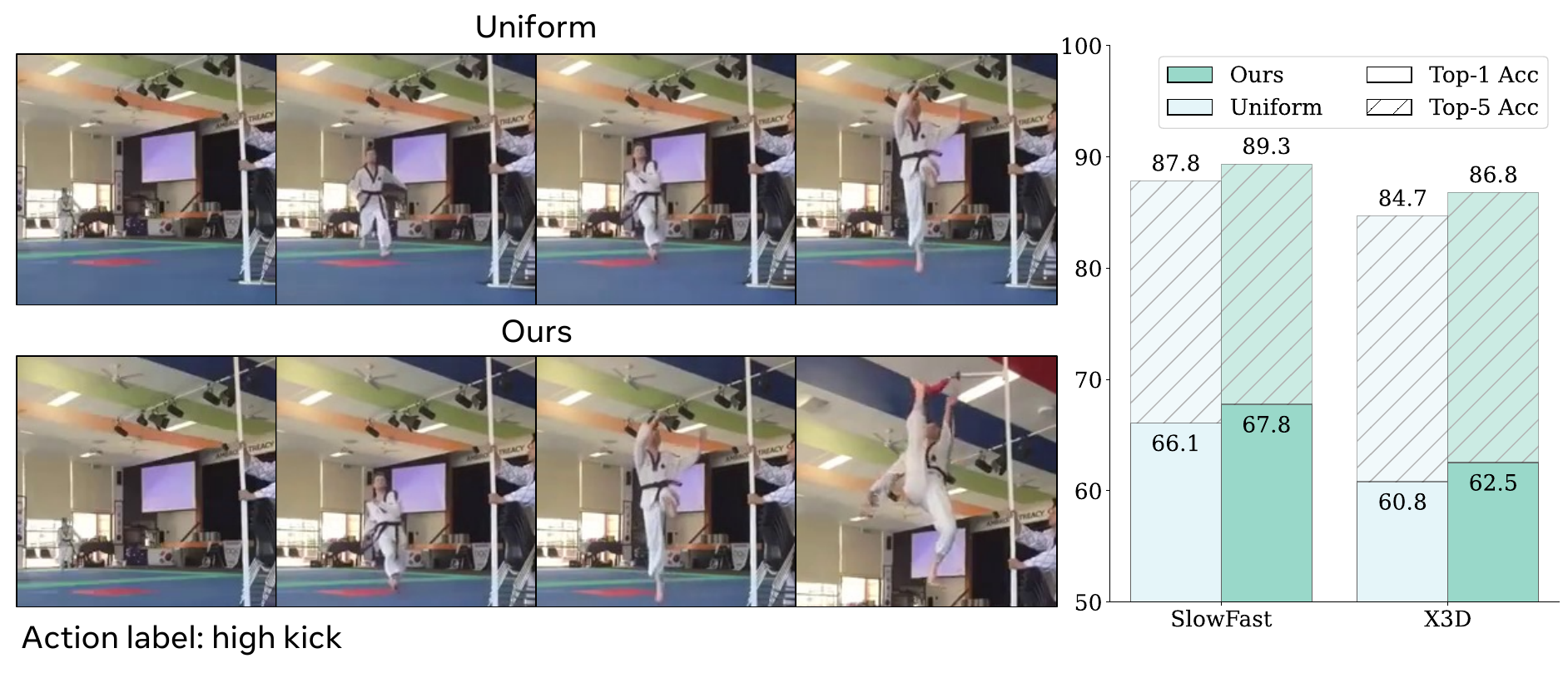}
    \vspace*{-2mm}
   \caption{On Kinetics test videos, ProgressCaptioner selects four frames that are more informative of the action than uniform sampling, resulting in improved action recognition accuracy.}
   \vspace*{-2mm}
   \label{fig:keyframe_k400}
\end{figure}

\custompar{Keyframes for Action Recognition}  \KG{Not only is keyframe selection useful for human viewers to quickly preview a longer video, but it can also extract the most informative portions of a video to benefit activity recognition~\cite{korbar2019scsampler}. To illustrate, we next}
%Furthermore, 
apply our keyframe selection mechanism to Kinetics~\cite{carreira2017quo} Temporal~\cite{sevilla2021only} subset that necessitates multi-frame reasoning.
% test videos from Kinetics~\cite{carreira2017quo}, specifically the Temporal subset~\cite{sevilla2021only} that requires multi-frame reasoning beyond single-frame analysis. 
Given that the original video clips are short (sampled at 1FPS, resulting in sequences of 10 frames), we employ two models that take four frames as input: Slow backbone from SlowFast~\cite{feichtenhofer2019slowfast} and X3D-XS~\cite{feichtenhofer2020x3d}. We prompt GPT-4o to select four representative frames from frame-wise captions produced by ProgressCaptioner. We take the two model checkpoints that have been trained on the Kinetics training set and replace uniformly sampled frames with our selected keyframes during inference.  %\KGnote{so they are trained on all 10 frames? but only downselect during testing?} \SXnote{the action recognition models takes 4 frames as input, both during training and inference}
Figure~\ref{fig:keyframe_k400} presents a qualitative comparison, highlighting performance gains such as a +1.7\% increase in top-1 accuracy for both SlowFast and X3D models. \KG{Even among just 10 candidate frames, our method's fine-grained ability to identify the 4 most informative ones translates into better recognition.}

\custompar{\SX{Advancing Video Understanding}}
% \custompar{Zero-shot Frame Classification}
The precise, frame-wise captions generated by ProgressCaptioner enhance our understanding of videos. To demonstrate this, we consider two video tasks that demand temporally fine-grained understanding: (1) frame-wise classification on HowToChange~\cite{xue2024learning} and Penn Action~\cite{zhang2013actemes},
and (2) video question answering (QA) on NExT-QA~\cite{xiao2021next} (ATP-Hard~\cite{buch2022revisiting}). These tasks are chosen because they challenge the model to comprehend not just the overarching content of a video, but also the more fine-grained event progression within a video. 
% Specifically, the HowToChange and Penn Action test set provide frame-wise labels that detail stages of an object’s state change or an action phase, respectively. Accurately predicting these labels requires a deep understanding of each frame’s content. Similarly, the NexTQA ATP-Hard set poses temporally challenging questions that demand multi-frame reasoning, such as identifying if one event precedes or follows another, highlighting the need for precise temporal comprehension of the video. 
The HowToChange and Penn Action test sets provide frame-wise labels detailing object state changes or action phases, requiring frame-level understanding. Similarly, NexTQA (ATP-Hard) poses temporally challenging questions that demand multi-frame reasoning, such as determining event order, emphasizing the need for precise temporal comprehension.
For baseline comparisons, we evaluate against the LLAVA-OV-7B~\cite{llava-ov} model, from which ProgressCaptioner is initialized, to highlight the enhancements that our specialized training on FrameCap brings to video understanding tasks. 
\SX{For the first task, as we pioneer a zero-shot, language-guided approach to this traditionally vision-centric problem (details below), no other zero-shot baselines exist.}
% For the first task, no other baselines exist since existing approaches~\cite{xue2024learning,dwibedi2019temporal} rely on pseudo object state labels or require training videos, which does not align with our zero-shot focus. 
For the second task, we compare ProgressCaptioner against two existing zero-shot approaches~\cite{momeni2023verbs,wang2024videoagent}.
% to underscore the benefits of our learning paradigm. 

% \subsubsection{Advancing Video Understanding}
% We posit that temporally accurate and fine-grained captions entail a more precise understanding of the video content. To illustrate this capability of ProgressCaptioner, we consider two video question answering (QA) evaluations.
%\KGnote{while the tasks and datasets below are well motivated, it can leave reviewer wondering why these were ``the ones" to try versus something else.  it doesn't help that the baselines vary in the two tables, which also vary from Fig 8.  can you make a more top-level statement here to help see cohesion?} \SXnote{rewrote section 4.2}

\begin{figure}[!t]
  \centering
   \includegraphics[width=0.9\linewidth]{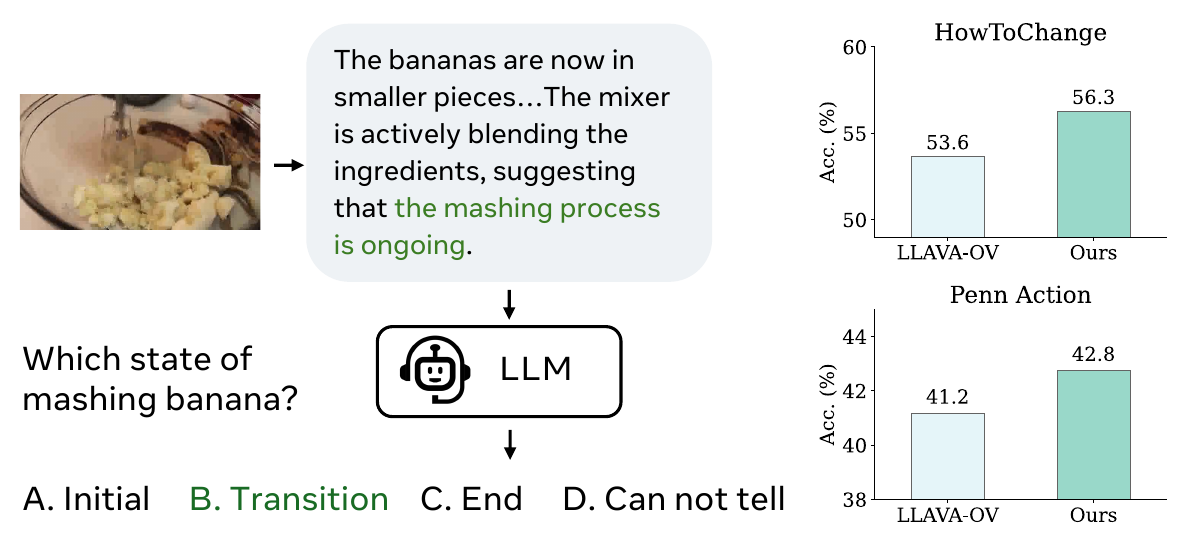}
    \vspace*{-2mm}
   \caption{ProgressCaptioner delivers precise and detailed per-frame descriptions, leading to enhanced zero-shot frame-wise classification performance when compared with LLAVA-OV. }
   \label{fig:frame_cls}
\end{figure}

\custompar{\KGCR{(a)} Zero-shot Frame Classification} We repurpose zero-shot frame-wise classification task into a multi-choice QA format, using frame-wise captions to guide an LLM in identifying the correct label per frame, evaluating caption accuracy and granularity (Figure~\ref{fig:frame_cls} left).
% leveraging frame-wise captions as the basis for the task. An LLM is tasked with identifying the correct label for each frame using only the provided captions, to assess the accuracy and granularity of the captions (as illustrated in Figure~\ref{fig:frame_cls} left). 
Results (Figure~\ref{fig:frame_cls} right) show that ProgressCaptioner consistently outperforms LLAVA-OV across both datasets. Notably, our training involves no signals related to these frame-wise labels, underscoring its generalizability and effectiveness in enhancing video frame-level understanding. %in a zero-shot manner. 

% Next, we evaluate on the NExT-QA~\cite{xiao2021next} ATP-hard~\cite{buch2022revisiting} set, which includes temporally challenging data from the original dataset and is designed to test a model's multi-frame understanding ability and mitigate single-frame bias. For this evaluation, we continue to utilize ProgressCaptioner to generate per-frame descriptions. These descriptions then serve as input for an LLM (we use GPT-4o) to answer the dataset's video-level questions. For comparison, we also include per-frame descriptions generated by LLAVA-OV, with GPT-4o performing the QA, along with the latest reported approaches on this benchmark. 
\custompar{\KGCR{(b)} Video QA} 
Finally, we report results using frame-wise descriptions for video QA, where an LLM (we use GPT-4o) is employed to answer questions on NExT-QA (ATP-Hard) set. As shown in Table~\ref{tab:result_nextqa}, ProgressCaptioner achieves the best results on this benchmark, outperforming the previous leader VideoAgent~\cite{wang2024videoagent} by +3.4\%. Compared with a similar setup using LLAVA-OV, ProgressCaptioner achieves a +4.7\% gain in the temporal subset, highlighting its superior ability to produce fine-grained, temporally precise descriptions and bring enhanced video understanding.
% Moreover, when compared with a similar setup using LLAVA-OV to generate per-frame captions, ProgressCaptioner shows a great performance gain of +4.7\% in the temporal subset. This highlights ProgressCaptioner’s enhanced ability to produce temporally fine-grained descriptions, showcasing its utility in improving temporal video understanding.  %\KGnote{to be clear, can we claim absolute SOTA numbers for this existign benchmark?  was it VideoAgent til now?} \SXnote{yes, it is SOTA on the NextQA ATP set, which includes challenging questions requiring multi-frame reasoning. The full NextQA set has more competitive results, and we didn't test the model (I do not think we will be SOTA there).}

\begin{table}[!t]
\tablestyle{3pt}{1.1}
  \centering
\begin{tabular}{l|ccc}
\thickhline
Model & Acc@C & Acc@T & Acc@All \\
\hline
VFC~\cite{momeni2023verbs} & 32.2 & 30.0 & 31.4 \\
VideoAgent~\cite{wang2024videoagent} & 57.8 & 58.8 & 58.4 \\
LLAVA-OV~\cite{llava-ov} + GPT-4o & 62.6 & 53.4 & 58.8 \\
\rowcolor{LighterGray} 
ProgressCaptioner + GPT-4o (ours) & 64.4 & 58.1 & \textbf{61.8} \\
\thickhline
\end{tabular}
\vspace*{-2mm}
\caption{Video QA results on NExT-QA (ATP-Hard). C and T denote causal and temporal subsets, respectively. 
% \cc{Our approach leads the benchmark, demonstrating that ProgressCaptioner's ability to generate accurate and temporally detailed per-frame captions enhances video QA that demand nuanced multi-frame reasoning.}
}
\label{tab:result_nextqa}
\end{table}

\section{Conclusion} \label{sec:conclusion}
We introduce progress-aware video frame captioning, which necessitates a significant enhancement in current captioning models’ capability to describe temporal action dynamics. 
Towards this end, we develop ProgressCaptioner and show its effectiveness in enhancing the temporal precision and alignment of captions with corresponding frames.
Furthermore, we demonstrate its practical applications: keyframe selection and enhanced video understanding. 
% To address this, we develop ProgressCaptioner and show its effectiveness and practical applications.
By setting a new standard for temporal precision in video captioning, we hope our work inspires further development in this domain. 
% Our work sets a new standard for temporal precision in video captioning, fostering further research in this evolving field.

% \KGCR{\noindent\textbf{Acknowledgements}  
\section*{Acknowledgements} 
Research supported in part by the IFML NSF AI Institute.

{
    \small
    \bibliographystyle{ieeenat_fullname}
    \bibliography{main}

\begin{thebibliography}{87}
\providecommand{\natexlab}[1]{#1}
\providecommand{\url}[1]{\texttt{#1}}
\expandafter\ifx\csname urlstyle\endcsname\relax
  \providecommand{\doi}[1]{doi: #1}\else
  \providecommand{\doi}{doi: \begingroup \urlstyle{rm}\Url}\fi

\bibitem[Abdar et~al.(2023)Abdar, Kollati, Kuraparthi, Pourpanah, McDuff, Ghavamzadeh, Yan, Mohamed, Khosravi, Cambria, et~al.]{abdar2023review}
Moloud Abdar, Meenakshi Kollati, Swaraja Kuraparthi, Farhad Pourpanah, Daniel McDuff, Mohammad Ghavamzadeh, Shuicheng Yan, Abduallah Mohamed, Abbas Khosravi, Erik Cambria, et~al.
\newblock A review of deep learning for video captioning.
\newblock \emph{arXiv preprint arXiv:2304.11431}, 2023.

\bibitem[Achiam et~al.(2023)Achiam, Adler, Agarwal, Ahmad, Akkaya, Aleman, Almeida, Altenschmidt, Altman, Anadkat, et~al.]{achiam2023gpt}
Josh Achiam, Steven Adler, Sandhini Agarwal, Lama Ahmad, Ilge Akkaya, Florencia~Leoni Aleman, Diogo Almeida, Janko Altenschmidt, Sam Altman, Shyamal Anadkat, et~al.
\newblock Gpt-4 technical report.
\newblock \emph{arXiv preprint arXiv:2303.08774}, 2023.

\bibitem[Agrawal et~al.(2019)Agrawal, Desai, Wang, Chen, Jain, Johnson, Batra, Parikh, Lee, and Anderson]{agrawal2019nocaps}
Harsh Agrawal, Karan Desai, Yufei Wang, Xinlei Chen, Rishabh Jain, Mark Johnson, Dhruv Batra, Devi Parikh, Stefan Lee, and Peter Anderson.
\newblock Nocaps: Novel object captioning at scale.
\newblock In \emph{IEEE/CVF International Conference on Computer Vision (ICCV)}, pages 8948--8957, 2019.

\bibitem[Awal et~al.(2024)Awal, Ahmadi, Zhang, and Agrawal]{awal2024vismin}
Rabiul Awal, Saba Ahmadi, Le Zhang, and Aishwarya Agrawal.
\newblock Vismin: Visual minimal-change understanding.
\newblock \emph{arXiv preprint arXiv:2407.16772}, 2024.

\bibitem[Buch et~al.(2022)Buch, Eyzaguirre, Gaidon, Wu, Fei-Fei, and Niebles]{buch2022revisiting}
Shyamal Buch, Crist{\'o}bal Eyzaguirre, Adrien Gaidon, Jiajun Wu, Li Fei-Fei, and Juan~Carlos Niebles.
\newblock Revisiting the ``video" in video-language understanding.
\newblock In \emph{IEEE/CVF Conference on Computer Vision and Pattern Recognition (CVPR)}, pages 2917--2927, 2022.

\bibitem[Cai et~al.(2024)Cai, Tan, Zhang, Zou, Zhang, Yao, Zhu, Gu, Zhong, Shang, et~al.]{cai2024temporalbench}
Mu Cai, Reuben Tan, Jianrui Zhang, Bocheng Zou, Kai Zhang, Feng Yao, Fangrui Zhu, Jing Gu, Yiwu Zhong, Yuzhang Shang, et~al.
\newblock Temporalbench: Benchmarking fine-grained temporal understanding for multimodal video models.
\newblock \emph{arXiv preprint arXiv:2410.10818}, 2024.

\bibitem[Carreira and Zisserman(2017)]{carreira2017quo}
Joao Carreira and Andrew Zisserman.
\newblock Quo vadis, action recognition? a new model and the kinetics dataset.
\newblock In \emph{IEEE/CVF Conference on Computer Vision and Pattern Recognition (CVPR)}, pages 6299--6308, 2017.

\bibitem[Chai et~al.(2024)Chai, Song, Du, Meng, Madhavan, Bar-Tal, Hwang, Xie, and Manning]{chai2024auroracap}
Wenhao Chai, Enxin Song, Yilun Du, Chenlin Meng, Vashisht Madhavan, Omer Bar-Tal, Jeng-Neng Hwang, Saining Xie, and Christopher~D Manning.
\newblock Auroracap: Efficient, performant video detailed captioning and a new benchmark.
\newblock \emph{arXiv preprint arXiv:2410.03051}, 2024.

\bibitem[Chen and Dolan(2011)]{chen2011collecting}
David Chen and William~B Dolan.
\newblock Collecting highly parallel data for paraphrase evaluation.
\newblock In \emph{Proceedings of the 49th annual meeting of the association for computational linguistics: human language technologies}, pages 190--200, 2011.

\bibitem[Chen et~al.(2024{\natexlab{a}})Chen, Liao, Lin, Yu, Chen, and Wang]{chen2024rextime}
Jr-Jen Chen, Yu-Chien Liao, Hsi-Che Lin, Yu-Chu Yu, Yen-Chun Chen, and Yu-Chiang~Frank Wang.
\newblock Rextime: A benchmark suite for reasoning-across-time in videos.
\newblock \emph{arXiv preprint arXiv:2406.19392}, 2024{\natexlab{a}}.

\bibitem[Chen et~al.(2024{\natexlab{b}})Chen, Wei, Li, Dong, Zhang, Zang, Chen, Duan, Lin, Tang, et~al.]{chen2024sharegpt4video}
Lin Chen, Xilin Wei, Jinsong Li, Xiaoyi Dong, Pan Zhang, Yuhang Zang, Zehui Chen, Haodong Duan, Bin Lin, Zhenyu Tang, et~al.
\newblock Sharegpt4video: Improving video understanding and generation with better captions.
\newblock \emph{arXiv preprint arXiv:2406.04325}, 2024{\natexlab{b}}.

\bibitem[Chen et~al.(2024{\natexlab{c}})Chen, Siarohin, Menapace, Deyneka, Chao, Jeon, Fang, Lee, Ren, Yang, et~al.]{chen2024panda}
Tsai-Shien Chen, Aliaksandr Siarohin, Willi Menapace, Ekaterina Deyneka, Hsiang-wei Chao, Byung~Eun Jeon, Yuwei Fang, Hsin-Ying Lee, Jian Ren, Ming-Hsuan Yang, et~al.
\newblock Panda-70m: Captioning 70m videos with multiple cross-modality teachers.
\newblock In \emph{IEEE/CVF Conference on Computer Vision and Pattern Recognition (CVPR)}, pages 13320--13331, 2024{\natexlab{c}}.

\bibitem[Chen et~al.(2015)Chen, Fang, Lin, Vedantam, Gupta, Doll{\'a}r, and Zitnick]{chen2015microsoft}
Xinlei Chen, Hao Fang, Tsung-Yi Lin, Ramakrishna Vedantam, Saurabh Gupta, Piotr Doll{\'a}r, and C~Lawrence Zitnick.
\newblock Microsoft coco captions: Data collection and evaluation server.
\newblock \emph{arXiv preprint arXiv:1504.00325}, 2015.

\bibitem[Dubey et~al.(2024)Dubey, Jauhri, Pandey, Kadian, Al-Dahle, Letman, Mathur, Schelten, Yang, Fan, et~al.]{dubey2024llama}
Abhimanyu Dubey, Abhinav Jauhri, Abhinav Pandey, Abhishek Kadian, Ahmad Al-Dahle, Aiesha Letman, Akhil Mathur, Alan Schelten, Amy Yang, Angela Fan, et~al.
\newblock The llama 3 herd of models.
\newblock \emph{arXiv preprint arXiv:2407.21783}, 2024.

\bibitem[Dunlap et~al.(2024)Dunlap, Zhang, Wang, Zhong, Darrell, Steinhardt, Gonzalez, and Yeung-Levy]{dunlap2024describing}
Lisa Dunlap, Yuhui Zhang, Xiaohan Wang, Ruiqi Zhong, Trevor Darrell, Jacob Steinhardt, Joseph~E Gonzalez, and Serena Yeung-Levy.
\newblock Describing differences in image sets with natural language.
\newblock In \emph{IEEE/CVF Conference on Computer Vision and Pattern Recognition (CVPR)}, pages 24199--24208, 2024.

\bibitem[Feichtenhofer(2020)]{feichtenhofer2020x3d}
Christoph Feichtenhofer.
\newblock X3d: Expanding architectures for efficient video recognition.
\newblock In \emph{IEEE/CVF Conference on Computer Vision and Pattern Recognition (CVPR)}, pages 203--213, 2020.

\bibitem[Feichtenhofer et~al.(2019)Feichtenhofer, Fan, Malik, and He]{feichtenhofer2019slowfast}
Christoph Feichtenhofer, Haoqi Fan, Jitendra Malik, and Kaiming He.
\newblock Slowfast networks for video recognition.
\newblock In \emph{IEEE/CVF International Conference on Computer Vision (ICCV)}, pages 6202--6211, 2019.

\bibitem[Grunde-McLaughlin et~al.(2021)Grunde-McLaughlin, Krishna, and Agrawala]{grunde2021agqa}
Madeleine Grunde-McLaughlin, Ranjay Krishna, and Maneesh Agrawala.
\newblock Agqa: A benchmark for compositional spatio-temporal reasoning.
\newblock In \emph{IEEE/CVF Conference on Computer Vision and Pattern Recognition (CVPR)}, pages 11287--11297, 2021.

\bibitem[Guan et~al.(2024)Guan, Liu, Wu, Xian, Li, Liu, Wang, Chen, Huang, Yacoob, et~al.]{guan2024hallusionbench}
Tianrui Guan, Fuxiao Liu, Xiyang Wu, Ruiqi Xian, Zongxia Li, Xiaoyu Liu, Xijun Wang, Lichang Chen, Furong Huang, Yaser Yacoob, et~al.
\newblock Hallusionbench: an advanced diagnostic suite for entangled language hallucination and visual illusion in large vision-language models.
\newblock In \emph{IEEE/CVF Conference on Computer Vision and Pattern Recognition (CVPR)}, pages 14375--14385, 2024.

\bibitem[Han et~al.(2023{\natexlab{a}})Han, Bain, Nagrani, Varol, Xie, and Zisserman]{han2023autoad}
Tengda Han, Max Bain, Arsha Nagrani, G{\"u}l Varol, Weidi Xie, and Andrew Zisserman.
\newblock Autoad: Movie description in context.
\newblock In \emph{IEEE/CVF Conference on Computer Vision and Pattern Recognition (CVPR)}, pages 18930--18940, 2023{\natexlab{a}}.

\bibitem[Han et~al.(2023{\natexlab{b}})Han, Bain, Nagrani, Varol, Xie, and Zisserman]{han2023autoad2}
Tengda Han, Max Bain, Arsha Nagrani, Gul Varol, Weidi Xie, and Andrew Zisserman.
\newblock Autoad ii: The sequel-who, when, and what in movie audio description.
\newblock In \emph{IEEE/CVF International Conference on Computer Vision (ICCV)}, pages 13645--13655, 2023{\natexlab{b}}.

\bibitem[Han et~al.(2024)Han, Bain, Nagrani, Varol, Xie, and Zisserman]{han2024autoad}
Tengda Han, Max Bain, Arsha Nagrani, G{\"u}l Varol, Weidi Xie, and Andrew Zisserman.
\newblock Autoad iii: The prequel-back to the pixels.
\newblock In \emph{IEEE/CVF Conference on Computer Vision and Pattern Recognition (CVPR)}, pages 18164--18174, 2024.

\bibitem[Hossain et~al.(2019)Hossain, Sohel, Shiratuddin, and Laga]{hossain2019comprehensive}
MD~Zakir Hossain, Ferdous Sohel, Mohd~Fairuz Shiratuddin, and Hamid Laga.
\newblock A comprehensive survey of deep learning for image captioning.
\newblock \emph{ACM Computing Surveys (CsUR)}, 51\penalty0 (6):\penalty0 1--36, 2019.

\bibitem[Hua et~al.(2024)Hua, Tang, Xu, and Luo]{hua2024v2xum}
Hang Hua, Yunlong Tang, Chenliang Xu, and Jiebo Luo.
\newblock V2xum-llm: Cross-modal video summarization with temporal prompt instruction tuning.
\newblock \emph{arXiv preprint arXiv:2404.12353}, 2024.

\bibitem[Huang et~al.(2016)Huang, Ferraro, Mostafazadeh, Misra, Agrawal, Devlin, Girshick, He, Kohli, Batra, et~al.]{huang2016visual}
Ting-Hao Huang, Francis Ferraro, Nasrin Mostafazadeh, Ishan Misra, Aishwarya Agrawal, Jacob Devlin, Ross Girshick, Xiaodong He, Pushmeet Kohli, Dhruv Batra, et~al.
\newblock Visual storytelling.
\newblock In \emph{Proceedings of the 2016 conference of the North American chapter of the association for computational linguistics: Human language technologies}, pages 1233--1239, 2016.

\bibitem[Islam et~al.(2024)Islam, Ho, Yang, Nagarajan, Torresani, and Bertasius]{islam2024video}
Md~Mohaiminul Islam, Ngan Ho, Xitong Yang, Tushar Nagarajan, Lorenzo Torresani, and Gedas Bertasius.
\newblock Video recap: Recursive captioning of hour-long videos.
\newblock In \emph{IEEE/CVF Conference on Computer Vision and Pattern Recognition (CVPR)}, pages 18198--18208, 2024.

\bibitem[Jhamtani and Berg-Kirkpatrick(2018)]{jhamtani2018learning}
Harsh Jhamtani and Taylor Berg-Kirkpatrick.
\newblock Learning to describe differences between pairs of similar images.
\newblock \emph{arXiv preprint arXiv:1808.10584}, 2018.

\bibitem[Jiao et~al.(2024)Jiao, Chen, Huang, Li, and Shen]{jiao2024img}
Qirui Jiao, Daoyuan Chen, Yilun Huang, Yaliang Li, and Ying Shen.
\newblock Img-diff: Contrastive data synthesis for multimodal large language models.
\newblock \emph{arXiv preprint arXiv:2408.04594}, 2024.

\bibitem[Korbar et~al.(2019)Korbar, Tran, and Torresani]{korbar2019scsampler}
Bruno Korbar, Du Tran, and Lorenzo Torresani.
\newblock Scsampler: Sampling salient clips from video for efficient action recognition.
\newblock In \emph{IEEE/CVF International Conference on Computer Vision (ICCV)}, pages 6232--6242, 2019.

\bibitem[Krishna et~al.(2017)Krishna, Hata, Ren, Fei-Fei, and Carlos~Niebles]{krishna2017dense}
Ranjay Krishna, Kenji Hata, Frederic Ren, Li Fei-Fei, and Juan Carlos~Niebles.
\newblock Dense-captioning events in videos.
\newblock In \emph{IEEE/CVF International Conference on Computer Vision (ICCV)}, pages 706--715, 2017.

\bibitem[Lauren{\c{c}}on et~al.(2024)Lauren{\c{c}}on, Tronchon, Cord, and Sanh]{idefics2}
Hugo Lauren{\c{c}}on, L{\'e}o Tronchon, Matthieu Cord, and Victor Sanh.
\newblock What matters when building vision-language models?
\newblock \emph{arXiv preprint arXiv:2405.02246}, 2024.

\bibitem[Lei et~al.(2022)Lei, Berg, and Bansal]{lei2022revealing}
Jie Lei, Tamara~L Berg, and Mohit Bansal.
\newblock Revealing single frame bias for video-and-language learning.
\newblock \emph{arXiv preprint arXiv:2206.03428}, 2022.

\bibitem[Li et~al.(2024{\natexlab{a}})Li, Zhang, Guo, Zhang, Li, Zhang, Zhang, Li, Liu, and Li]{llava-ov}
Bo Li, Yuanhan Zhang, Dong Guo, Renrui Zhang, Feng Li, Hao Zhang, Kaichen Zhang, Yanwei Li, Ziwei Liu, and Chunyuan Li.
\newblock Llava-onevision: Easy visual task transfer.
\newblock \emph{arXiv preprint arXiv:2408.03326}, 2024{\natexlab{a}}.

\bibitem[Li et~al.(2024{\natexlab{b}})Li, Zhu, Tian, Tan, Chen, Lu, Cui, Veer, Ehrlich, Philion, et~al.]{li2024wolf}
Boyi Li, Ligeng Zhu, Ran Tian, Shuhan Tan, Yuxiao Chen, Yao Lu, Yin Cui, Sushant Veer, Max Ehrlich, Jonah Philion, et~al.
\newblock Wolf: Captioning everything with a world summarization framework.
\newblock \emph{arXiv preprint arXiv:2407.18908}, 2024{\natexlab{b}}.

\bibitem[Li et~al.(2024{\natexlab{c}})Li, Zhang, Zhang, Zhang, Li, Li, Ma, and Li]{llava-next}
Feng Li, Renrui Zhang, Hao Zhang, Yuanhan Zhang, Bo Li, Wei Li, Zejun Ma, and Chunyuan Li.
\newblock Llava-next-interleave: Tackling multi-image, video, and 3d in large multimodal models.
\newblock \emph{arXiv preprint arXiv:2407.07895}, 2024{\natexlab{c}}.

\bibitem[Li et~al.(2019{\natexlab{a}})Li, Wong, Zhao, and Kankanhalli]{li2019video}
Junnan Li, Yongkang Wong, Qi Zhao, and Mohan~S Kankanhalli.
\newblock Video storytelling: Textual summaries for events.
\newblock \emph{IEEE Transactions on Multimedia}, 22\penalty0 (2):\penalty0 554--565, 2019{\natexlab{a}}.

\bibitem[Li et~al.(2023{\natexlab{a}})Li, Li, Savarese, and Hoi]{li2023blip}
Junnan Li, Dongxu Li, Silvio Savarese, and Steven Hoi.
\newblock Blip-2: Bootstrapping language-image pre-training with frozen image encoders and large language models.
\newblock In \emph{International Conference on Machine Learning (ICML)}, pages 19730--19742. PMLR, 2023{\natexlab{a}}.

\bibitem[Li et~al.(2019{\natexlab{b}})Li, Tao, Li, and Fu]{li2019visual}
Sheng Li, Zhiqiang Tao, Kang Li, and Yun Fu.
\newblock Visual to text: Survey of image and video captioning.
\newblock \emph{IEEE Transactions on Emerging Topics in Computational Intelligence}, 3\penalty0 (4):\penalty0 297--312, 2019{\natexlab{b}}.

\bibitem[Li et~al.(2023{\natexlab{b}})Li, Li, Ren, Liu, Liu, Gao, Sun, and Hou]{li2023vitatecs}
Shicheng Li, Lei Li, Shuhuai Ren, Yuanxin Liu, Yi Liu, Rundong Gao, Xu Sun, and Lu Hou.
\newblock Vitatecs: A diagnostic dataset for temporal concept understanding of video-language models.
\newblock \emph{arXiv preprint arXiv:2311.17404}, 2023{\natexlab{b}}.

\bibitem[Lin et~al.(2024)Lin, Yin, Ping, Molchanov, Shoeybi, and Han]{lin2024vila}
Ji Lin, Hongxu Yin, Wei Ping, Pavlo Molchanov, Mohammad Shoeybi, and Song Han.
\newblock Vila: On pre-training for visual language models.
\newblock In \emph{IEEE/CVF Conference on Computer Vision and Pattern Recognition (CVPR)}, pages 26689--26699, 2024.

\bibitem[Liu et~al.(2024{\natexlab{a}})Liu, Li, Li, and Lee]{liu2024improved}
Haotian Liu, Chunyuan Li, Yuheng Li, and Yong~Jae Lee.
\newblock Improved baselines with visual instruction tuning.
\newblock In \emph{IEEE/CVF Conference on Computer Vision and Pattern Recognition (CVPR)}, pages 26296--26306, 2024{\natexlab{a}}.

\bibitem[Liu et~al.(2024{\natexlab{b}})Liu, Li, Wu, and Lee]{llava}
Haotian Liu, Chunyuan Li, Qingyang Wu, and Yong~Jae Lee.
\newblock Visual instruction tuning.
\newblock \emph{Advances in Neural Information Processing Systems (NeurIPS)}, 36, 2024{\natexlab{b}}.

\bibitem[Liu et~al.(2024{\natexlab{c}})Liu, Li, Liu, Wang, Ren, Li, Chen, Sun, and Hou]{liu2024tempcompass}
Yuanxin Liu, Shicheng Li, Yi Liu, Yuxiang Wang, Shuhuai Ren, Lei Li, Sishuo Chen, Xu Sun, and Lu Hou.
\newblock Tempcompass: Do video llms really understand videos?
\newblock \emph{arXiv preprint arXiv:2403.00476}, 2024{\natexlab{c}}.

\bibitem[Mangalam et~al.(2023)Mangalam, Akshulakov, and Malik]{mangalam2023egoschema}
Karttikeya Mangalam, Raiymbek Akshulakov, and Jitendra Malik.
\newblock Egoschema: A diagnostic benchmark for very long-form video language understanding.
\newblock \emph{Advances in Neural Information Processing Systems (NeurIPS)}, 36:\penalty0 46212--46244, 2023.

\bibitem[Momeni et~al.(2023)Momeni, Caron, Nagrani, Zisserman, and Schmid]{momeni2023verbs}
Liliane Momeni, Mathilde Caron, Arsha Nagrani, Andrew Zisserman, and Cordelia Schmid.
\newblock Verbs in action: Improving verb understanding in video-language models.
\newblock In \emph{IEEE/CVF International Conference on Computer Vision (ICCV)}, pages 15579--15591, 2023.

\bibitem[Nadeem et~al.(2024)Nadeem, Sardari, Dawes, Husain, Hilton, and Mustafa]{nadeem2024narrativebridge}
Asmar Nadeem, Faegheh Sardari, Robert Dawes, Syed~Sameed Husain, Adrian Hilton, and Armin Mustafa.
\newblock Narrativebridge: Enhancing video captioning with causal-temporal narrative.
\newblock \emph{arXiv preprint arXiv:2406.06499}, 2024.

\bibitem[Nguyen et~al.(2024)Nguyen, Bi, Vosoughi, Tian, Fazli, and Xu]{nguyen2024oscar}
Nguyen Nguyen, Jing Bi, Ali Vosoughi, Yapeng Tian, Pooyan Fazli, and Chenliang Xu.
\newblock Oscar: Object state captioning and state change representation.
\newblock \emph{arXiv preprint arXiv:2402.17128}, 2024.

\bibitem[Park et~al.(2019)Park, Darrell, and Rohrbach]{park2019robust}
Dong~Huk Park, Trevor Darrell, and Anna Rohrbach.
\newblock Robust change captioning.
\newblock In \emph{IEEE/CVF International Conference on Computer Vision (ICCV)}, pages 4624--4633, 2019.

\bibitem[Perrett et~al.(2024)Perrett, Han, Damen, and Zisserman]{perrett2024s}
Toby Perrett, Tengda Han, Dima Damen, and Andrew Zisserman.
\newblock It's just another day: Unique video captioning by discriminative prompting.
\newblock \emph{arXiv preprint arXiv:2410.11702}, 2024.

\bibitem[Polyak et~al.(2024)Polyak, Zohar, Brown, Tjandra, Sinha, Lee, Vyas, Shi, Ma, Chuang, et~al.]{polyak2024movie}
Adam Polyak, Amit Zohar, Andrew Brown, Andros Tjandra, Animesh Sinha, Ann Lee, Apoorv Vyas, Bowen Shi, Chih-Yao Ma, Ching-Yao Chuang, et~al.
\newblock Movie gen: A cast of media foundation models.
\newblock \emph{arXiv preprint arXiv:2410.13720}, 2024.

\bibitem[Radford et~al.(2021)Radford, Kim, Hallacy, Ramesh, Goh, Agarwal, Sastry, Askell, Mishkin, Clark, et~al.]{radford2021learning}
Alec Radford, Jong~Wook Kim, Chris Hallacy, Aditya Ramesh, Gabriel Goh, Sandhini Agarwal, Girish Sastry, Amanda Askell, Pamela Mishkin, Jack Clark, et~al.
\newblock Learning transferable visual models from natural language supervision.
\newblock In \emph{International Conference on Machine Learning (ICML)}, pages 8748--8763. PMLR, 2021.

\bibitem[Rafailov et~al.(2024)Rafailov, Sharma, Mitchell, Manning, Ermon, and Finn]{rafailov2024direct}
Rafael Rafailov, Archit Sharma, Eric Mitchell, Christopher~D Manning, Stefano Ermon, and Chelsea Finn.
\newblock Direct preference optimization: Your language model is secretly a reward model.
\newblock \emph{Advances in Neural Information Processing Systems (NeurIPS)}, 36, 2024.

\bibitem[Reid et~al.(2024)Reid, Savinov, Teplyashin, Lepikhin, Lillicrap, Alayrac, Soricut, Lazaridou, Firat, Schrittwieser, et~al.]{reid2024gemini}
Machel Reid, Nikolay Savinov, Denis Teplyashin, Dmitry Lepikhin, Timothy Lillicrap, Jean-baptiste Alayrac, Radu Soricut, Angeliki Lazaridou, Orhan Firat, Julian Schrittwieser, et~al.
\newblock Gemini 1.5: Unlocking multimodal understanding across millions of tokens of context.
\newblock \emph{arXiv preprint arXiv:2403.05530}, 2024.

\bibitem[Ren et~al.(2024)Ren, Yao, Li, Sun, and Hou]{ren2024timechat}
Shuhuai Ren, Linli Yao, Shicheng Li, Xu Sun, and Lu Hou.
\newblock Timechat: A time-sensitive multimodal large language model for long video understanding.
\newblock In \emph{IEEE/CVF Conference on Computer Vision and Pattern Recognition (CVPR)}, pages 14313--14323, 2024.

\bibitem[Rousseeuw(1987)]{rousseeuw1987silhouettes}
Peter~J Rousseeuw.
\newblock Silhouettes: a graphical aid to the interpretation and validation of cluster analysis.
\newblock \emph{Journal of computational and applied mathematics}, 20:\penalty0 53--65, 1987.

\bibitem[Sachdeva and Zisserman(2023)]{sachdeva2023change}
Ragav Sachdeva and Andrew Zisserman.
\newblock The change you want to see (now in 3d).
\newblock In \emph{IEEE/CVF International Conference on Computer Vision (ICCV)}, pages 2060--2069, 2023.

\bibitem[Sevilla-Lara et~al.(2021)Sevilla-Lara, Zha, Yan, Goswami, Feiszli, and Torresani]{sevilla2021only}
Laura Sevilla-Lara, Shengxin Zha, Zhicheng Yan, Vedanuj Goswami, Matt Feiszli, and Lorenzo Torresani.
\newblock Only time can tell: Discovering temporal data for temporal modeling.
\newblock In \emph{IEEE Winter Conference on Applications of Computer Vision (WACV)}, pages 535--544, 2021.

\bibitem[Singer et~al.(2022)Singer, Polyak, Hayes, Yin, An, Zhang, Hu, Yang, Ashual, Gafni, et~al.]{singer2022make}
Uriel Singer, Adam Polyak, Thomas Hayes, Xi Yin, Jie An, Songyang Zhang, Qiyuan Hu, Harry Yang, Oron Ashual, Oran Gafni, et~al.
\newblock Make-a-video: Text-to-video generation without text-video data.
\newblock \emph{arXiv preprint arXiv:2209.14792}, 2022.

\bibitem[Tang et~al.(2019)Tang, Ding, Rao, Zheng, Zhang, Zhao, Lu, and Zhou]{tang2019coin}
Yansong Tang, Dajun Ding, Yongming Rao, Yu Zheng, Danyang Zhang, Lili Zhao, Jiwen Lu, and Jie Zhou.
\newblock Coin: A large-scale dataset for comprehensive instructional video analysis.
\newblock In \emph{IEEE/CVF Conference on Computer Vision and Pattern Recognition (CVPR)}, pages 1207--1216, 2019.

\bibitem[Vinyals et~al.(2016)Vinyals, Toshev, Bengio, and Erhan]{vinyals2016show}
Oriol Vinyals, Alexander Toshev, Samy Bengio, and Dumitru Erhan.
\newblock Show and tell: Lessons learned from the 2015 mscoco image captioning challenge.
\newblock \emph{IEEE Transactions on Pattern Analysis and Machine Intelligence}, 39\penalty0 (4):\penalty0 652--663, 2016.

\bibitem[Wang et~al.(2024{\natexlab{a}})Wang, Yuan, and Zhang]{wang2024tarsier}
Jiawei Wang, Liping Yuan, and Yuchen Zhang.
\newblock Tarsier: Recipes for training and evaluating large video description models.
\newblock \emph{arXiv preprint arXiv:2407.00634}, 2024{\natexlab{a}}.

\bibitem[Wang et~al.(2024{\natexlab{b}})Wang, Bai, Tan, Wang, Fan, Bai, Chen, Liu, Wang, Ge, Fan, Dang, Du, Ren, Men, Liu, Zhou, Zhou, and Lin]{qwen2}
Peng Wang, Shuai Bai, Sinan Tan, Shijie Wang, Zhihao Fan, Jinze Bai, Keqin Chen, Xuejing Liu, Jialin Wang, Wenbin Ge, Yang Fan, Kai Dang, Mengfei Du, Xuancheng Ren, Rui Men, Dayiheng Liu, Chang Zhou, Jingren Zhou, and Junyang Lin.
\newblock Qwen2-vl: Enhancing vision-language model's perception of the world at any resolution.
\newblock \emph{arXiv preprint arXiv:2409.12191}, 2024{\natexlab{b}}.

\bibitem[Wang et~al.(2019)Wang, Wu, Chen, Li, Wang, and Wang]{wang2019vatex}
Xin Wang, Jiawei Wu, Junkun Chen, Lei Li, Yuan-Fang Wang, and William~Yang Wang.
\newblock Vatex: A large-scale, high-quality multilingual dataset for video-and-language research.
\newblock In \emph{IEEE/CVF International Conference on Computer Vision (ICCV)}, pages 4581--4591, 2019.

\bibitem[Wang et~al.(2024{\natexlab{c}})Wang, Zhang, Zohar, and Yeung-Levy]{wang2024videoagent}
Xiaohan Wang, Yuhui Zhang, Orr Zohar, and Serena Yeung-Levy.
\newblock Videoagent: Long-form video understanding with large language model as agent.
\newblock \emph{arXiv preprint arXiv:2403.10517}, 2024{\natexlab{c}}.

\bibitem[Wang et~al.(2024{\natexlab{d}})Wang, Wang, Zhao, Xie, and Zheng]{wang2024videohallucer}
Yuxuan Wang, Yueqian Wang, Dongyan Zhao, Cihang Xie, and Zilong Zheng.
\newblock Videohallucer: Evaluating intrinsic and extrinsic hallucinations in large video-language models.
\newblock \emph{arXiv preprint arXiv:2406.16338}, 2024{\natexlab{d}}.

\bibitem[Wang et~al.(2023)Wang, Sung, Cheng, Bertasius, and Bansal]{wang2023unified}
Ziyang Wang, Yi-Lin Sung, Feng Cheng, Gedas Bertasius, and Mohit Bansal.
\newblock Unified coarse-to-fine alignment for video-text retrieval.
\newblock In \emph{IEEE/CVF International Conference on Computer Vision (ICCV)}, pages 2816--2827, 2023.

\bibitem[Wu et~al.(2023)Wu, Luo, Fang, Wang, and Ouyang]{wu2023cap4video}
Wenhao Wu, Haipeng Luo, Bo Fang, Jingdong Wang, and Wanli Ouyang.
\newblock Cap4video: What can auxiliary captions do for text-video retrieval?
\newblock In \emph{IEEE/CVF Conference on Computer Vision and Pattern Recognition (CVPR)}, pages 10704--10713, 2023.

\bibitem[Wu et~al.(2024)Wu, Wang, Luo, Wang, Yang, and Ouyang]{wu2024cap4video++}
Wenhao Wu, Xiaohan Wang, Haipeng Luo, Jingdong Wang, Yi Yang, and Wanli Ouyang.
\newblock Cap4video++: Enhancing video understanding with auxiliary captions.
\newblock \emph{IEEE Transactions on Pattern Analysis and Machine Intelligence}, 2024.

\bibitem[Xiao et~al.(2021)Xiao, Shang, Yao, and Chua]{xiao2021next}
Junbin Xiao, Xindi Shang, Angela Yao, and Tat-Seng Chua.
\newblock Next-qa: Next phase of question-answering to explaining temporal actions.
\newblock In \emph{IEEE/CVF Conference on Computer Vision and Pattern Recognition (CVPR)}, pages 9777--9786, 2021.

\bibitem[Xu et~al.(2016)Xu, Mei, Yao, and Rui]{xu2016msr}
Jun Xu, Tao Mei, Ting Yao, and Yong Rui.
\newblock Msr-vtt: A large video description dataset for bridging video and language.
\newblock In \emph{IEEE/CVF Conference on Computer Vision and Pattern Recognition (CVPR)}, pages 5288--5296, 2016.

\bibitem[Xu et~al.(2024)Xu, Zhao, Zhou, Lin, Ng, and Feng]{xu2024pllava}
Lin Xu, Yilin Zhao, Daquan Zhou, Zhijie Lin, See~Kiong Ng, and Jiashi Feng.
\newblock Pllava: Parameter-free llava extension from images to videos for video dense captioning.
\newblock \emph{arXiv preprint arXiv:2404.16994}, 2024.

\bibitem[Xue et~al.(2024)Xue, Ashutosh, and Grauman]{xue2024learning}
Zihui Xue, Kumar Ashutosh, and Kristen Grauman.
\newblock Learning object state changes in videos: An open-world perspective.
\newblock In \emph{IEEE/CVF Conference on Computer Vision and Pattern Recognition (CVPR)}, pages 18493--18503, 2024.

\bibitem[Yang et~al.(2021)Yang, Miech, Sivic, Laptev, and Schmid]{yang2021just}
Antoine Yang, Antoine Miech, Josef Sivic, Ivan Laptev, and Cordelia Schmid.
\newblock Just ask: Learning to answer questions from millions of narrated videos.
\newblock In \emph{IEEE/CVF International Conference on Computer Vision (ICCV)}, pages 1686--1697, 2021.

\bibitem[Yang et~al.(2023)Yang, Nagrani, Seo, Miech, Pont-Tuset, Laptev, Sivic, and Schmid]{yang2023vid2seq}
Antoine Yang, Arsha Nagrani, Paul~Hongsuck Seo, Antoine Miech, Jordi Pont-Tuset, Ivan Laptev, Josef Sivic, and Cordelia Schmid.
\newblock Vid2seq: Large-scale pretraining of a visual language model for dense video captioning.
\newblock In \emph{IEEE/CVF Conference on Computer Vision and Pattern Recognition (CVPR)}, pages 10714--10726, 2023.

\bibitem[Ye et~al.(2023)Ye, Xu, Xu, Ye, Yan, Zhou, Wang, Hu, Shi, Shi, et~al.]{ye2023mplug}
Qinghao Ye, Haiyang Xu, Guohai Xu, Jiabo Ye, Ming Yan, Yiyang Zhou, Junyang Wang, Anwen Hu, Pengcheng Shi, Yaya Shi, et~al.
\newblock mplug-owl: Modularization empowers large language models with multimodality.
\newblock \emph{arXiv preprint arXiv:2304.14178}, 2023.

\bibitem[You et~al.(2016)You, Jin, Wang, Fang, and Luo]{you2016image}
Quanzeng You, Hailin Jin, Zhaowen Wang, Chen Fang, and Jiebo Luo.
\newblock Image captioning with semantic attention.
\newblock In \emph{IEEE/CVF Conference on Computer Vision and Pattern Recognition (CVPR)}, pages 4651--4659, 2016.

\bibitem[Yu and Grauman(2014)]{yu2014fine}
Aron Yu and Kristen Grauman.
\newblock Fine-grained visual comparisons with local learning.
\newblock In \emph{IEEE/CVF Conference on Computer Vision and Pattern Recognition (CVPR)}, pages 192--199, 2014.

\bibitem[Yu et~al.(2016)Yu, Wang, Huang, Yang, and Xu]{yu2016video}
Haonan Yu, Jiang Wang, Zhiheng Huang, Yi Yang, and Wei Xu.
\newblock Video paragraph captioning using hierarchical recurrent neural networks.
\newblock In \emph{Proceedings of the IEEE conference on computer vision and pattern recognition}, pages 4584--4593, 2016.

\bibitem[Zeng et~al.(2017)Zeng, Chen, Chuang, Liao, Niebles, and Sun]{zeng2017leveraging}
Kuo-Hao Zeng, Tseng-Hung Chen, Ching-Yao Chuang, Yuan-Hong Liao, Juan~Carlos Niebles, and Min Sun.
\newblock Leveraging video descriptions to learn video question answering.
\newblock In \emph{Association for the Advancement of Artificial Intelligence (AAAI)}, 2017.

\bibitem[Zhai et~al.(2023)Zhai, Mustafa, Kolesnikov, and Beyer]{zhai2023sigmoid}
Xiaohua Zhai, Basil Mustafa, Alexander Kolesnikov, and Lucas Beyer.
\newblock Sigmoid loss for language image pre-training.
\newblock In \emph{IEEE/CVF International Conference on Computer Vision (ICCV)}, pages 11975--11986, 2023.

\bibitem[Zhang et~al.(2023)Zhang, Li, and Bing]{videollama}
Hang Zhang, Xin Li, and Lidong Bing.
\newblock Video-llama: An instruction-tuned audio-visual language model for video understanding.
\newblock \emph{arXiv preprint arXiv:2306.02858}, 2023.

\bibitem[Zhang et~al.(2013)Zhang, Zhu, and Derpanis]{zhang2013actemes}
Weiyu Zhang, Menglong Zhu, and Konstantinos~G Derpanis.
\newblock From actemes to action: A strongly-supervised representation for detailed action understanding.
\newblock In \emph{IEEE/CVF International Conference on Computer Vision (ICCV)}, pages 2248--2255, 2013.

\bibitem[Zhang et~al.(2024)Zhang, Wu, Li, Li, Ma, Liu, and Li]{llava-video}
Yuanhan Zhang, Jinming Wu, Wei Li, Bo Li, Zejun Ma, Ziwei Liu, and Chunyuan Li.
\newblock Video instruction tuning with synthetic data.
\newblock \emph{arXiv preprint arXiv:2410.02713}, 2024.

\bibitem[Zhao et~al.(2023)Zhao, Wang, Ouyang, Dong, Wang, and He]{zhao2023beyond}
Zhiyuan Zhao, Bin Wang, Linke Ouyang, Xiaoyi Dong, Jiaqi Wang, and Conghui He.
\newblock Beyond hallucinations: Enhancing lvlms through hallucination-aware direct preference optimization.
\newblock \emph{arXiv preprint arXiv:2311.16839}, 2023.

\bibitem[Zhou et~al.(2018{\natexlab{a}})Zhou, Xu, and Corso]{youcook}
Luowei Zhou, Chenliang Xu, and Jason Corso.
\newblock Towards automatic learning of procedures from web instructional videos.
\newblock In \emph{Association for the Advancement of Artificial Intelligence (AAAI)}, 2018{\natexlab{a}}.

\bibitem[Zhou et~al.(2018{\natexlab{b}})Zhou, Xu, and Corso]{zhou2018towards}
Luowei Zhou, Chenliang Xu, and Jason Corso.
\newblock Towards automatic learning of procedures from web instructional videos.
\newblock In \emph{Association for the Advancement of Artificial Intelligence (AAAI)}, 2018{\natexlab{b}}.

\bibitem[Zhou et~al.(2024)Zhou, Arnab, Buch, Yan, Myers, Xiong, Nagrani, and Schmid]{zhou2024streaming}
Xingyi Zhou, Anurag Arnab, Shyamal Buch, Shen Yan, Austin Myers, Xuehan Xiong, Arsha Nagrani, and Cordelia Schmid.
\newblock Streaming dense video captioning.
\newblock In \emph{IEEE/CVF Conference on Computer Vision and Pattern Recognition (CVPR)}, pages 18243--18252, 2024.

\end{thebibliography}
}
\clearpage
\setcounter{page}{1}
\maketitlesupplementary
\setcounter{section}{0}
\section{Dataset}
\subsection{FrameCap Training Data}
To construct the FramePair dataset, we employ a suite of open-source VLMs as captioners for initial pseudo label generation, including VILA~\cite{lin2024vila}, Qwen2-VL~\cite{qwen2}, LLAVA-Next-Video~\cite{llava-next}, LLAVA-Video~\cite{llava-video} and LLAVA-OV~\cite{llava-ov}. Training videos are sourced from HowToChange and COIN, with frames extracted at 1FPS. We prepare pairs of frames for stage-I and multi-frame sequences for stage-II; the frame sequence length ranges from 3 to 6, as our preliminary experiments suggest that extending beyond 6 frames causes multiple issues with our captioners,  such as overly brief captions, memory overflows, and great temporal mismatches.

We then process the data through our custom-designed tasks: progression detection and caption matching, to filter for high-quality data. The progression detection uses LLAMA-3.1-70B-Instruct~\cite{dubey2024llama}, and for caption matching, we use VILA~\cite{lin2024vila}, chosen for its open-source availability and strong performance. Specifically, we assess caption matching precision by comparing model-generated answers against human responses on a subset of 90 questions. Gemini-1.5-Pro~\cite{reid2024gemini} achieves a precision of 0.89, while VILA achieves 0.75, the highest among open-source VLMs. Given that Gemini-1.5-Pro API usage incurs a cost, we reserve it for evaluation while utilizing the cost-free VILA as the caption matching evaluation VLM during the pseudo labeling stage. 

\begin{figure}[!t]
  \centering
   \includegraphics[width=1.0\linewidth]{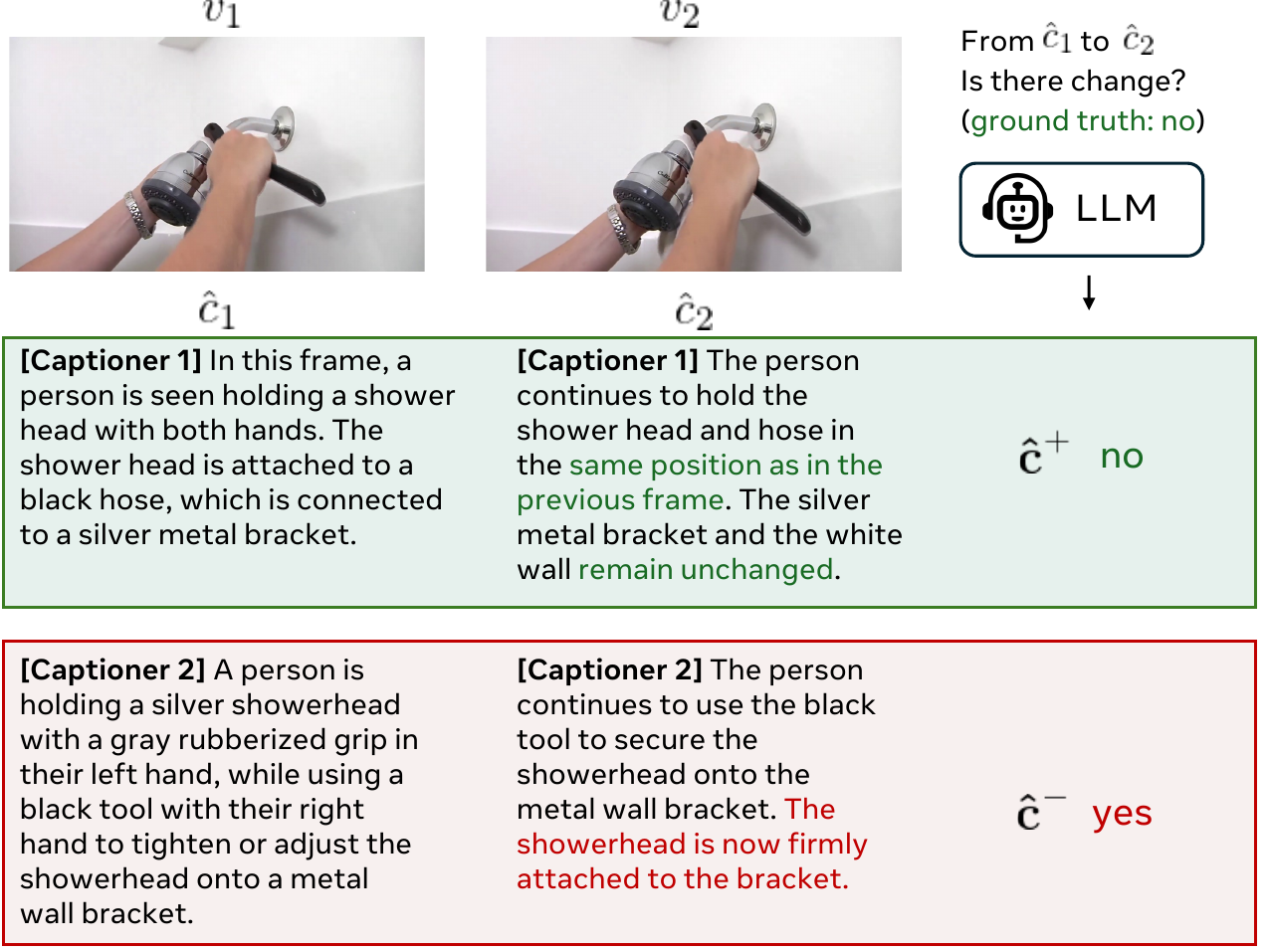}
   \caption{Example of a frame pair (decided by progression detection). The upper caption pair is marked as ``accepted'' by the evaluation LLM, aligning with the ground truth progression label (no progression), while the lower caption pair is marked as ``rejected'' because it incorrectly suggests progression.}
   \label{fig:prefpair_pd}
\end{figure}

\begin{figure}[!t]
  \centering
   \includegraphics[width=1.0\linewidth]{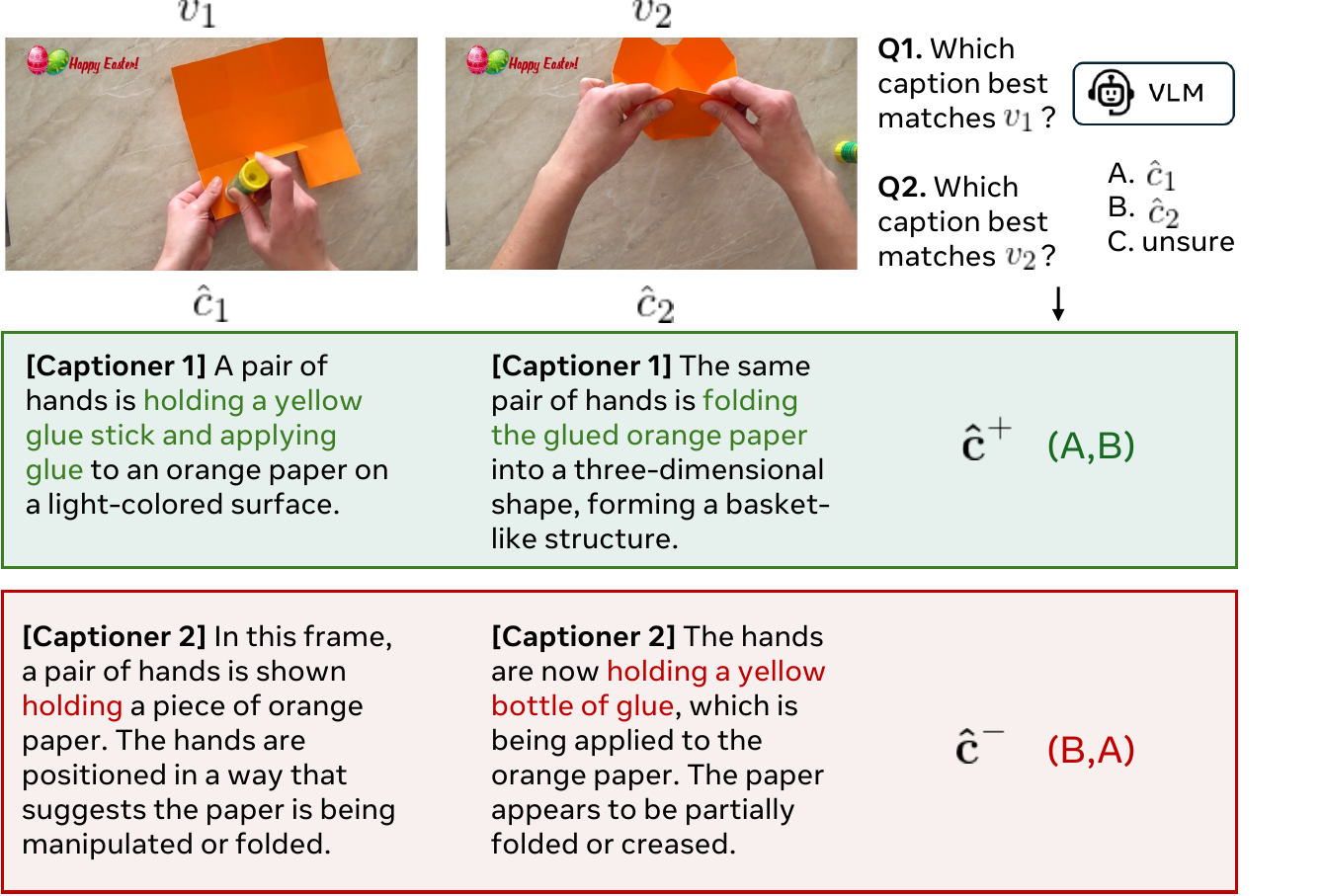}
   \caption{Example of a frame pair (decided by caption matching). The upper caption pair is marked as ``accepted'' since the evaluation VLM correctly answers the caption matching questions as (A, B), demonstrating good alignment. In contrast, the lower pair is ``rejected'' due to its answers (B, A), indicating poor correspondence between the frame and the generated captions.}
   \label{fig:prefpair_cm}
\end{figure}

\begin{figure*}[!t]
  \centering
   \includegraphics[width=1.0\linewidth]{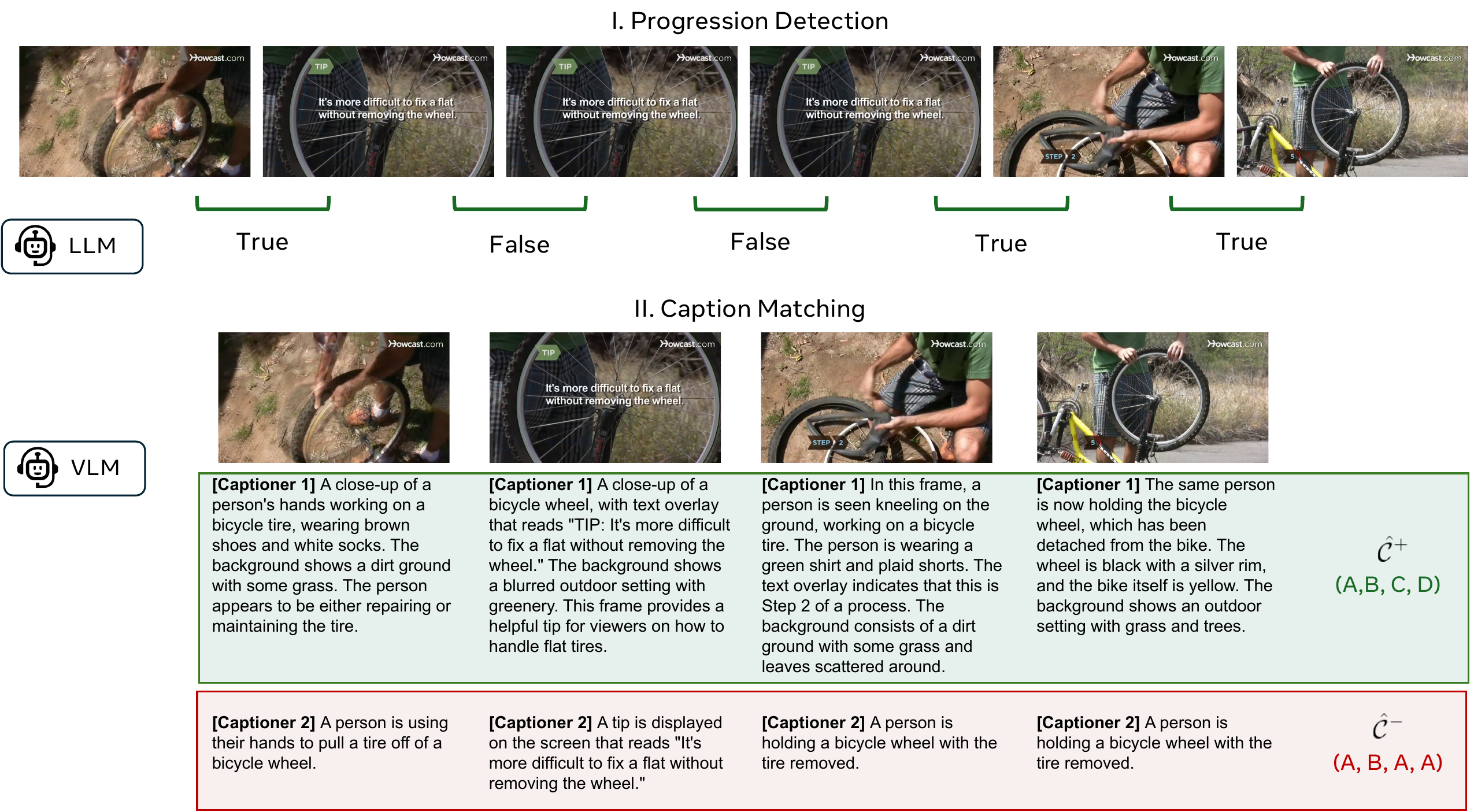}
   \caption{Example of a frame sequence. Progression detection is first applied to each adjacent frame pair to determine the visual-change label and identify $M$ distinct frames. Caption matching then evaluates the captions corresponding to these $M$ frames. The upper caption sequence is ``accepted'' as the evaluation VLM correctly answers (A, B, C, D), whereas the lower caption sequence, leading to erroneous responses, is marked as ``rejected''.}
   \label{fig:prefpair_seq}
\end{figure*}

For each frame sequence, the caption sequence that passes and fails these evaluations forms our preference data, which is utilized for DPO training of ProgressCaptioner. See Figure~\ref{fig:prefpair_pd} and ~\ref{fig:prefpair_cm} for examples of frame pair data obtained from progression detection and caption matching, respectively, and Figure~\ref{fig:prefpair_seq} for an illustration of the frame sequence data preparation process. Table~\ref{tab:train_data} provides a summary of the training data statistics. The first data preparation stage collects a total of 240K frame-caption pairs for supervised fine-tuning (SFT) and 21K preference pairs for direct preference optimization (DPO). The second stage further expands the dataset to include 34K multi-frame and caption sequences for SFT, along with 26K frame-caption sequences for DPO.

\begin{table}[!t]
\tablestyle{2pt}{1.3}
  \centering
\begin{tabular}{l|cccccc}
\thickhline
\multirow{2}{*}{Dataset} & \multirow{2}{*}{\# Videos} & \multirow{2}{*}{\# Frames} & \multicolumn{2}{c}{\# Pair} & \multicolumn{2}{c}{\# Seq} \\
 & & & SFT & DPO & SFT & DPO \\
\hline
HowToChange~\cite{xue2024learning} & 7,812 & 101,369 & 83,383 & 8,453 & 13,602 & 8,362 \\
COIN~\cite{tang2019coin} & 9,030 & 103,791 & 156,858 & 12,622 & 20,704 & 17,826\\
Total & 16,842 & 205,160 & 240,241 & 21,075 & 34,306 & 26,188 \\
\thickhline
\end{tabular}
\caption{We propose the FrameCap data collection, offering large-scale frame and caption sequences for fine-grained frame-level video captioning.}
\label{tab:train_data}
\end{table}

\subsection{FrameCapEval Benchmark}
For the FrameCapEval benchmark, we source videos from four action-focused datasets: HowToChange~\cite{xue2024learning}, COIN~\cite{tang2019coin}, Penn Action~\cite{zhang2013actemes} and Kinetics~\cite{carreira2017quo}. We ensure a balanced selection of videos from each action category across these datasets and follow their original validation or test splits. We are mindful of the single frame bias~\cite{lei2022revealing}—a recognized issue in video understanding where some actions are not distinctly temporal and can be adequately depicted with a single frame. To address this, we conduct a manual verification of all videos to eliminate frame sequences that lack fine-grained action progression, as these scenarios are straightforward and can be adequately managed by image captioning models. Frames are extracted at 1 FPS and grouped using $K$-means clustering based on CLIP features~\cite{radford2021learning}, with $K$ determined by silhouette scores~\cite{rousseeuw1987silhouettes} and ranging from 3 to 6. To each sequence, we add a frame with the smallest CLIP feature distance from a randomly chosen frame, so that the final sequence captures scenarios with and without action progression. See Table~\ref{tab:eval_data} for detailed evaluation data statistics. 

\begin{table}[!t]
\tablestyle{10pt}{1.3}
  \centering
\begin{tabular}{l|cc}
\thickhline
Dataset & \# Videos & \# Frames \\
\hline
HowToChange~\cite{xue2024learning} & 306 (102) & 1101 \\
COIN~\cite{tang2019coin} & 271 (139) & 1063 \\
Penn Action~\cite{zhang2013actemes} & 51 (47) & 235 \\
Kinetics600~\cite{carreira2017quo} & 56 (52) & 451 \\
\thickhline
\end{tabular}
\caption{FrameCapEval data statistics. The numbers in parentheses represent the count of videos used for caption matching. We manually verify all selected frame sequences to assign action progression labels and filter out low-quality (easy) examples lacking clear action progression. This process ensures a robust testbed for evaluating a model's capability to generate temporally fine-grained descriptions.}
\label{tab:eval_data}
\end{table}

FrameCap and FrameCapEval offer unique resources for temporally fine-grained descriptions at the frame level, which can be a valuable enhancement to current VLM's training data.
We will publicly release the two datasets and hope that these resources help advance the temporal precision in video understanding capabilities of VLMs.

\section{Experiments}
\subsection{Experimental setup}
\paragraph{Evaluation Metric Design} Progression detection evaluates a model's action progress awareness, using caption pairs generated for each frame pair. It functions as a binary classification task, where label = 0 identifies scenarios with no visual progression to detect hallucinations, and label = 1 signifies visual progression to assess the model’s ability to capture detailed temporal changes. We measure performance using \emph{balanced accuracy}, which averages the true positive and true negative rates to account for data imbalance. To enhance the reliability and quality of our evaluations, we manually annotate visual progression between frames in the FrameCapEval dataset. Llama-3.1-70B-Instruct~\cite{dubey2024llama} is employed as the evaluation LLM to determine if a caption pair describes visual progression.

Caption matching assesses both the accuracy and the temporal granularity of captions. The evaluation is conducted on $T$-frame sequences that depict action progression, which are manually validated to ensure reliability. Gemini-1.5-Pro~\cite{reid2024gemini} is adopted as the evaluation VLM and tasked with performing the frame-wise caption matching task. We measure \emph{sequence-level accuracy}, defined as the proportion of sequences where every frame is correctly identified by the evaluation VLM among all test sequences. It reflects how many caption sequences are entirely correct, which effectively rules out the possibility of random guessing being successful for a few frames within the sequence, providing a more robust assessment of caption sequence quality.

\CR{\paragraph{User Study} To evaluate the subjective quality of generated captions, we conduct a user study involving 15 graduate student participants fluent in English.} The study utilized 85 randomly sampled frame sequences (totaling 364 frames) from the FrameCapEval benchmark. We evaluate the captions from four leading models—two open-source (LLAVA-OV~\cite{llava-ov}, Qwen2-VL~\cite{qwen2}) and two proprietary (Gemini-1.5-Pro~\cite{reid2024gemini}, GPT-4o~\cite{achiam2023gpt})—alongside our ProgressCaptioner. Note that image captioning baselines are excluded due to their excessively lengthy captions and complete lack of temporal coherence. Participants are presented with captions produced by these five models, randomly shuffled for each sequence, and asked to choose the best and second best (with an additional ``none'' option available) for each frame's caption. The \emph{average selection rate per model} is reported, providing insights into subjective caption quality preferences. 

\paragraph{Implementation} The Stage-I (frame pair captioning) and Stage-II (frame sequence captioning) models are trained with the same hyperparameters and undergo the same training processes: SFT followed by DPO. In the SFT phase, learning rates are set at 1e-5 for the LLM and projector, and 2e-6 for the vision encoder, with a batch size of 64. For DPO, the learning rate is reduced to 5e-7 with a batch size of 8. We set the preference scaling parameter $\alpha$ = 1.0 and the temperature parameter $\beta$ = 0.2.

During inference, ProgressCaptioner takes frame sequences ranging from 2 to 6 frames. This limit is set because, as discussed earlier, all models experience severe performance degradation with longer frame sequences; hence, we cap at 6 frames when preparing training data and keep the inference protocol consistent with training. For sequences exceeding this length, ProgressCaptioner can operate in a sliding window mode. 

For results in Section~\ref{sec:exp_benchmark}, direct inference is applied on $T$ frames. For results in Section~\ref{sec:exp_application}, we employ a 2-frame sliding window, where ProgressCaptioner performs frame pair captioning (except for NeXT-QA, where we uniformly sample 6 frames from the original video and apply direct inference on these 6-frame sequences without a sliding window). A single frame ($v_t$) can receive two captions: one from the pair ($v_{t-1}$, $v_{t}$) and another from ($v_{t}$, $v_{t+1}$). We concatenate the two captions for frame classification tasks to provide richer contextual information, aiding the LLM in frame label prediction. For keyframe selection, we use the caption from the pair ($v_{t-1}$, $v_{t}$) for $v_t$ to maintain caption sequence coherence. 

\subsection{Prompt used} 
We design the following prompt for VLMs to perform the frame-wise video captioning task:
\begin{tcolorbox}[title=Caption Generation Prompt,colbacktitle=gray!20,coltitle=black,fonttitle=\bfseries, myboxstyle] \textbf{Instructions:}

These are {\texttt{T}} frames extracted from a video sequence depicting {\texttt{action}}. Provide a detailed description for each frame.

\medskip

\textbf{Requirement:}

(1) Ensure each frame's description is specific to the corresponding frame, not referencing other frames. 

(2) The description should focus on the specific action being performed, capturing the progression of the action.
There is no need to comment on other elements, such as the background or unrelated objects.

\medskip

\textbf{Reply with the following format:}

\texttt{<Frame 1>}: Your description

\quad\vdots

\texttt{<Frame {T}>}: Your description. 
\end{tcolorbox}

where {\texttt{T}} represents the number of frames in the sequence, and {\texttt{action}} is the video-level action label. The prompt is selected based on preliminary experiments on a small set of data, and we manually review the generated captions to ensure their effectiveness. We use the same prompt consistently for pseudo labeling training data and for evaluating current VLMs, both for our model and existing ones.

% \clearpage 

The progression detection prompt provided to the LLM is as follows: 

\begin{tcolorbox}[title=Progression Detection Prompt (Pseudo-labeling),colbacktitle=gray!20,coltitle=black,fonttitle=\bfseries, myboxstyle]
\textbf{Instructions:}

You will be provided with two image descriptions. Your task is to determine the relationship between the two images based on these descriptions.

\medskip

\textbf{Image 1 description}: {\texttt{desc1}}

\textbf{Image 2 description}: {\texttt{desc2}}

\medskip

Choose the most appropriate option from the following:

\begin{enumerate}[label=\Alph*.] 
\item The images likely look similar (no significant change). 
\item There are noticeable changes between Image 1 and Image 2.
\item It is not possible to determine the similarity or difference based on the descriptions. 
\vspace{-3mm}
\end{enumerate} 
\end{tcolorbox}

\begin{tcolorbox}[title=Progression Detection Prompt (Evaluation),colbacktitle=gray!20,coltitle=black,fonttitle=\bfseries, myboxstyle]
\textbf{Instructions:}

You will be provided with two image descriptions depicting an action. Your task is to determine the relationship between the actions in the two images based on the descriptions provided.

\medskip

\textbf{Action}: {\texttt{action}}

\textbf{The image descriptions are:}

\textbf{Image 1}: {\texttt{desc1}}

\textbf{Image 2}: {\texttt{desc2}}

\medskip

Choose one of the following options:

\begin{enumerate}[label=\Alph*.] 
\item Action Progression: The action has advanced from Image 1 to Image 2 (e.g., more of the task has been completed in Image 2). 
\item No Action Progression: The action remains the same between Image 1 and Image 2 (e.g., the images may show a change in viewpoint, hand position, or slight object adjustments, but the action itself has not progressed). 
\item Uncertain: It is unclear whether the action has progressed or not. 
\vspace{-3mm}
\end{enumerate} 
\end{tcolorbox}

In these prompts, {\texttt{desc1}} and {\texttt{desc2}} represent the descriptions of Image 1 and Image 2, respectively, and {\texttt{action}} is the video-level action label. The progression detection prompts differ between training and evaluation as they serve distinct purposes. For training, we aim to identify visually different frames within a sequence to ensure that the frame sequences processed later by caption matching are composed of distinct frames. Therefore, the training prompt focuses on detecting any visual changes, regardless of their nature. For evaluation, the objective shifts to determining whether the caption sequence is progress-aware; we manually annotate each frame sequence with progression labels for this purpose. As such, the evaluation prompt is designed to discern whether there is action progression or no action progression, rather than identifying simple visual changes. It is important to note that ``changes'' can encompass broader aspects than ``progression'', as explained in the prompt, changes may include viewpoint change or background object adjustments, which do not necessarily indicate a progression in the ongoing action. 

Consider a sequence of $M$ visually distinct frames $\mathcal V_M = \{v_i\}_{i=1}^M$, as detailed in Sec.~\ref{sec:method_model}. We task an evaluation VLM to perform caption matching for each frame $v_m \in \mathcal V_M$ with the following prompt:
\begin{tcolorbox}[title=Caption Matching Prompt,colbacktitle=gray!20,coltitle=black,fonttitle=\bfseries, myboxstyle]

Which caption best describes the image?

\medskip

\textless Frame $v_m$\textgreater

\medskip

A. Caption $\hat c_1$

\qquad $\vdots$

M. Caption $\hat c_M$  

M+1. None of the above descriptions match the image, are hard to determine, or contain incorrect information about the image.

\medskip 

Reply with only the corresponding letter (A, B, C, etc.)
\end{tcolorbox}
where \textless Frame $v_m$\textgreater~denotes the image input (the $m$-th frame), and $\{\hat c_i\}_{i=1}^M$ is the caption sequence to be evaluated. 

\subsection{Results}
\paragraph{Ablation Study} Table~\ref{tab:result_ablation} presents an ablation study using HowToChange videos from FrameCapEval, focusing on three key variables: (1) comparisons between Stage-I and Stage-II models; (2) the effect of training datasets—HowToChange alone versus HowToChange combined with COIN; (3) the impact of SFT alone versus SFT followed by DPO. The results demonstrate that all three factors are crucial for optimal performance. First, the Stage-I model, limited to frame pair captioning, does not provide caption matching accuracy for $T$-frame sequences and shows lower progression detection performance compared to the Stage-II model, which benefits from additional frame sequence training (see row 1 vs. row 2). Second, regarding training data, while evaluation is conducted on HowToChange, incorporating COIN data for training greatly improves performance, particularly in caption matching, highlighting the benefits of data scaling (see row 2 vs. row 4). This indicates potential for further enhancements by adding more datasets in the future. Finally, direct preference optimization (DPO) proves critical as its absence leads to great performance declines (see row 3 vs. row 4). 

\begin{table*}[!htbp]
  \centering
  \begin{tabularx}{\textwidth}{X X}
    % Left column: The table
    \begin{minipage}[t]{\linewidth}
      \centering
      \small % Reduce font size in the table
      \setlength{\tabcolsep}{4pt} % Reduce horizontal padding between columns
      \begin{tabular}{ccc|cc}
        \thickhline
        Model & Training Data & Training Strategy & Cap & Prog \\
        \hline
        Stage-I & HTC & SFT+DPO & -- & 70.6 \\
        Stage-II & HTC & SFT+DPO & 28.4 & 73.1 \\
        Stage-II & HTC+COIN & SFT & 24.5 & 68.3 \\
        \rowcolor{LighterGray} 
        Stage-II & HTC+COIN & SFT+DPO & \textbf{37.3} & \textbf{73.6} \\
        \thickhline
      \end{tabular}
      \captionof{table}{Ablation study of ProgressCaptioner on HowToChange (HTC) evaluation videos: examining the impact of model stages, training data, and optimization strategies on performance.}
      \label{tab:result_ablation}
    \end{minipage}
    &
    % Right column: The paragraph
    \begin{minipage}[t]{\linewidth}
      \setlength{\parindent}{0pt} % Remove paragraph indentation
      \vspace{-12mm} % Adjust vertical alignment if necessary
      \textbf{More Qualitative Results}\par
      Supplementing Figure~\ref{fig:qualitative} in the main paper, Figures~\ref{fig:qualitative_supp1}--\ref{fig:qualitative_supp3} provide more qualitative predictions, comparing ProgressCaptioner with the four leading VLMs. As can be seen from these examples, while all baseline VLMs exhibit temporal inaccuracies in their descriptions, ProgressCaptioner consistently provides temporally precise and progress-aware captions, highlighting its superior performance.
    \end{minipage}
  \end{tabularx}
\end{table*}

\FloatBarrier % Prevents floats from moving past this point

\begin{figure*}[!htbp]
  \centering
  \includegraphics[width=0.85\linewidth]{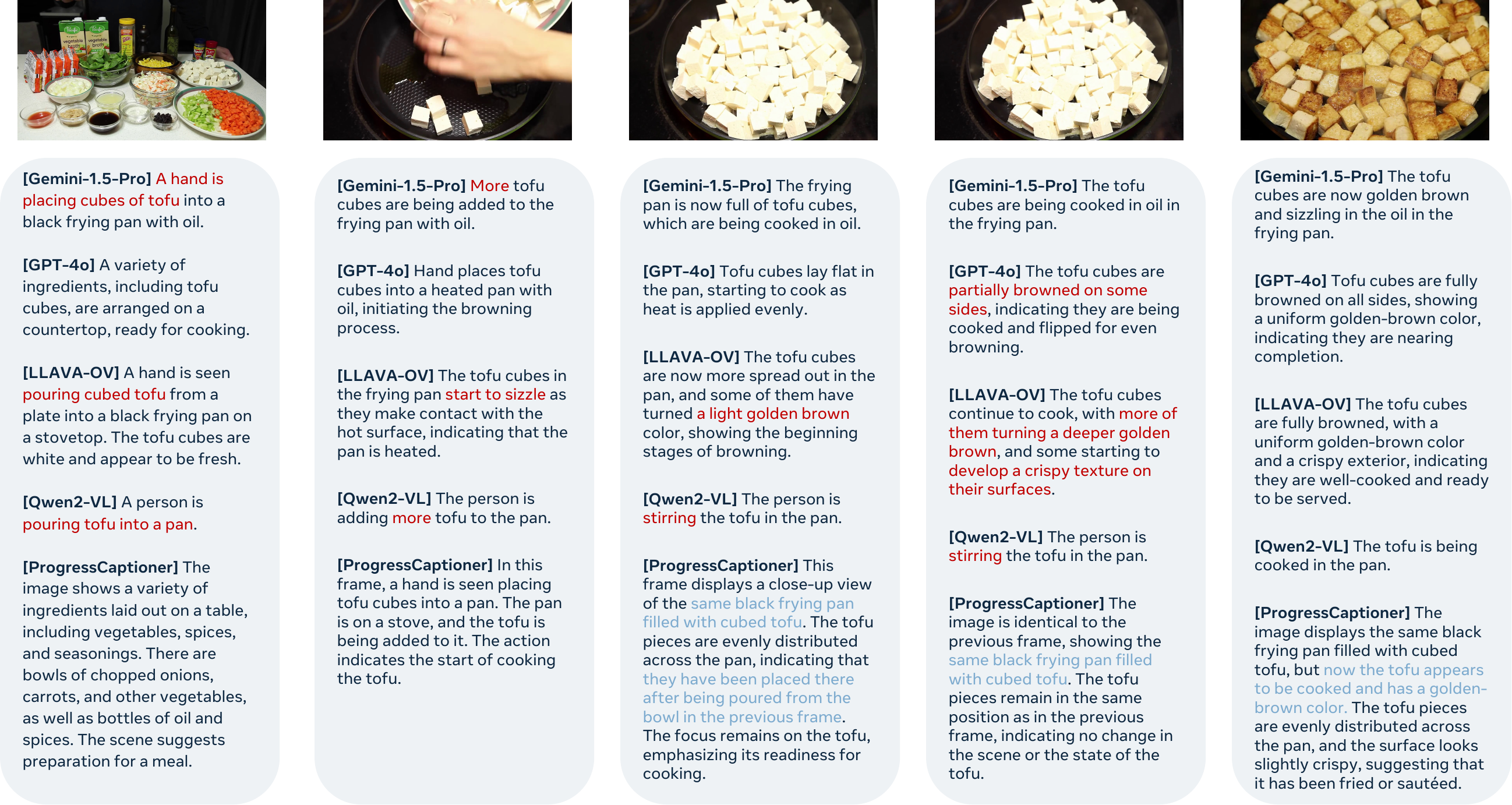}
  \includegraphics[width=0.85\linewidth]{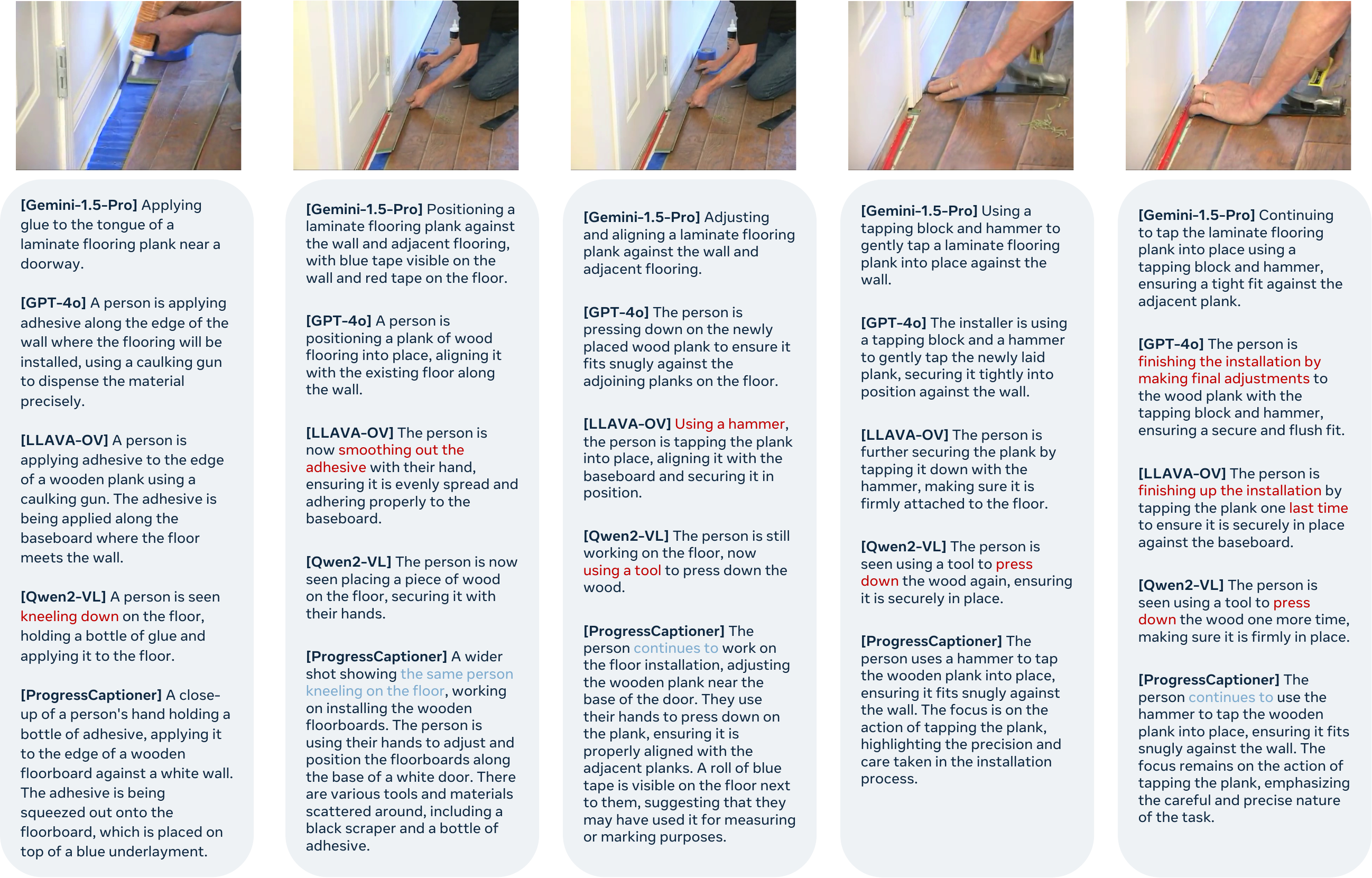}
  \caption{Qualitative comparisons of ProgressCaptioner with SOTA VLMs (I). Red text identifies inaccuracies in the generated captions, while blue text highlights how our progress-aware captions build on prior content to clearly delineate what is changing or continuing.}
  \label{fig:qualitative_supp1}
\end{figure*}

\FloatBarrier % Ensures that no floats are placed beyond this point
% *****

\begin{figure*}[htbp]
  \centering
   \includegraphics[width=0.85\linewidth]{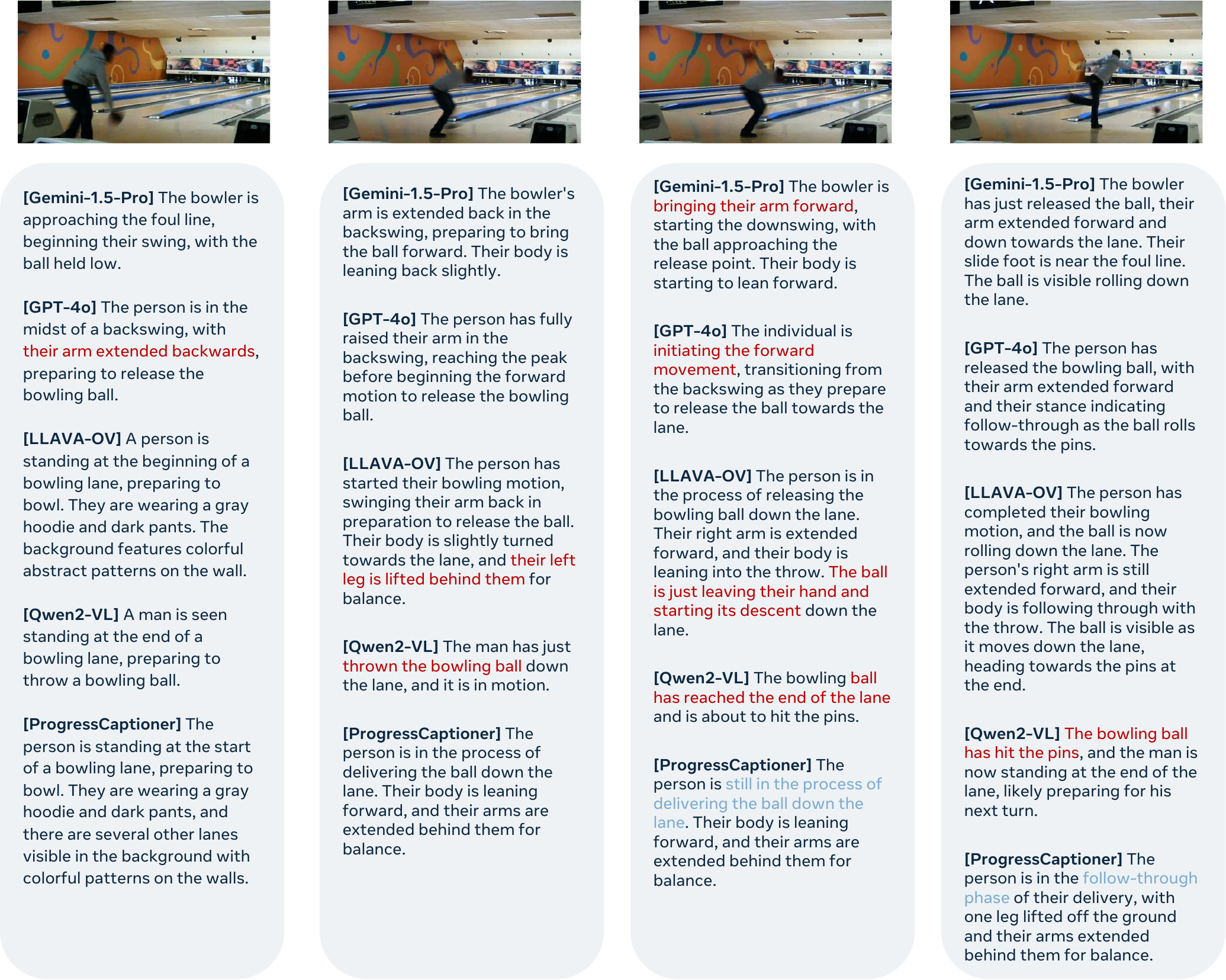}
   \includegraphics[width=1.0\linewidth]{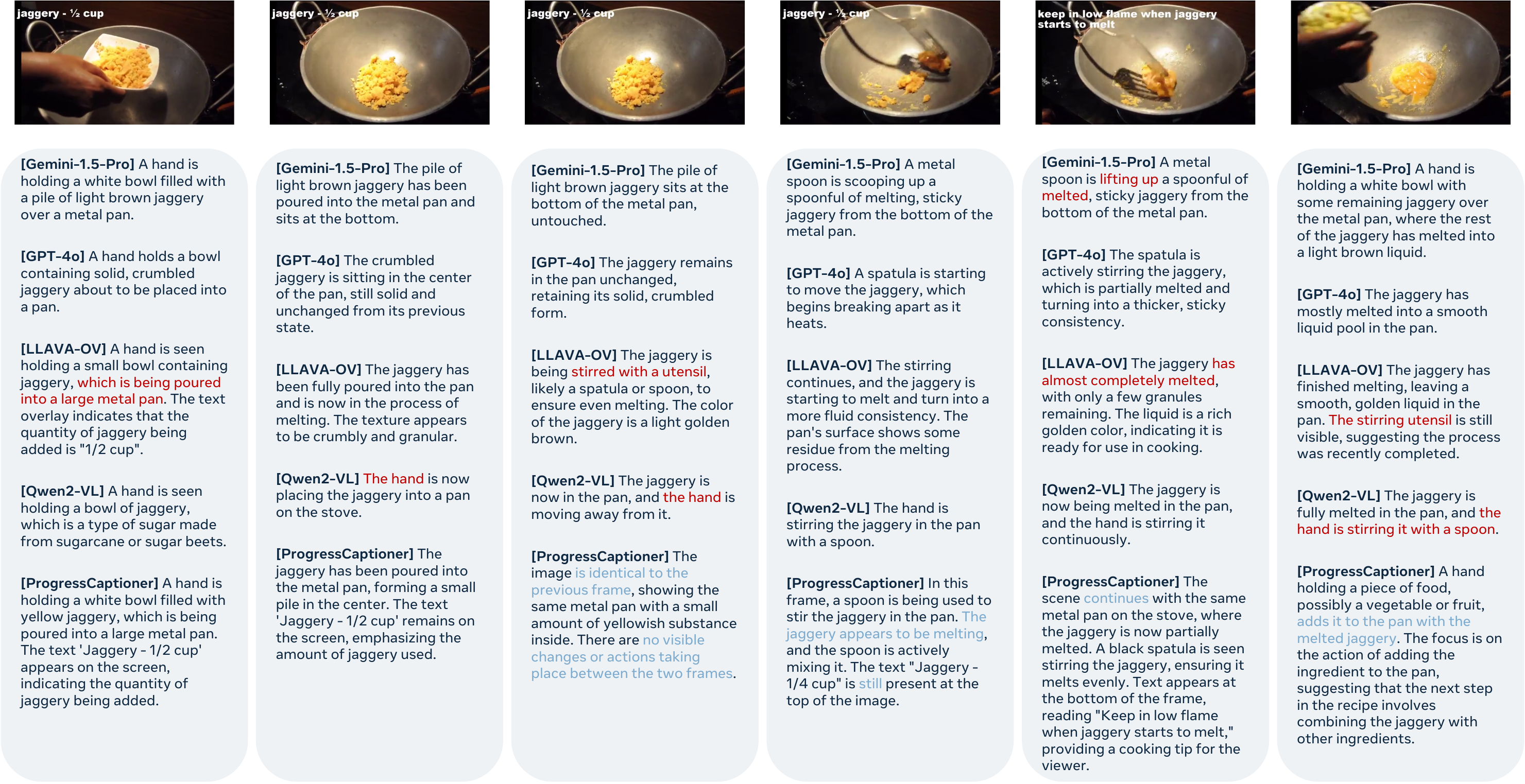}
   \caption{Qualitative comparisons of ProgressCaptioner with SOTA VLMs (II). Red text identifies inaccuracies in the generated captions, while blue text highlights how our progress-aware captions build on prior content to clearly delineate what is changing or continuing.}
   \label{fig:qualitative_supp2}
\end{figure*}

\begin{figure*}[htbp]
  \centering
   \includegraphics[width=1.0\linewidth]{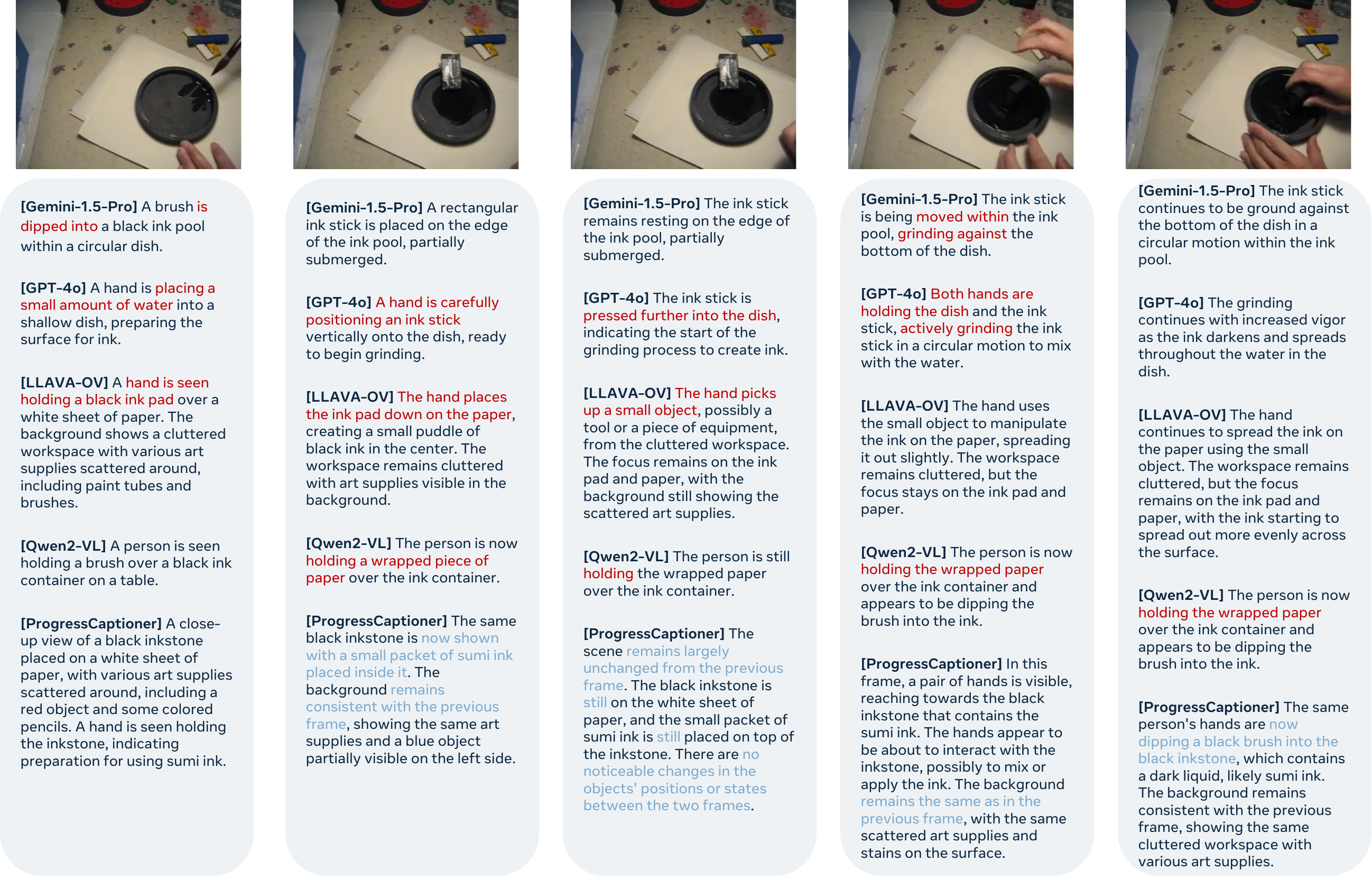}
   \includegraphics[width=1.0\linewidth]{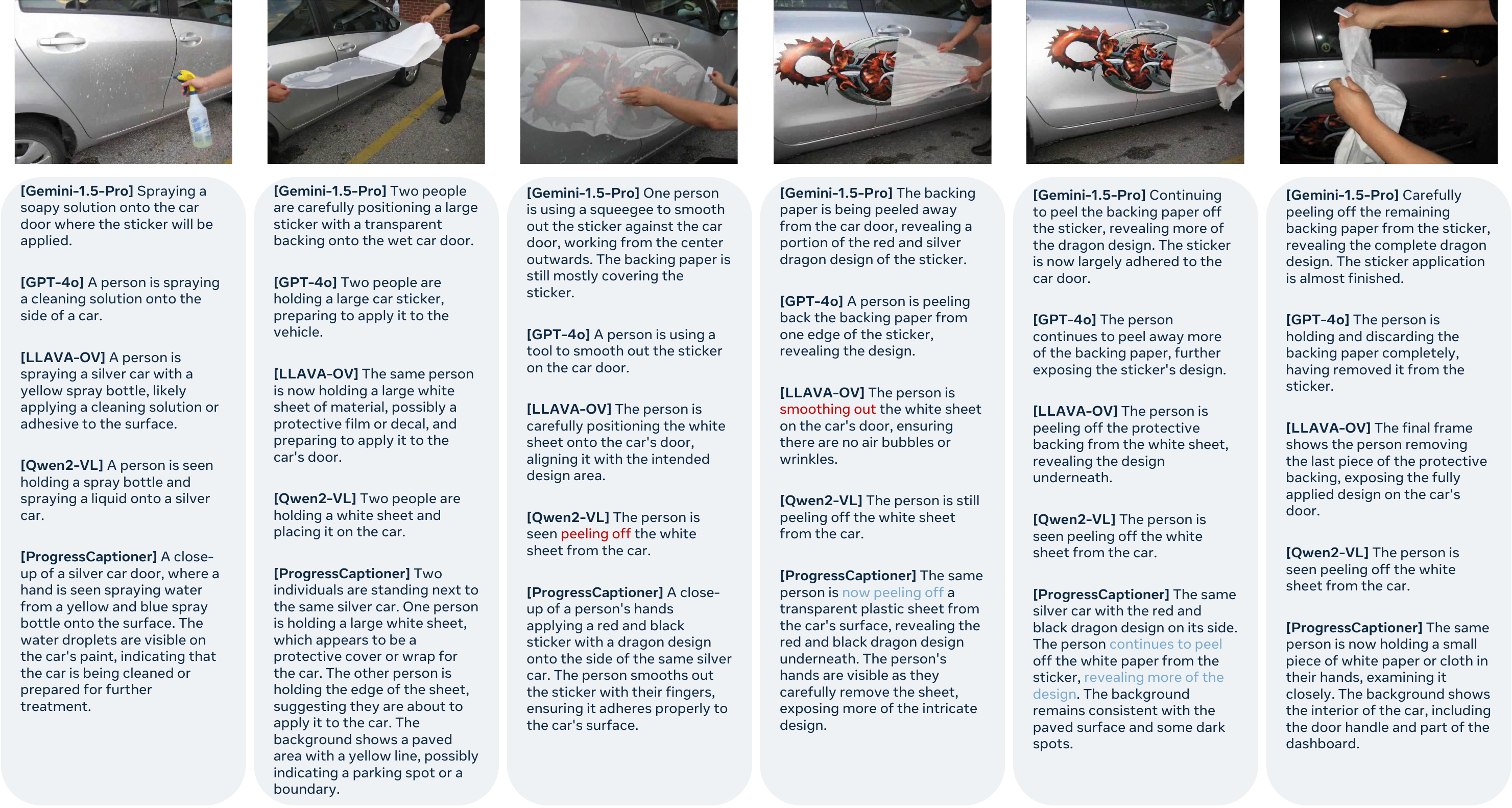}
   \caption{Qualitative comparisons of ProgressCaptioner with SOTA VLMs (III). Red text identifies inaccuracies in the generated captions, while blue text highlights how our progress-aware captions build on prior content to clearly delineate what is changing or continuing.}
   \label{fig:qualitative_supp3}
\end{figure*}
\FloatBarrier

\begin{figure*}[!htb]
  \centering
   \includegraphics[width=0.83\linewidth]{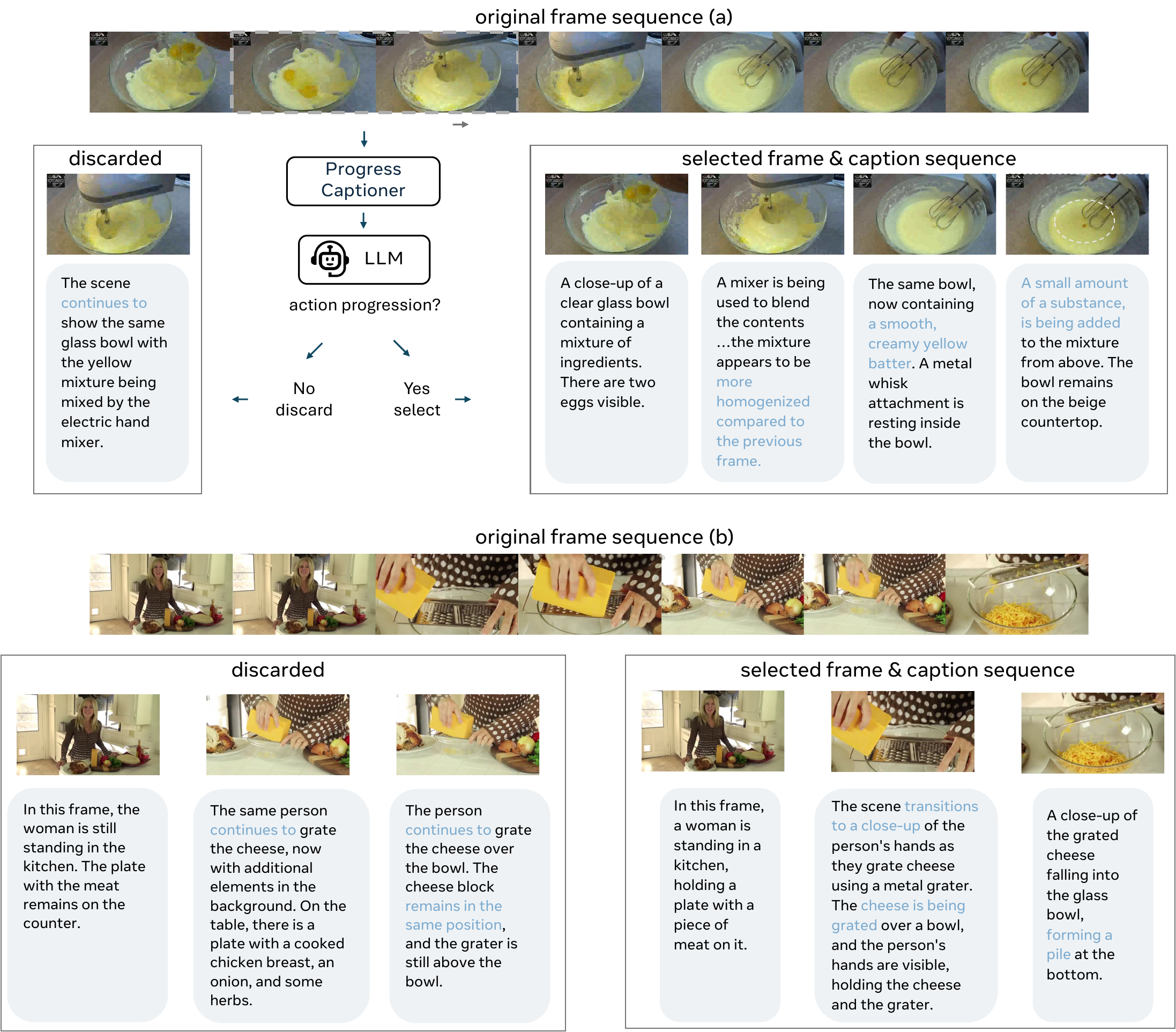}
   \caption{Captions produced by ProgressCaptioner and processed by an LLM enable us to automatically select representative frames that clearly depict action progression from densely sampled frame sequences. For each frame sequence, the bottom left box displays discarded frames alongside their captions, while the bottom right box showcases selected frames and their corresponding captions. This process effectively removes duplicate frames that depict the same action progression and enhances the selected frames with captions.}
   \label{fig:keyframe_design}
\end{figure*}

\paragraph{Keyframe Selection} 
We propose to utilize frame-wise captions from ProgressCaptioner to select frames that depict action progression. The key idea is to ``encode'' a sequence of densely sampled video frames into per-frame captions, allowing an LLM to subsequently ``decode'' and identify key frames from this rich textual representation. The temporally fine-grained descriptions act as a condensed frame representation, focusing on action progression while remaining robust to visual disturbances such as changes in viewpoint or background objects. Figure~\ref{fig:keyframe_design} illustrates one potential design for such a keyframe selection feature. With ProgressCaptioner, we employ a sliding two-frame window for captioning, followed by an LLM (we use Llama-3.1-70B-Instruct) processing the generated captions. Specifically, for a sequence of densely sampled frames $\{v_t\}_{t=1}^T$, starting from $t=1$, ProgressCaptioner generates caption ($c_1$, $c_2$) for ($v_1$, $v_2$). We then ask the LLM to determine if there is action progression between $c_1$ and $c_2$. If the answer is yes, frame $v_2$ gets selected; if no, $v_2$ is skipped to avoid redundancy as it likely depicts the same action stage as $v_1$.  The process is repeated by advancing the window to ($v_2$, $v_3$) and continuing through the sequence.

Our approach offers two key advantages: (1) it efficiently filters out non-essential frames to ensure that selected frames distinctly represent action progression, and (2) it dynamically determines the size of the keyframe set based on the sequence content, eliminating the need for manually specifying the number of frames to sub-sample. To better illustrate this, we compare our method with the pseudo labeling strategy used in a recent video summarization work, V2Xum~\cite{hua2024v2xum}. V2Xum employs an image captioning model followed by an LLM to perform extractive document summarization based on per-frame captions for keyframe selection.

\begin{figure*}[!htb]
  \centering
   \includegraphics[width=0.83\linewidth]{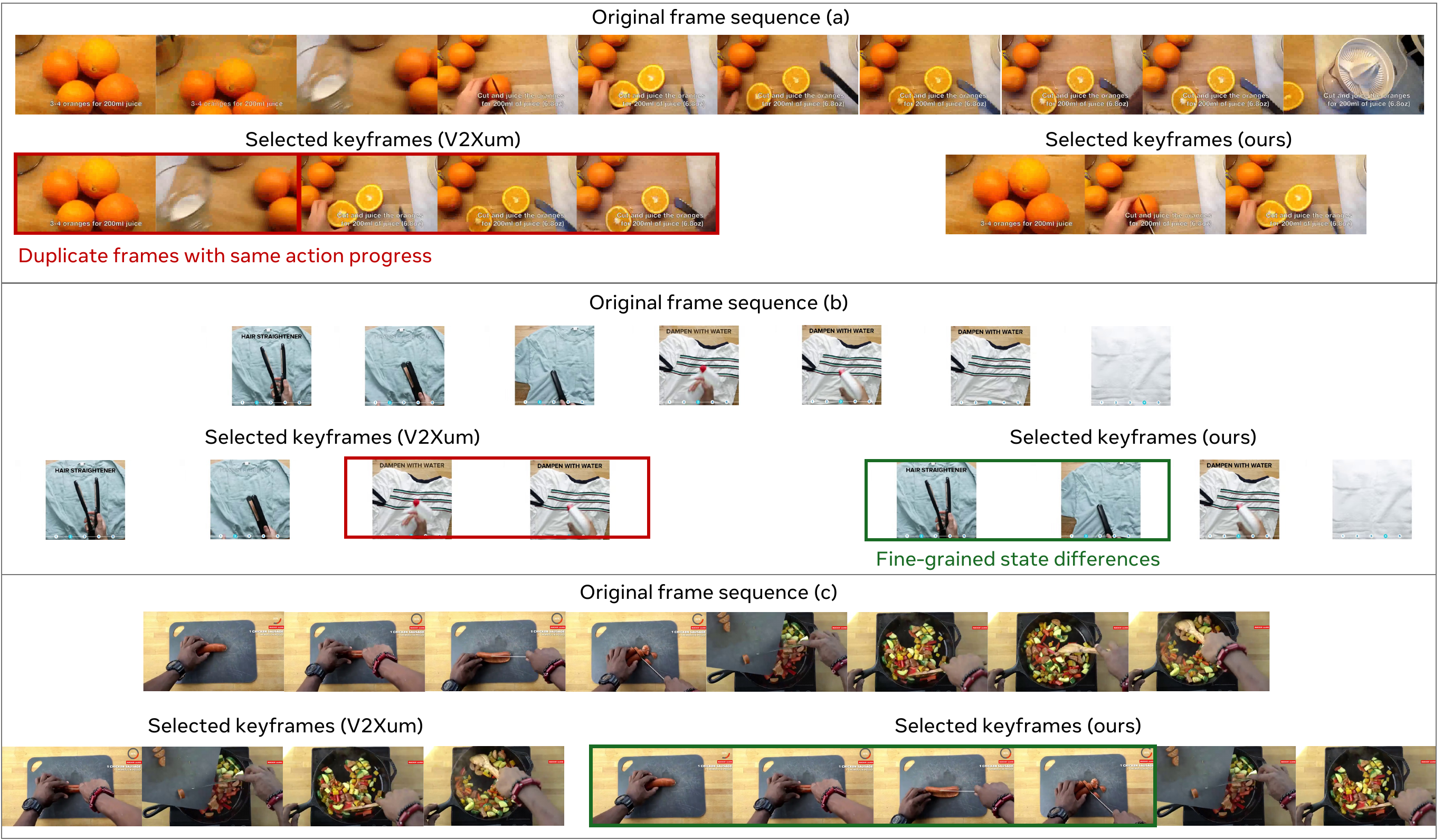}
   \caption{Comparison of our keyframe selection with V2Xum~\cite{hua2024v2xum}. Leveraging precise and progress-aware captions from ProgressCaptioner, our approach selects keyframes that accurately represent stages of the action process. In contrast, V2Xum's method often includes duplicate frames or overlooks frames that show subtle but important differences.}
   \label{fig:keyframe_cmp}
\end{figure*}

As shown in Figure~\ref{fig:keyframe_cmp}, V2Xum's approach results in duplicate keyframes for sequence (a), where the first and second frames depict the same action progression despite a viewpoint change, and the last three frames similarly represent the action progression of oranges being sliced in half. In contrast, our method, leveraging the more accurate and temporally fine-grained captions produced by ProgressCaptioner, precisely identifies three distinct stages of this slicing action sequence. For sequence (c), V2Xum selects only one frame from the first four, despite depicting various stages of cutting a sausage (from whole to partially cut, fully cut, and then to chunks). Conversely, our approach accurately identifies all these frames as markers of action progression. It adaptively determines the size of the keyframe set, which can vary from small to large depending on the actual content, offering flexibility without requiring manual specification.

To conclude, our keyframe selection approach effectively highlights critical moments within action sequences. We believe such a system has significant potential for providing focused insights in educational tutorials and sports analysis, benefiting learners and analysts alike.

\begin{figure}[t!]
    \centering
    \includegraphics[width=1.0\linewidth]{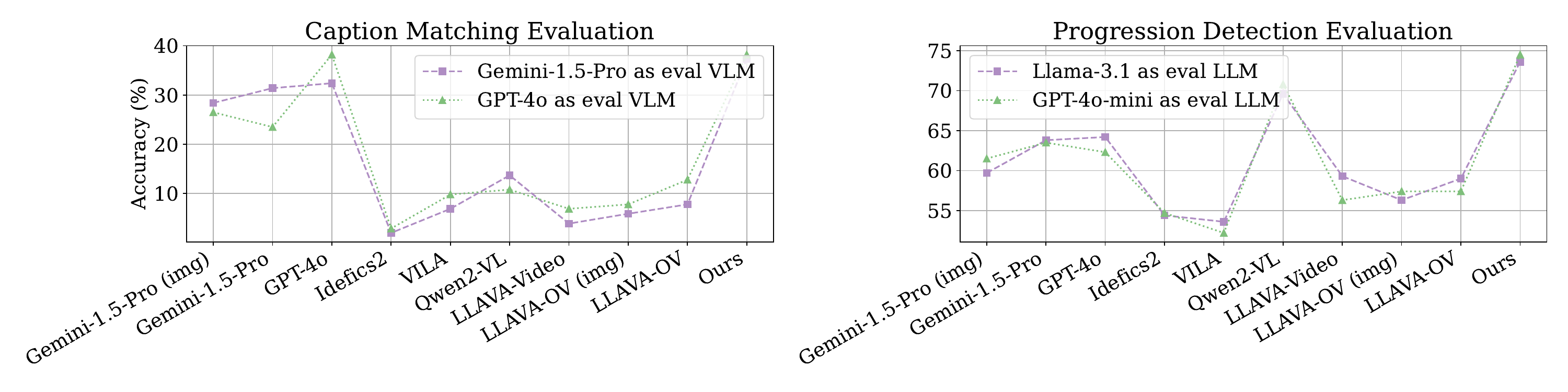}
    \caption{Caption matching (left) and progress detection (right) evaluation results on HowToChange, with different VLM/LLM as evaluators.}
    \label{fig:caption_eval}
\end{figure}

\begin{figure}[t]
  \centering
   \includegraphics[width=1.0\linewidth]{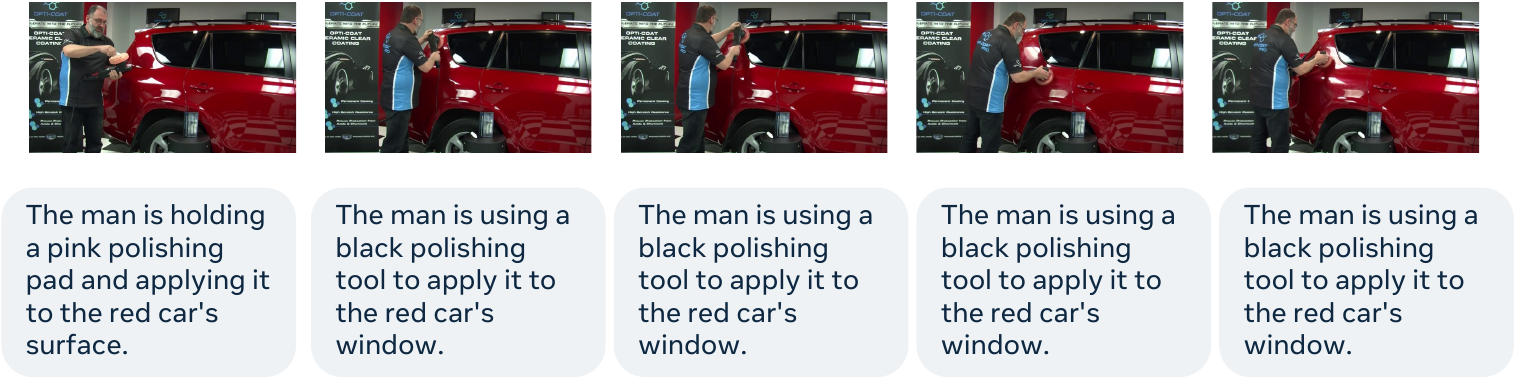}
   \caption{One failure case of ProgressCaptioner, where it fails to discern fine-grained spatial differences among the last four frames and thus produces identical captions.}
   \label{fig:failure_case}
\end{figure}
% \FloatBarrier

\CR{\paragraph{Justification of Automatic Evaluation} Due to the lack of existing datasets with frame-wise ground truth captions, direct reference-based evaluation is infeasible.  Therefore, we propose two automatic evaluation tasks, progression detection and caption matching, to assess frame-wise caption quality. To validate the reliability of these two metrics, we conduct experiments using different LLM/VLMs as evaluators for the two metrics (we pick the most widely adopted ones—Gemini and GPT for VLMs, Llama and GPT for LLMs). Figure~\ref{fig:caption_eval} demonstrates consistent trends across these different evaluators, confirming the robustness of our evaluation methodology.}

\paragraph{Limitations} Despite the enhanced performance of ProgressCaptioner, it still faces several challenges. Firstly, while we have developed an advanced pseudo labeling refinement process, the training data sourced from existing VLMs inherently limits the quality of the captions. Moreover, the automation of data filtering using evaluation LLMs and VLMs introduces noise—though less costly, it's not as reliable as human annotation. Secondly, we observe that captioning longer frame sequences presents increased difficulties; for instance, accurately captioning six-frame sequences is notably more challenging than two-frame sequences. Addressing this challenge to extend ProgressCaptioner's capabilities to handle longer sequences remains a critical area for future development. In addition, Figure~\ref{fig:failure_case} illustrates a failure case where ProgressCaptioner produces identical captions for the last four frames, failing to recognize fine-grained spatial changes—an area that current VLMs consistently fall short of. This underscores the need for further advancements in this area.

Finally, we emphasize that the task of video frame captioning introduces a significant challenge by demanding high temporal precision. We recognize the limitations of ProgressCaptioner in its current stage and view this work as an initial step toward resolving this problem.

\clearpage

\end{document}